\documentclass{article}

\usepackage{arxiv}
\usepackage[utf8]{inputenc} 
\usepackage[T1]{fontenc}    
\usepackage{hyperref}       
\usepackage{url}            
\usepackage{booktabs}       
\usepackage{amsfonts}       
\usepackage{nicefrac}       
\usepackage{microtype}      
\usepackage{lipsum}		
\usepackage{graphicx}
\usepackage{natbib}
\usepackage{doi}
\usepackage{float}
\usepackage{eqndefns-left}
\RequirePackage{newtxtext}
\RequirePackage{newtxmath}
\RequirePackage{bm}
\RequirePackage{endnotes}

\usepackage{algorithm}
\usepackage{algpseudocode}
\usepackage{multirow}
\usepackage{hyperref}
\usepackage{tikz}
\usepackage{booktabs}
\usetikzlibrary{positioning,arrows.meta}
\usepackage{graphicx}
\usepackage{subcaption} 
\usepackage{xcolor}       

\makeatletter
\@ifundefined{openbox}{}{}
\makeatother

\usepackage{amsthm}

\theoremstyle{definition}
\newtheorem{definition}{Definition}[section]
\newtheorem{assumption}{Assumption}[section]

\theoremstyle{plain}
\newtheorem{theorem}{Theorem}[section]
\newtheorem{proposition}{Proposition}[section]

\theoremstyle{remark}
\newtheorem{remark}{Remark}[section]

\usepackage{threeparttable}
\usepackage{caption}

\usepackage{natbib}
\bibpunct[, ]{(}{)}{,}{a}{}{,}%

\title{Finance-Informed Neural Network: Learning the Geometry of Option Pricing}

\author{
Amine Mohamed Aboussalah \\
NYU Tandon School of Engineering \\
New York University \\
New York, NY, USA \\
\texttt{ama10288@nyu.edu}
\And
Xuanze Li \\
Department of Mechanical and Industrial Engineering \\
University of Toronto \\
Toronto, ON, Canada \\
\texttt{xuanze.li@mail.utoronto.ca}
\And
Cheng Chi \\
Department of Mechanical and Industrial Engineering \\
University of Toronto \\
Toronto, ON, Canada \\
\texttt{cheng.chi9948@gmail.com}
\And
Raj G Patel \\
Princeton University \\
Princeton, NJ, USA \\
\texttt{rp7104@princeton.edu}
}

\renewcommand{\shorttitle}{Finance-Informed Neural Network}

\hypersetup{
pdftitle={A template for the arxiv style},
pdfsubject={q-bio.NC, q-bio.QM},
pdfauthor={David S.~Hippocampus, Elias D.~Striatum},
pdfkeywords={First keyword, Second keyword, More},
}

\begin{document}
\maketitle

\begin{abstract}
We propose a Finance-Informed Neural Network (FINN) for option pricing and hedging that integrates financial theory directly into machine learning. Instead of training on observed option prices, FINN is learned through a self-supervised replication objective based on dynamic hedging, ensuring economic consistency by construction. We show theoretically that minimizing replication error recovers the arbitrage-free pricing operator and yields economically meaningful sensitivities. Empirically, FINN accurately recovers classical Black--Scholes prices and performs robustly in stochastic volatility environments, including the Heston model, while remaining stable in settings where analytical solutions are unavailable or unreliable. Fundamental pricing relationships such as put--call parity emerge endogenously. When applied to implied-volatility surface reconstruction, FINN produces surfaces that are consistently closer to observed market-implied volatilities than those obtained from Heston calibrations, indicating superior out-of-sample adaptability and reduced structural bias. Importantly, FINN extends beyond liquid option markets: it can be trained directly on historical spot prices to construct coherent option prices and Greeks for assets with no listed options. More broadly, FINN defines a new paradigm for financial pricing, in which prices are learned from replication and risk-control principles rather than inferred from parametric assumptions or direct supervision on option prices. By reframing option pricing as the learning of a pricing operator rather than the fitting of prices, FINN offers practitioners a practical and scalable tool for pricing, hedging, and risk management across both established and emerging financial markets.
\end{abstract}

\keywords{Option pricing; Pricing geometry; Replication and hedging; Machine learning in finance; Neural networks; Implied volatility surface; Stochastic volatility}

\section{Introduction}\label{sec:Intro}
Accurate pricing and hedging of financial options is a central task for institutions engaged in trading, risk management, and capital allocation. In practice, however, managers and quantitative analysts face a persistent dilemma regarding the models used to support these decisions. On the one hand, traditional principle-driven models are grounded in economic theory and no-arbitrage arguments, making them interpretable and auditable, but they often rely on restrictive assumptions that limit empirical accuracy. On the other hand, modern machine learning (ML) models can achieve high predictive performance by easily fitting market data, yet they typically operate as black boxes and may violate fundamental financial principles. This conflict between theoretical consistency and empirical performance poses a central challenge for model risk management, validation, and regulatory compliance in quantitative finance.

Existing approaches to option pricing largely fall on one side of this dichotomy. Classical methods based on partial differential equations (PDEs) and stochastic control derive prices from no-arbitrage conditions and dynamic replication arguments \cite{black_scholes}. While theoretically sound, their applicability is often constrained by simplifying assumptions, such as constant volatility, that are inconsistent with observed market behavior \cite{YANG20111702}, and by computational challenges in high-dimensional or path-dependent settings \cite{PARK201213243}. In contrast, data-driven techniques leverage advances in ML to capture complex nonlinear relationships in financial data and often deliver strong empirical performance \cite{Anders1998369}. However, because these methods are typically trained to minimize statistical prediction error rather than economically meaningful criteria, they may produce prices and sensitivities that violate no-arbitrage conditions, weakening interpretability, robustness, and trustworthiness, especially in regimes with limited or biased data \cite{Anders1998369}.

To address this trade-off, we propose the \emph{Finance-Informed Neural Network} (FINN), a hybrid framework that integrates financial theory directly into the learning objective of a neural network. The key idea underlying FINN is to replace supervision by observed option prices with a self-supervised, economically grounded training signal. Specifically, the network is trained to minimize the replication error of a dynamically hedged portfolio, thus embedding no-arbitrage conditions derived from hedging arguments directly into the loss function. By learning prices and sensitivities that support effective replication, FINN enforces economic consistency by construction while retaining the flexibility of modern deep learning architectures. As a result, the framework avoids reliance on labeled price data and is not restricted to a specific parametric form of the underlying asset dynamics.

We evaluate FINN across a range of market environments. In settings characterized by constant volatility, the framework accurately recovers the classical Black--Scholes pricing function. More importantly, FINN exhibits strong robustness and improved accuracy in more complex environments governed by stochastic volatility, such as the Heston model, which better reflects empirical market behavior. Despite not being explicitly imposed during training, fundamental no-arbitrage relationships—especially put--call parity—emerge naturally from the learned pricing functions. We further demonstrate the practical relevance of the approach by extending FINN to support advanced risk management strategies, showing that delta--gamma hedging based on FINN substantially improves hedging performance relative to traditional approaches. The framework is easy to extend, allowing higher-order sensitivities to be incorporated as needed for more complex risk profiles.

The primary contributions of this paper are fourfold:
\begin{enumerate}
    \item \textbf{A finance-informed learning framework:} We introduce FINN, a hybrid neural network approach that embeds no-arbitrage principles derived from dynamic replication directly into the training objective, bridging theory-driven and data-driven methods.
    \item \textbf{Self-supervision via replication consistency:} We propose a self-supervised learning methodology in which option prices and sensitivities are learned by minimizing economically meaningful hedging errors rather than statistical pricing errors.
    \item \textbf{Robust generalization across market regimes:} We demonstrate that FINN generalizes effectively from simple diffusion models to stochastic volatility environments, delivering stable and accurate pricing and hedging performance, and we also prove its robustness using examples from real-world market data.
    \item \textbf{Implications for risk management:} We show that the framework naturally supports advanced hedging strategies, including delta--gamma hedging, highlighting its practical value for decision-making in risk management contexts.
\end{enumerate}

The remainder of the paper is organized as follows. Section~2 reviews related work on machine learning methods for option pricing. Section~3 introduces the FINN framework and its theoretical foundations. Section~4 describes the experimental setup, while Section~5 presents and discusses the empirical results, including several case studies. Section~6 concludes and outlines directions for future research.

\section{Related Literature}\label{sec:lit_review}

Machine learning (ML) has emerged as a powerful alternative to classical option pricing approaches grounded in stochastic calculus \cite{black_scholes}. Recent research falls into three main streams: (i) \emph{direct price prediction}, (ii) \emph{neural PDE solvers}, and (iii) emerging \emph{hybrid frameworks} that embed financial theory into ML architectures.

\subsection{Direct Price Prediction}
Early applications of artificial neural networks (ANNs) demonstrated their ability to approximate complex pricing relationships without rigid parametric assumptions \cite{Hutchinson1994}. Over time, richer input features—underlying asset price, strike, maturity, dividend yield, interest rate, and implied volatility—improved accuracy \cite{ANN_OP_2017}, while more expressive architectures such as recurrent neural networks (RNNs) and long short-term memory (LSTM) models captured temporal dependencies and integrated alternative signals like market sentiment \cite{time_series_OP, sentiment_OP}. Additional innovations include modular designs that segment the input space by moneyness or maturity \cite{Gradojevic2009626}, and probabilistic methods such as Gaussian Processes that provide uncertainty estimates alongside predictions \cite{HAN2008515, YANG20111702, PARK201213243}. Reinforcement learning (RL) has also been applied to jointly price and hedge derivatives in a sequential decision-making framework \cite{grassl2010, martin2014, halperin2019, Marzban03102023}. While these methods offer flexibility and strong empirical performance \cite{PARK20145227, IVASCU2021113799}, they generally lack explicit enforcement of no-arbitrage constraints, leaving open the possibility of economically inconsistent prices and hedge ratios.

\subsection{Neural PDE Solvers}

A second stream uses neural networks to solve the partial differential equations (PDEs) that arise from stochastic models of asset dynamics. By linking stochastic differential equations (SDEs) to PDEs via Itô’s lemma, the pricing problem can be recast as a forward–backward stochastic differential equation (FBSDE) with terminal conditions given by the payoff. Within this stream, physics-informed approaches often solve these problems through fitting a neural network by penalizing the PDE residual together with boundary/terminal condition violations, therefore turning the option pricing problem into a constrained function-approximation problem \cite{PINN, PINNa}. On the other side, deep BSDE/FBSDE approaches exploit the probabilistic representation directly: they simulate forward paths and train a network to learn the backward components (value and gradients) so that the terminal payoff is matched in expectation, which is particularly attractive in higher dimensions where grid-based PDE methods break down because of the curse of dimensionality \cite{we2017deep, beck2018solving, han2018solving}. Finance-focused variants of these neural PDE solvers extend this line of work by adapting network architectures, sampling schemes, and training objectives to derivative payoffs and standard market practices \cite{ML_SDE_OP, raj-paper-1, raj-paper-2, raj-paper-3}. These approaches have been successfully applied in practice to price a wide range of financial derivatives \cite{finn_1, finn_2, finn_3, finn_4, finn_5}. However, these models often operate as black boxes, making it difficult to ensure consistency with fundamental economic and financial principles, such as no-arbitrage, and to provide interpretable risk metrics. Their accuracy is also highly sensitive to data coverage and can deteriorate under regime shifts or rare market events.

\subsection{Hybrid and Theory-Embedded Approaches: The Gap}
Recognizing these limitations, recent work has explored hybrid approaches that incorporate elements of financial theory into ML models, for example by constraining outputs or augmenting loss functions. For example, \citet{handal2024kanop} proposes KANOP, a data-efficient extension of least-squares Monte Carlo for American-style options that replaces fixed basis functions with Kolmogorov--Arnold networks, improving both pricing and delta estimation under simulation scenarios. \citet{liu2024kolmogorov_arnold_finance_informed_nn} studies KAFIN, which combines the Kolmogorov--Arnold representation with a PINN-style composite loss for European option pricing in synthetically generated settings, and reports improved accuracy and training stability. \citet{dhiman2023physics_informed_option_pricing} applies PINNs as PDE solvers for both European and American put option pricing, using the Black--Scholes operator with boundary/terminal conditions and benchmarking against market prices. Finally, \citet{pinn2025} develops an uncertainty-aware PINN for the Black--Scholes equation that incorporates an ensemble-based fine-tuning stage to produce prediction bands alongside point estimates.

Yet, to our knowledge, no existing approach achieves all three of the following:
\begin{enumerate}
    \item \textbf{Theoretical consistency:} strict adherence to arbitrage-free pricing relationships;
    \item \textbf{Robustness:} stable performance with noisy, sparse, or out-of-distribution data;
    \item \textbf{Practical hedgeability:} accurate, consistent Greeks for effective trading and risk management.
\end{enumerate}
This gap motivates our \emph{Finance-Informed Neural Network (FINN)}, which embeds no-arbitrage and dynamic hedging principles directly into the network’s training objective, enabling models that are data-adaptive, robust, and theoretically sound.

\section{Methodology: The FINN Framework}
\label{sec:Methodology}
This section formalizes the \emph{Finance-Informed Neural Network (FINN)} and the economic principle that drives its learning signal. The key idea is to replace supervision by observed option prices with a replication-consistency objective: FINN is trained by minimizing the hedging (replication) error of a dynamically managed portfolio. We denote by $g^\theta$ the option pricing function parameterized by a neural network with parameters $\theta$, mapping market states to option values. In the population limit, the resulting objective is equivalent to enforcing the standard no-arbitrage condition underlying arbitrage-free pricing. Our theoretical results are therefore framed in terms of identification and consistency: (i) the replication residual induced by $g^\theta$ reflects the standard no-arbitrage pricing principle, and (ii) under standard approximation and stability conditions, FINN can approximate the arbitrage-free price and its sensitivities arbitrarily well on compact domains.

\subsection{The Arbitrage-Free Pricing Function}

\begin{definition}[Market Environment]
Let $(\Omega,\mathcal{F},(\mathcal{F}_t)_{t\in[0,T]},\mathbb{P})$ be a filtered probability space satisfying the usual conditions. We consider a market with
\begin{itemize}
  \item a risk-free asset $B_t$ satisfying $dB_t = r_t B_t\,dt$,
  \item an underlying asset $S_t$, and
  \item optionally, an additional liquid, traded derivative $H_t$ that can be used as a hedging instrument.
\end{itemize}
\end{definition}

\begin{assumption}[No-Arbitrage, Regularity, and Completeness]
\label{ass:regularity_main}
The market is arbitrage-free and admits at least one equivalent martingale measure $\mathbb{Q}\sim\mathbb{P}$. Under $\mathbb{Q}$, the tradable asset vector $(S_t,H_t)$ (when $H$ is present) follows an It\^{o} diffusion driven by a multi-dimensional Brownian motion, with coefficients satisfying standard local Lipschitz and linear growth conditions.

\medskip

\noindent
\textbf{Completeness.}  
The market is dynamically complete in the sense that every square-integrable $\mathcal{F}_T$-measurable payoff can be replicated by a self-financing trading strategy in the available traded assets. Equivalently, the equivalent martingale measure $\mathbb{Q}$ is unique.

\medskip

\noindent
\textbf{Payoff regularity.}  
The contingent claim payoff $f:\mathbb{R}_+\to\mathbb{R}$ is continuous and satisfies a polynomial growth condition: there exist constants $C>0$ and $k\ge0$ such that
\[
|f(s)| \le C(1+s^k), \qquad s\in\mathbb{R}_+.
\]
\end{assumption}

\begin{proposition}[Existence and Uniqueness of the Pricing Function $g^*$]
\label{prop:existence}
Under Assumption~\ref{ass:regularity_main}, for a European claim with payoff $f(S_T)$, an arbitrage-free time-$t$ price can be written as:
\begin{equation}
g^*(t,S_t \mid H_t)=\mathbb{E}^{\mathbb{Q}}\!\left[e^{-\int_t^T r_s ds}\,f(S_T)\mid\mathcal{F}_t\right].
\end{equation}
If, in addition, the pricing equation admits a unique sufficiently smooth solution on a compact domain $D\subset[0,T)\times\mathbb{R}_+$, then $g^*\in C^{1,2}(D)$ and the Delta and Gamma, $\partial_s g^*$ and $\partial_s^2 g^*$, are well-defined and continuous on $D$.
\end{proposition}

\textit{Remark.} Proposition ~\ref{prop:existence} is essential for the theoretical validity of FINN. It ensures that the target function we aim to learn is not simply a statistical artifact, but a well-defined mathematical object with continuous derivatives. This guarantees that the ``Greeks'' (Delta and Gamma) extracted from the neural network via automatic differentiation converge to the true economic sensitivities. The formal proof, which relies on the Feynman-Kac representation and standard parabolic regularity theory, is provided in Appendix ~\ref{app:proofs}.

\begin{remark}[Incomplete markets and learned pricing]
The completeness assumption ensures uniqueness of the arbitrage-free price. In incomplete markets, where multiple equivalent martingale measures exist, the pricing function is no longer unique and only an arbitrage-free price interval can be identified. In such settings, FINN should be interpreted as selecting a particular economically admissible pricing functional, determined by its self-financing replication error minimization objective, rather than as recovering a unique theoretical price.
\end{remark}

\subsection{FINN Parameterization and Replication-Based Self-Supervision}
\label{FINN-Parameterization_Self-Supervision}

\begin{definition}[FINN Parameterization]
FINN parameterizes the pricing function by a neural network $g^\theta(t,s)$ with parameters $\theta\in\Theta$, where $\Theta\subset\mathbb{R}^d$ is compact. We assume the activation functions are sufficiently smooth so that $\Delta_{g^\theta}:=\partial_s g^\theta$ and $\Gamma_{g^\theta}:=\partial_s^2 g^\theta$ exist and can be computed via automatic differentiation.
\end{definition}

\begin{remark}[Self-Supervision via Replication Consistency]
FINN does not require labeled training pairs $(s,g^*(t,s))$. Instead, it learns $g^\theta$ by enforcing a no-arbitrage replication criterion: a locally riskless, self-financing hedged portfolio should earn the risk-free rate. This aligns model training with the economic logic used in hedging and model validation.
\end{remark}

\begin{remark}[Delta-only versus delta--gamma training.]
FINN can be trained using delta hedging with the underlying $S_t$ alone. When a second traded derivative $H_t$ is available, FINN can additionally enforce gamma neutrality. We present the delta--gamma construction below; the delta-only case is recovered by setting $\eta_t\equiv 0$.
\end{remark}

\begin{assumption}[Nondegeneracy of the Hedging Instrument]
\label{ass:gamma_nonzero}
When delta--gamma hedging is used, the chosen hedging instrument satisfies $\Gamma_H(t,s)\neq 0$ on the training domain, where $\Delta_H$ and $\Gamma_H$ denote its Delta and Gamma with respect to $S$.
\end{assumption}

\begin{definition}[Delta--Gamma Hedged Portfolio]
\label{def:portfolio}
Consider the portfolio
\begin{equation}
\Pi_t=-g^\theta(t,S_t)+\alpha_t S_t+\eta_t H_t+\beta_t B_t.
\end{equation}
The portfolio Delta and Gamma with respect to $S_t$ are
\begin{equation}
\Delta_{\Pi_t} = \frac{\partial \Pi_t}{\partial S_t} = -\Delta_{g^\theta}(t,S_t)+\alpha_t+\eta_t\Delta_H(t,S_t),
\end{equation}
\begin{equation}
\Gamma_{\Pi_t}=\frac{\partial \Delta_{\Pi_t}}{\partial S_t}=-\Gamma_{g^\theta}(t,S_t)+\eta_t\Gamma_H(t,S_t).
\end{equation}
Imposing local delta and gamma neutrality, $\Delta_{\Pi_t}=0$ and $\Gamma_{\Pi_t}=0$, yields the hedge ratios
\begin{equation}
\eta_t=\frac{\Gamma_{g^\theta}(t,S_t)}{\Gamma_H(t,S_t)},\qquad
\alpha_t=\Delta_{g^\theta}(t,S_t)-\eta_t\,\Delta_H(t,S_t),
\end{equation}
which are well-defined under Assumption~\ref{ass:gamma_nonzero}. The control processes $\alpha_t$ and $\eta_t$ are computed entirely from the model’s own output derivatives, namely the estimated sensitivities $\Delta_{g^\theta}$ and $\Gamma_{g^\theta}$, and do not require knowledge of the true (and unobservable) option Greeks $\Delta_{g^*}$ and $\Gamma_{g^*}$. The sensitivities of the hedging instruments, $\Delta_H$ and $\Gamma_H$, are treated as known inputs, since they are obtained from standard market models (such as Black--Scholes or Heston) using observable market data.
\end{definition}

\begin{definition}[Replication-Error Loss (Discrete-Time Form)]
\label{def:loss}
Let $0=t_0<t_1<\cdots<t_N=T$ be a trading grid and denote $\Delta t_n=t_{n+1}-t_n$. For a self-financing strategy using hedge ratios computed from $g^\theta$ at time $t_n$, define the portfolio value $\Pi_{t_n}$ as in Definition~\ref{def:portfolio}. FINN trains $\theta$ by minimizing the mean squared deviation from risk-free growth:
\begin{equation}
\mathcal{L}(\theta)=
\mathbb{E}\left[\sum_{n=0}^{N-1}\left(\Pi_{t_{n+1}}-\exp\!\left(\int_{t_n}^{t_{n+1}} r_s ds\right)\Pi_{t_n}\right)^2\right].
\label{eq:loss_discrete}
\end{equation}
The expectation is taken under the data-generating measure (e.g., simulated paths under $\mathbb{Q}$ or empirical paths). This objective is self-supervised: it depends only on realizations of $S_t$, optionally $H_t$ in the case of Delta-Gamma hedging, the payoff specification, and the network output and derivatives.
\end{definition}

\begin{remark}[Geometric Interpretation]
The loss in \eqref{eq:loss_discrete} can be interpreted as an $L^2$ projection error: it measures the distance between the realized incremental portfolio value generated by $g^\theta$ and the risk-free target increment. Minimizing $\mathcal{L}(\theta)$ selects the pricing function whose induced hedging strategy best replicates the claim in mean-square sense using the available traded instruments.
\end{remark}

\subsection{Identification and Consistency}

\begin{proposition}[Replication Residual and the No-Arbitrage Pricing Operator]
\label{prop:residual_pde}
Consider the continuous-time limit of the replication residual underlying \eqref{eq:loss_discrete}. Under the regularity conditions of Assumption~\ref{ass:regularity_main}, the following equivalence holds on any compact domain $D\subset[0,T)\times\mathbb{R}_+$:

\begin{equation}
\begin{aligned}
&\text{(replication residual vanishes)}\\
&\iff\;
\text{$g^\theta$ satisfies the corresponding}\\
&\hspace{1.6em}\text{no-arbitrage pricing equation on $D$.}
\end{aligned}
\end{equation}

In particular, in diffusion settings, the vanishing of the residual yields the standard pricing PDE (or, more generally, the appropriate pricing operator associated with the underlying dynamics). We refer the reader to Appendix \ref{app:proofs} for the formal proof, which establishes this equivalence using standard stochastic calculus arguments.
\end{proposition}

\begin{theorem}[Consistency of Price and Hedging Sensitivities]
\label{thm:consistency}
Let $D \subset [0,T) \times \mathbb{R}_+$ be a compact domain. Suppose the arbitrage-free pricing equation admits a unique classical solution $g^* \in C^{1,2}(D)$, as established in Proposition \ref{prop:existence}. Assume the neural network architecture is sufficiently rich to approximate functions in $C^{1,2}(D)$.

\noindent (i) Uniform Approximation Capability.
For any precision $\varepsilon > 0$, there exists a network parameterization $\theta_\varepsilon \in \Theta$ such that the model approximates the true price and its hedging sensitivities (Delta and Gamma) uniformly on $D$:
\begin{align}
\sup_{(t,s) \in D} \big| g^{\theta_\varepsilon}(t,s) - g^*(t,s) \big| &< \varepsilon, \\
\sup_{(t,s) \in D} \big| \partial_s g^{\theta_\varepsilon}(t,s) - \partial_s g^*(t,s) \big| &< \varepsilon, \\
\sup_{(t,s) \in D} \big| \partial_s^2 g^{\theta_\varepsilon}(t,s) - \partial_s^2 g^*(t,s) \big| &< \varepsilon.
\end{align}

\noindent (ii) Convergence via Risk Minimization. 
Let $\{\theta_m\}_{m \in \mathbb{N}}$ be a sequence of trained parameters such that the replication loss $\mathcal{L}(\theta_m) \to 0$ as the discretization grid refines ($\Delta t \to 0$). Then, under standard parabolic stability conditions, the learned function sequence $\{g^{\theta_m}\}$ converges to the unique no-arbitrage solution $g^*$, implying simultaneous convergence of the implied hedging strategy to the optimal hedge.
\end{theorem}

\subsection{Special Case of the FINN Framework: Recovering the Black--Scholes PDE}

To make the link between replication-based learning and the classical pricing equation explicit, we provide a concrete derivation in the Black--Scholes setting. This result is standard, but it clarifies how FINN's replication objective recovers the Black--Scholes operator as a special case.

\begin{proposition}[Equivalence of Replication Consistency and the Black--Scholes Operator]
\label{prop:bs_equiv}
Let $S_t$ follow a geometric Brownian motion under the risk-neutral measure $\mathbb{Q}$,
\begin{equation}
dS_t=rS_t\,dt+\sigma S_t\,dW_t,
\end{equation}
and let $B_t$ satisfy $dB_t=rB_t\,dt$. Consider the delta-hedged portfolio $\Pi_t=-g^\theta(t,S_t)+\alpha_t S_t$ with $\alpha_t=\partial_s g^\theta(t,S_t)$. Then, in continuous time,
\begingroup
\small
\begin{equation}
\begin{aligned}
d\Pi_t = r\Pi_t\,dt
\;\iff\;
&\partial_t g^\theta
+ r s\,\partial_s g^\theta
+ \tfrac{1}{2}\sigma^2 s^2\,\partial_s^2 g^\theta \\
&\qquad - r g^\theta = 0 .
\end{aligned}
\end{equation}
\endgroup
Equivalently, the replication-consistency condition corresponds exactly to the Black--Scholes PDE.
\end{proposition}

\noindent The derivation is provided in Appendix \ref{app:proofs}. By strictly enforcing the self-financing condition of the hedging portfolio, we demonstrate that the only way for the portfolio to grow at the risk-free rate is if the neural network's pricing surface solves the Black-Scholes PDE. This confirms that our ``replication loss'' is not just a statistical objective, but a direct implementation of the no-arbitrage condition.

\subsection{Extension to American Options}
\label{sec:american_sketch}

American options introduce an early exercise feature, turning pricing into an optimal stopping problem. FINN can be extended by combining (i) replication consistency in the continuation region with (ii) enforcement of the value constraint $g \ge f$, where $g(t,s)$ denotes the option’s continuation (hold) value and $f(s)$ its intrinsic (immediate exercise) payoff. The inequality $g(t,s) > f(s)$ therefore characterizes the continuation region in which holding the option is optimal, while equality identifies the exercise boundary.

Concretely, one may parameterize $g^\theta(t,s)$ and train using a composite objective that (a) penalizes replication error only when the estimated continuation value exceeds the intrinsic value and (b) penalizes violations of the lower bound:

\begingroup
\scriptsize
\begin{align}
\mathcal{L}_A(\theta)
&=
\mathbb{E}\!\left[
\sum_{n}
\left(
\Pi_{t_{n+1}}
-
\exp\!\left(\int_{t_n}^{t_{n+1}} r_s\,ds\right)
\Pi_{t_n}
\right)^2
\mathbf{1}_{\{g^\theta(t_n,S_{t_n})>f(S_{t_n})\}}
\right] \notag\\
&\quad
+\lambda\,
\mathbb{E}\!\left[
\left(
\max\{f(S_{t_n})-g^\theta(t_n,S_{t_n}),0\}
\right)^2
\right],
\label{eq:american_loss}
\end{align}
\endgroup
with $\lambda>0$.
Under standard regularity and stability conditions for variational inequalities, the exercise boundary emerges endogenously as the set where $g^\theta(t,s)=f(s)$. Further technical details and supporting arguments are provided in Section \ref{sec:american_option_hedging}.

This extension maintains the core principles of FINN while adapting to the free boundary nature of American options, opening avenues for self-supervised pricing in more complex, path-dependent derivatives.

\section{Numerical Study}\label{sec:results}

This section provides a comprehensive empirical evaluation of the FINN framework, with the objective of assessing its accuracy, economic consistency, and practical relevance across progressively more challenging settings. We begin with controlled experiments under classical models, where analytical or high-precision numerical benchmarks are available, to verify that FINN recovers arbitrage-based prices and hedge ratios when standard assumptions hold. We then examine whether core no-arbitrage relationships emerge endogenously, and evaluate FINN’s ability to support advanced risk-management tasks such as delta--gamma hedging. Finally, building on these controlled validations, we study robustness under volatility regime shifts and transaction costs, and compare FINN to parametric alternatives in the reconstruction of implied volatility surfaces. Together, these experiments are designed to establish FINN as a theory-consistent, data-driven pricing operator that generalizes across models, market conditions, and data availability regimes.

\subsection{Experimental Design}
\label{sec:design}

\begin{figure}[h]
    \centering
    \includegraphics[width=\linewidth]{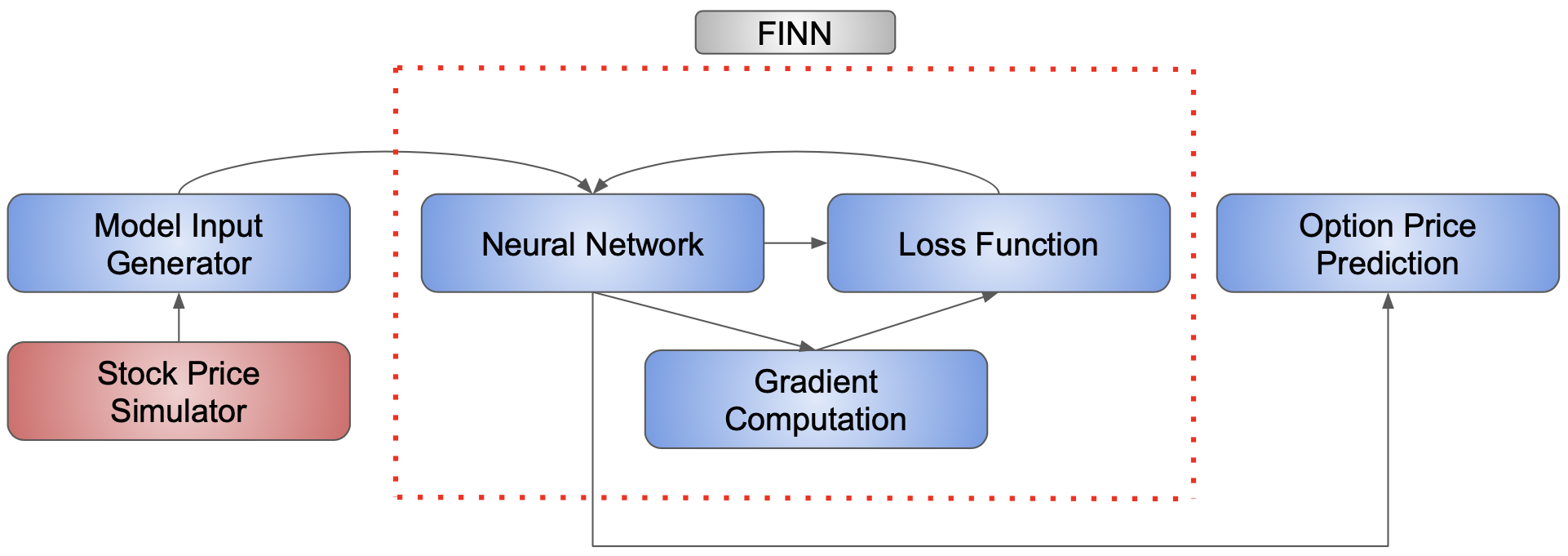}
    \caption{Overview of the FINN training and evaluation pipeline.}
    \label{fig:finn_system}
\end{figure}

We begin by describing the experimental design used to evaluate FINN. The experiments are constructed to assess four central claims of the framework: (i) recovery of classical pricing models under standard assumptions, (ii) robustness in environments with stochastic volatility, (iii) endogenous adherence to no-arbitrage conditions, and (iv) decision-relevant performance in hedging and risk management. An overview of the training and evaluation pipeline is provided in Figure~\ref{fig:finn_system}.

\subsubsection{Data Generation and Market Environments}
Our first batch of experiments focuses on synthetic data generated from controlled stochastic models of asset prices. This design choice allows us to clearly assess the model’s behavior and benchmark FINN against known ground truth. We consider two canonical market environments.

First, we use a Geometric Brownian Motion (GBM) model with constant volatility, corresponding to the classical Black--Scholes setting. This environment serves as a baseline in which the arbitrage-free price is known analytically. Second, we consider the Heston stochastic volatility model, which introduces time-varying and mean-reverting volatility and does not admit closed-form solutions for most option payoffs. This environment is used to assess robustness under more realistic and challenging dynamics. Full model specifications are provided in the online Appendix~\ref{appendix:GBM} and~\ref{appendix:Heston}.

In addition to these controlled synthetic settings, we also evaluate FINN on real-world market data through empirical experiments and case studies, which are used to demonstrate robustness and practical relevance in applied pricing and hedging scenarios.

\begin{figure}[h]
    \centering
    \includegraphics[width=\linewidth]{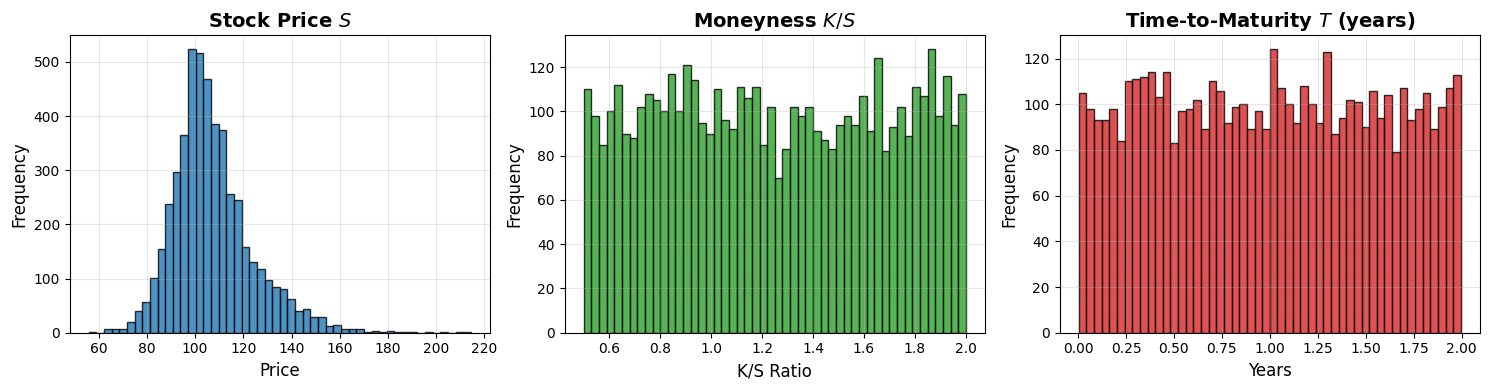}
    \caption{Distribution of underlying prices, moneyness, and time to maturity in the training data (illustrated on a subset of 5{,}000 samples).}
    \label{fig:coverage}
\end{figure}

Simulated price paths are segmented into rolling time windows. For each window, we randomly sample strike prices and times to maturity, ensuring broad coverage across moneyness and maturity dimensions. Figure~\ref{fig:coverage} illustrates the resulting distribution of training samples. This procedure is designed to prevent concentration in narrow regions of the state space and to evaluate generalization across option characteristics.

\subsubsection{Model Architecture and Training}
FINN is implemented as a fully connected feedforward neural network with two hidden layers of 50 units each. Smooth activation functions are used to ensure stable computation of sensitivities via automatic differentiation. The network is trained using the self-supervised dynamic hedging loss derived in equation ~\ref{eq:loss_discrete} (Section~\ref{sec:Methodology}), which penalizes deviations from replication consistency rather than discrepancies from labeled prices.

In the numerical experiments, the continuous-time loss is discretized along simulated paths. We work with discounted portfolio values and set the risk-free rate to zero, so the loss in \eqref{eq:loss_discrete} reduces to the mean squared increment of the hedged portfolio value. We denote by $\Pi_{t_n}$ the portfolio value at the $n$-th trading time $t_n$; for notational simplicity, we occasionally write $\Pi_n := \Pi_{t_n}$, and 
\(
\mathbb{E}\!\left[\sum_{n=0}^{N-1}
\left(\Pi_{n+1} - e^{r_n \Delta t_n}\,\Pi_n\right)^2
\right].
\)
Training is performed for a fixed number of epochs with early stopping based on a held-out validation set to mitigate overfitting. All reported results are averaged over ten independent runs with different random seeds.

\subsubsection{Evaluation Protocol and Benchmarks}
Model performance is evaluated on independently generated test paths drawn from the same data-generating processes. Option parameters are systematically varied to examine sensitivity with respect to moneyness, maturity, and volatility regimes. In the geometric Brownian motion (GBM) environment, FINN’s predictions are benchmarked against the analytical Black--Scholes prices and Greeks. In the Heston environment, we use Monte Carlo estimates with a large number of paths as the reference benchmark. Performance for both option prices and hedge ratios is evaluated using relative deviation (RMAD) and squared-error (RMSE) measures, with RMAD and RMSE reported in the main text and NMAD and MSE provided in the Appendix\footnote{The full description of all performance measures used in this section is provided in ~\ref{sec:measures}. The NMAD normalizes the mean absolute deviation by the empirical range of the benchmark values, while MSE reports errors in squared units; reporting both provides complementary views of magnitude and scale sensitivity.}.

Our experimental design here emphasizes not only the pricing accuracy, but also the stability of hedge ratios and its economic consistency, which are critical for downstream trading and risk management decisions. To ensure statistical robustness, all reported results are averaged over multiple independent runs with different random seeds. Beyond pricing accuracy, we also examine the stability and economic consistency of the learned hedge ratios, since these properties are critical for downstream hedging and risk-management applications.

\subsection{Baseline Validation: Black--Scholes and Stochastic Volatility}

We begin by validating FINN in controlled environments where benchmark solutions are available. The objective of this section is to verify that FINN recovers classical arbitrage-based prices and hedge ratios when model assumptions hold.

\subsubsection{Consistency with the Black--Scholes Model}
A minimal requirement for any learning-based pricing framework is consistency with the Black--Scholes model under Geometric Brownian Motion (GBM). We therefore evaluate FINN on European call options generated under GBM, spanning a wide range of underlying prices, strikes, maturities (60–120 days), and volatilities (12.5\%–17.5\%). Analytical Black--Scholes prices and deltas serve as the ground truth.

\begin{table}[t]
\centering
{\begin{tabular}{cc|cc|cc|cc|cc}
\hline
\multirow{2}{*}{Vol $(\sigma)$} & \multirow{2}{*}{TTM $(\tau)$} &
\multicolumn{4}{c|}{Call Option Price $(C)$} &
\multicolumn{4}{c}{Hedge Ratio $(\Delta)$} \\
\cline{3-10}
 & & \multicolumn{2}{c}{RMAD} & \multicolumn{2}{c|}{RMSE} &
     \multicolumn{2}{c}{RMAD} & \multicolumn{2}{c}{RMSE} \\
\hline
\multirow{7}{*}{0.125}
 & 0.24 & 0.022 & (0.005) & 0.020 & (0.004) & 0.041 & (0.009) & 0.044 & (0.009) \\
 & 0.28 & 0.024 & (0.005) & 0.021 & (0.005) & 0.043 & (0.009) & 0.044 & (0.009) \\
 & 0.32 & 0.025 & (0.006) & 0.022 & (0.005) & 0.044 & (0.009) & 0.045 & (0.009) \\
 & 0.36 & 0.026 & (0.006) & 0.022 & (0.005) & 0.045 & (0.010) & 0.045 & (0.009) \\
 & 0.40 & 0.026 & (0.006) & 0.023 & (0.006) & 0.045 & (0.010) & 0.044 & (0.009) \\
 & 0.44 & 0.027 & (0.007) & 0.023 & (0.006) & 0.044 & (0.010) & 0.043 & (0.010) \\
 & 0.48 & 0.026 & (0.007) & 0.022 & (0.006) & 0.043 & (0.010) & 0.042 & (0.010) \\
\hline
\multirow{7}{*}{0.150}
 & 0.24 & 0.029 & (0.006) & 0.025 & (0.006) & 0.050 & (0.011) & 0.049 & (0.010) \\
 & 0.28 & 0.031 & (0.007) & 0.027 & (0.006) & 0.051 & (0.011) & 0.050 & (0.011) \\
 & 0.32 & 0.033 & (0.007) & 0.028 & (0.006) & 0.052 & (0.011) & 0.050 & (0.011) \\
 & 0.36 & 0.033 & (0.008) & 0.028 & (0.007) & 0.052 & (0.012) & 0.050 & (0.011) \\
 & 0.40 & 0.034 & (0.008) & 0.028 & (0.007) & 0.052 & (0.012) & 0.049 & (0.011) \\
 & 0.44 & 0.034 & (0.009) & 0.028 & (0.007) & 0.051 & (0.012) & 0.048 & (0.011) \\
 & 0.48 & 0.033 & (0.009) & 0.028 & (0.007) & 0.049 & (0.012) & 0.046 & (0.011) \\
\hline
\multirow{7}{*}{0.175}
 & 0.24 & 0.037 & (0.009) & 0.031 & (0.007) & 0.058 & (0.012) & 0.055 & (0.011) \\
 & 0.28 & 0.039 & (0.009) & 0.033 & (0.007) & 0.059 & (0.013) & 0.056 & (0.012) \\
 & 0.32 & 0.041 & (0.010) & 0.034 & (0.008) & 0.060 & (0.014) & 0.056 & (0.012) \\
 & 0.36 & 0.042 & (0.010) & 0.034 & (0.008) & 0.060 & (0.014) & 0.055 & (0.012) \\
 & 0.40 & 0.042 & (0.011) & 0.034 & (0.008) & 0.059 & (0.014) & 0.055 & (0.012) \\
 & 0.44 & 0.042 & (0.011) & 0.034 & (0.009) & 0.058 & (0.014) & 0.053 & (0.012) \\
 & 0.48 & 0.041 & (0.011) & 0.034 & (0.009) & 0.056 & (0.014) & 0.052 & (0.012) \\
\hline
\end{tabular}}
\captionsetup{width=0.9\linewidth}
\caption{FINN performance against Black--Scholes (GBM) for European call options. Entries report mean RMAD and RMSE with standard deviations in parentheses (across 10 different training seeds).}
\label{finn_gbm}
\end{table}

\begin{figure}
{\centering
\begin{tabular}{@{}cc@{}}
\includegraphics[width=0.48\linewidth]{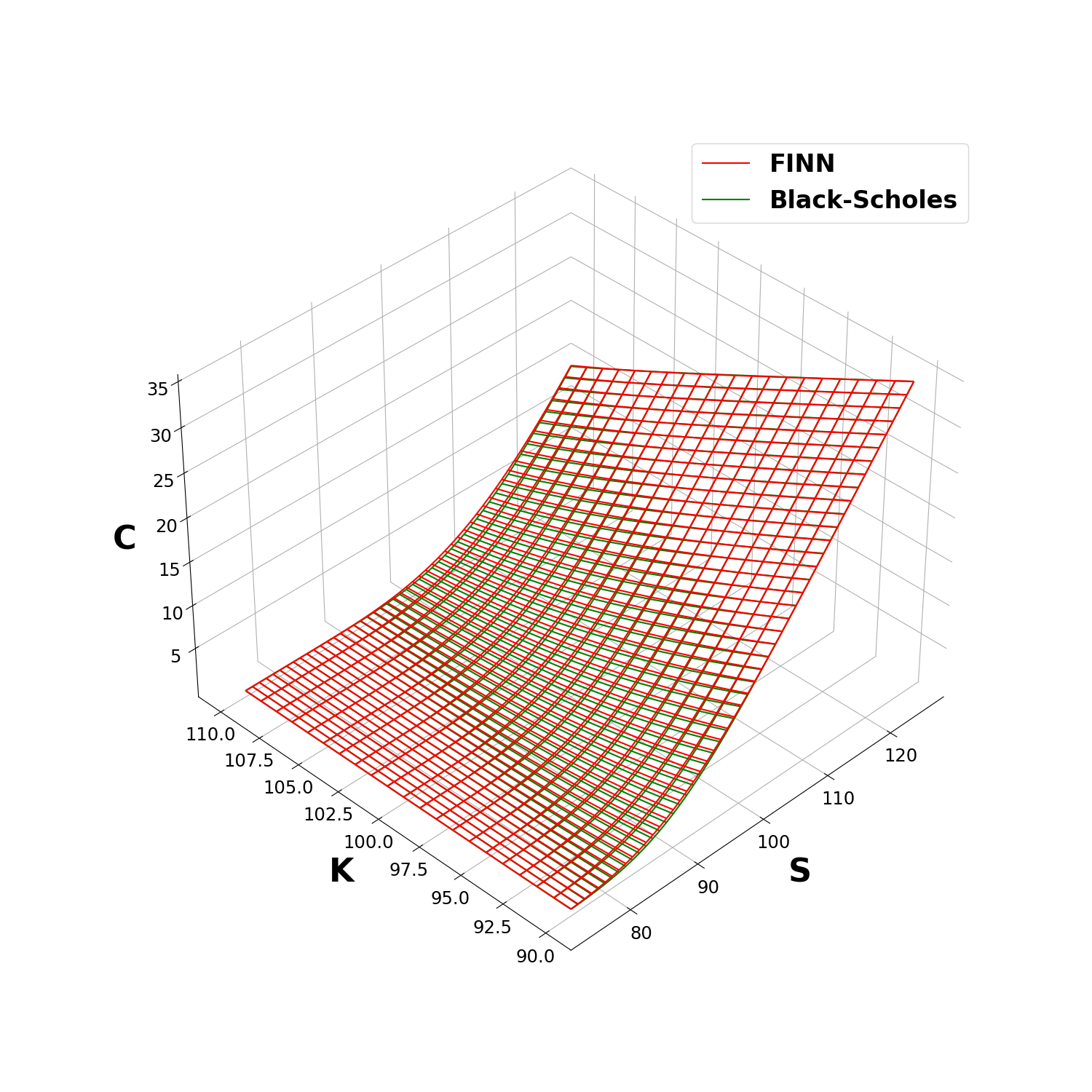} &
\includegraphics[width=0.48\linewidth]{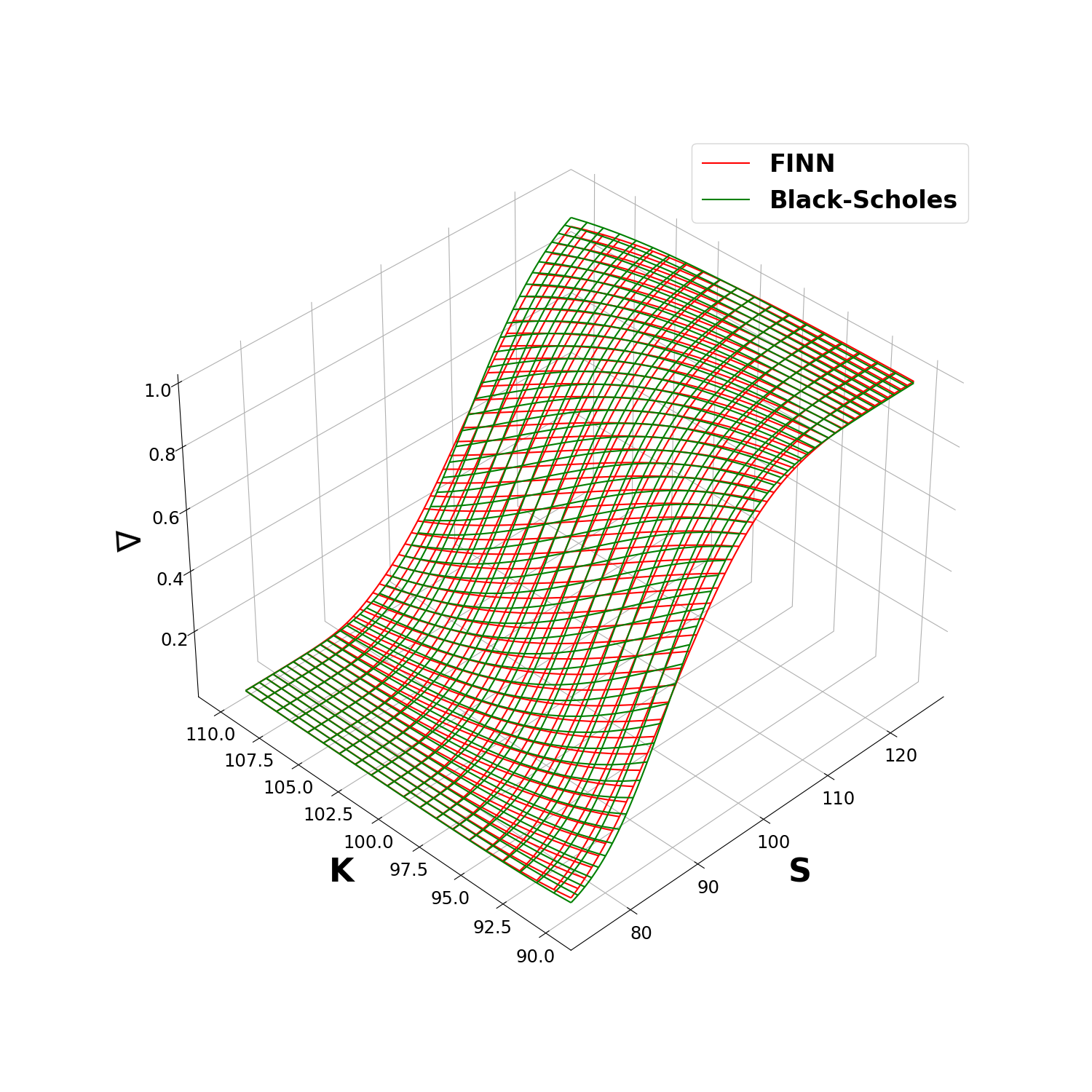} \\
{\footnotesize (a) Estimated call option price surface} &
{\footnotesize (b) Estimated hedge ratio $(\Delta)$ surface} \\
[2mm]
\includegraphics[width=0.48\linewidth]{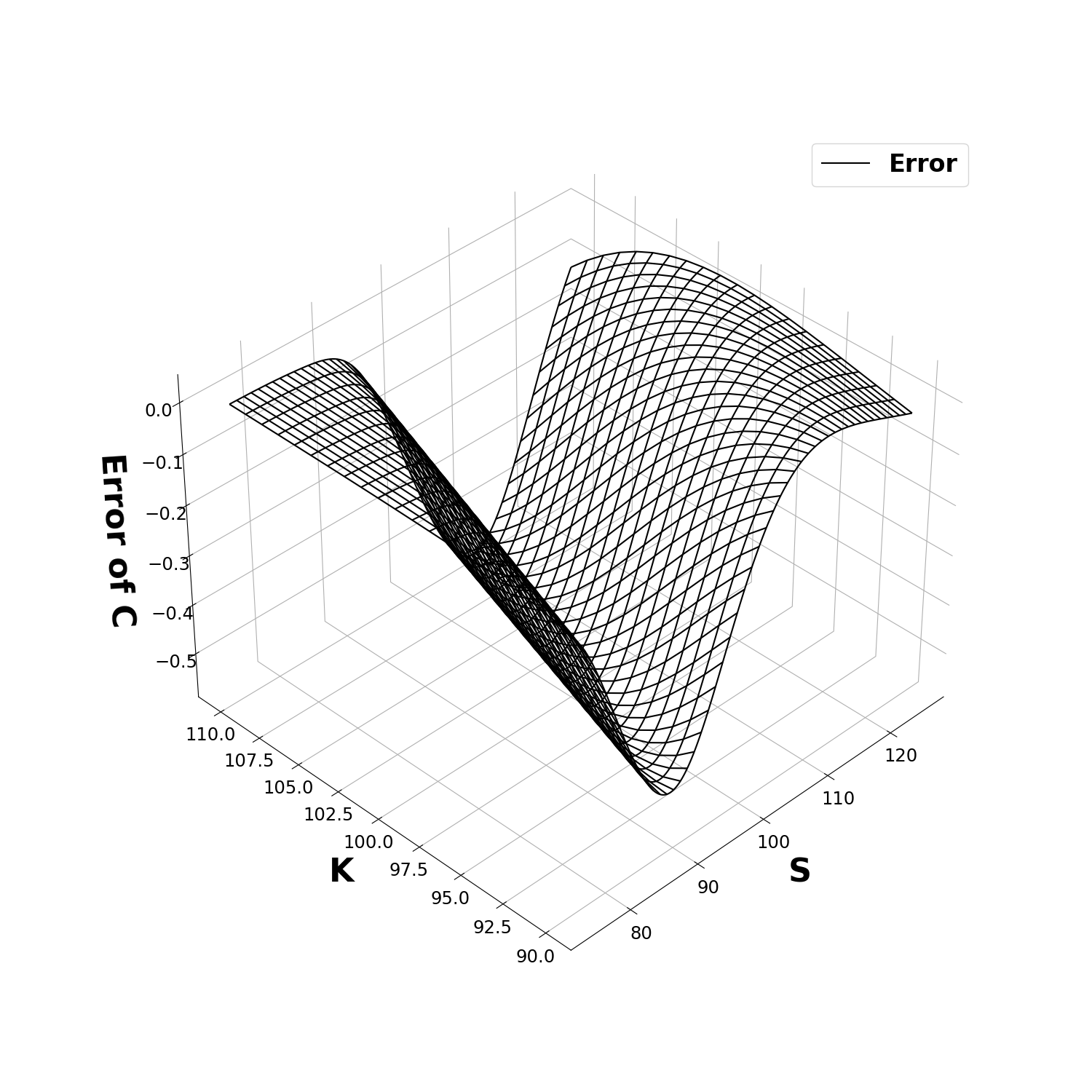} &
\includegraphics[width=0.48\linewidth]{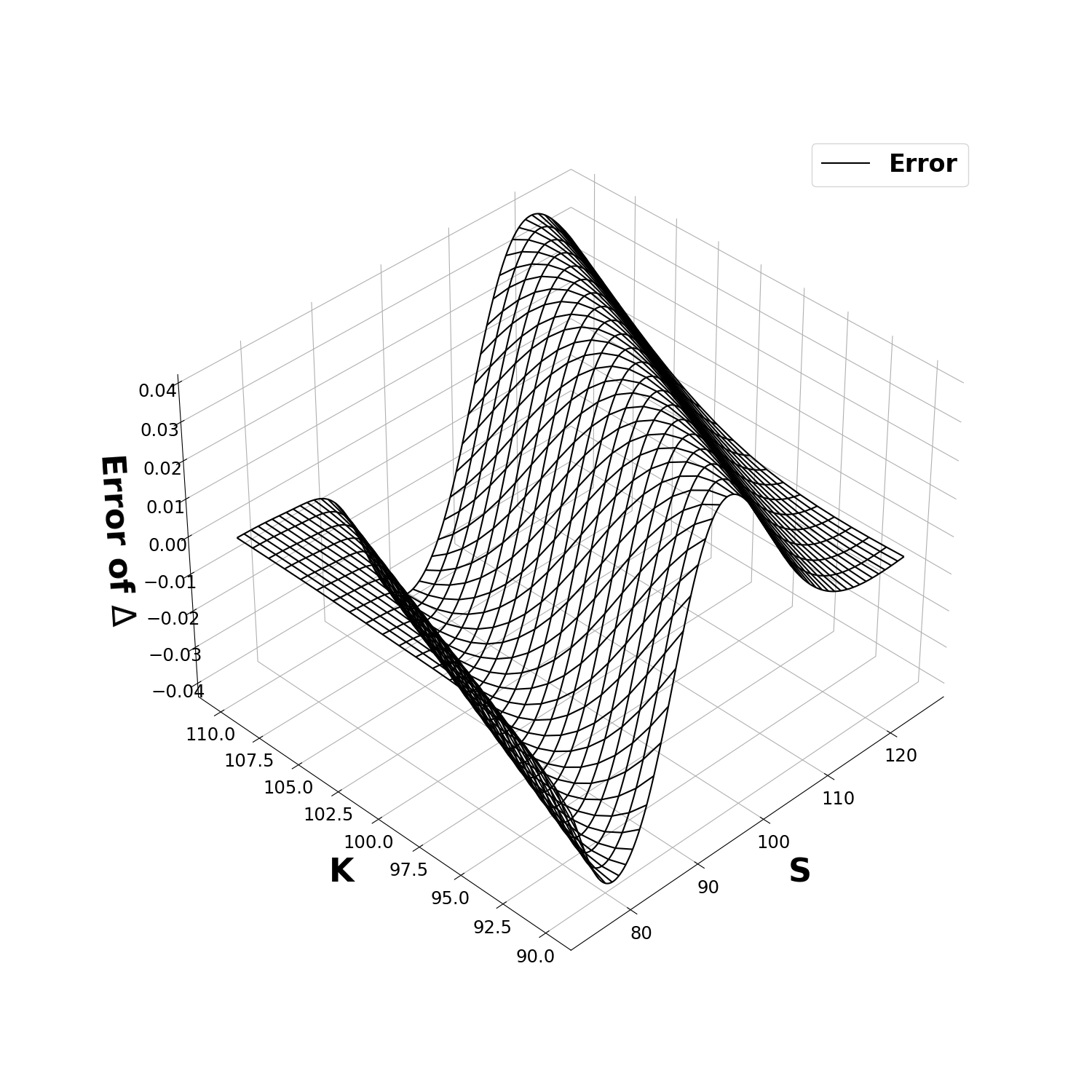} \\
{\footnotesize (c) Call price error} &
{\footnotesize (d) Hedge ratio $(\Delta)$ error} \\
\end{tabular}
}
\caption{Visual validation of FINN against the analytical Black--Scholes solution for a European call under GBM with $\sigma=0.15$ and $\tau=0.36$. Panels (a)--(b) show predicted and analytical surfaces; panels (c)--(d) show errors, concentrated near the at-the-money region.}
\label{fig:four_plots}
\end{figure}

Table~\ref{finn_gbm} summarizes pricing and hedging accuracy across volatility and maturity regimes. FINN achieves uniformly low pricing errors, with most relative mean absolute deviations well below 4\% across all configurations. Errors in hedge ratios are similarly small, with delta deviations typically in the range of 2–3\%. Standard deviations across independent runs are negligible, indicating stable training and convergence.

Consistent with financial intuition, errors tend to increase monotonically with volatility and time to maturity, reflecting greater curvature and uncertainty in the pricing function. These effects are smooth and systematic rather than random, suggesting that FINN internalizes the local geometry of the option surface rather than overfitting individual regions. Comparable accuracy is obtained for European put options (Appendix~\ref{secA2}), confirming that the self-supervised hedging objective generalizes across payoff structures.

Figure~\ref{fig:four_plots} provides a visual comparison between FINN’s predicted price and delta surfaces and their analytical Black--Scholes counterparts. Panels (a) and (b) show near-perfect alignment across the domain. Panels (c) and (d) report residual errors, which are small and concentrated around the at-the-money region where option convexity (Gamma) is highest. This localization is economically expected and indicates that FINN captures both global pricing behavior and local sensitivity structure.

\subsubsection{Robustness under Stochastic Volatility}

We next evaluate FINN in a Heston stochastic volatility environment, where volatility is time-varying and mean-reverting. In this setting, the Black–Scholes model is misspecified. While the Heston model admits semi-analytical pricing formulas, Greeks such as Delta and Gamma do not have simple closed-form expressions and are typically computed numerically. We therefore use Monte Carlo estimates as the reference solution.

\begin{table}[t]
\centering
{\begin{tabular}{cc|cc|cc|cc|cc}
\hline
\multirow{2}{*}{VolVol $(\xi)$} & \multirow{2}{*}{TTM $(\tau)$} &
\multicolumn{4}{c|}{Call Option Price $(C)$} &
\multicolumn{4}{c}{Hedge Ratio $(\Delta)$} \\
\cline{3-10}
 & & \multicolumn{2}{c}{RMAD} & \multicolumn{2}{c|}{RMSE} &
     \multicolumn{2}{c}{RMAD} & \multicolumn{2}{c}{RMSE} \\
\hline
\multirow{7}{*}{0.125}
 & 0.24 & 0.029 & (0.015) & 0.026 & (0.013) & 0.050 & (0.024) & 0.051 & (0.023) \\
 & 0.28 & 0.032 & (0.015) & 0.028 & (0.013) & 0.052 & (0.025) & 0.053 & (0.023) \\
 & 0.32 & 0.035 & (0.016) & 0.030 & (0.014) & 0.055 & (0.026) & 0.055 & (0.024) \\
 & 0.36 & 0.037 & (0.016) & 0.032 & (0.014) & 0.057 & (0.026) & 0.056 & (0.023) \\
 & 0.40 & 0.039 & (0.016) & 0.033 & (0.014) & 0.058 & (0.026) & 0.058 & (0.023) \\
 & 0.44 & 0.040 & (0.016) & 0.034 & (0.014) & 0.059 & (0.026) & 0.058 & (0.023) \\
 & 0.48 & 0.042 & (0.016) & 0.035 & (0.014) & 0.060 & (0.026) & 0.059 & (0.023) \\
\hline
\multirow{7}{*}{0.150}
 & 0.24 & 0.031 & (0.014) & 0.027 & (0.013) & 0.052 & (0.023) & 0.054 & (0.023) \\
 & 0.28 & 0.033 & (0.016) & 0.029 & (0.014) & 0.055 & (0.024) & 0.056 & (0.023) \\
 & 0.32 & 0.036 & (0.016) & 0.031 & (0.015) & 0.058 & (0.025) & 0.059 & (0.024) \\
 & 0.36 & 0.038 & (0.017) & 0.033 & (0.015) & 0.060 & (0.025) & 0.061 & (0.023) \\
 & 0.40 & 0.040 & (0.018) & 0.035 & (0.016) & 0.062 & (0.025) & 0.062 & (0.023) \\
 & 0.44 & 0.042 & (0.018) & 0.036 & (0.016) & 0.063 & (0.025) & 0.063 & (0.023) \\
 & 0.48 & 0.044 & (0.018) & 0.037 & (0.016) & 0.064 & (0.024) & 0.064 & (0.022) \\
\hline
\multirow{7}{*}{0.175}
 & 0.24 & 0.034 & (0.014) & 0.030 & (0.013) & 0.056 & (0.021) & 0.059 & (0.021) \\
 & 0.28 & 0.037 & (0.015) & 0.033 & (0.014) & 0.060 & (0.022) & 0.062 & (0.021) \\
 & 0.32 & 0.041 & (0.015) & 0.035 & (0.014) & 0.063 & (0.022) & 0.064 & (0.021) \\
 & 0.36 & 0.044 & (0.016) & 0.038 & (0.015) & 0.066 & (0.022) & 0.067 & (0.021) \\
 & 0.40 & 0.046 & (0.016) & 0.040 & (0.015) & 0.068 & (0.022) & 0.069 & (0.021) \\
 & 0.44 & 0.048 & (0.016) & 0.042 & (0.015) & 0.070 & (0.022) & 0.070 & (0.021) \\
 & 0.48 & 0.051 & (0.017) & 0.043 & (0.015) & 0.072 & (0.022) & 0.071 & (0.020) \\

\hline
\end{tabular}}
\captionsetup{width=0.9\linewidth}
\caption{FINN performance against the Heston stochastic volatility model for European call options. Entries report mean RMAD and RMSE with standard deviations in parentheses (across 10 different training seeds).}
\label{table_heston}
\end{table}

\begin{figure}
{\centering
\begin{tabular}{@{}cc@{}}
\includegraphics[width=0.48\linewidth]{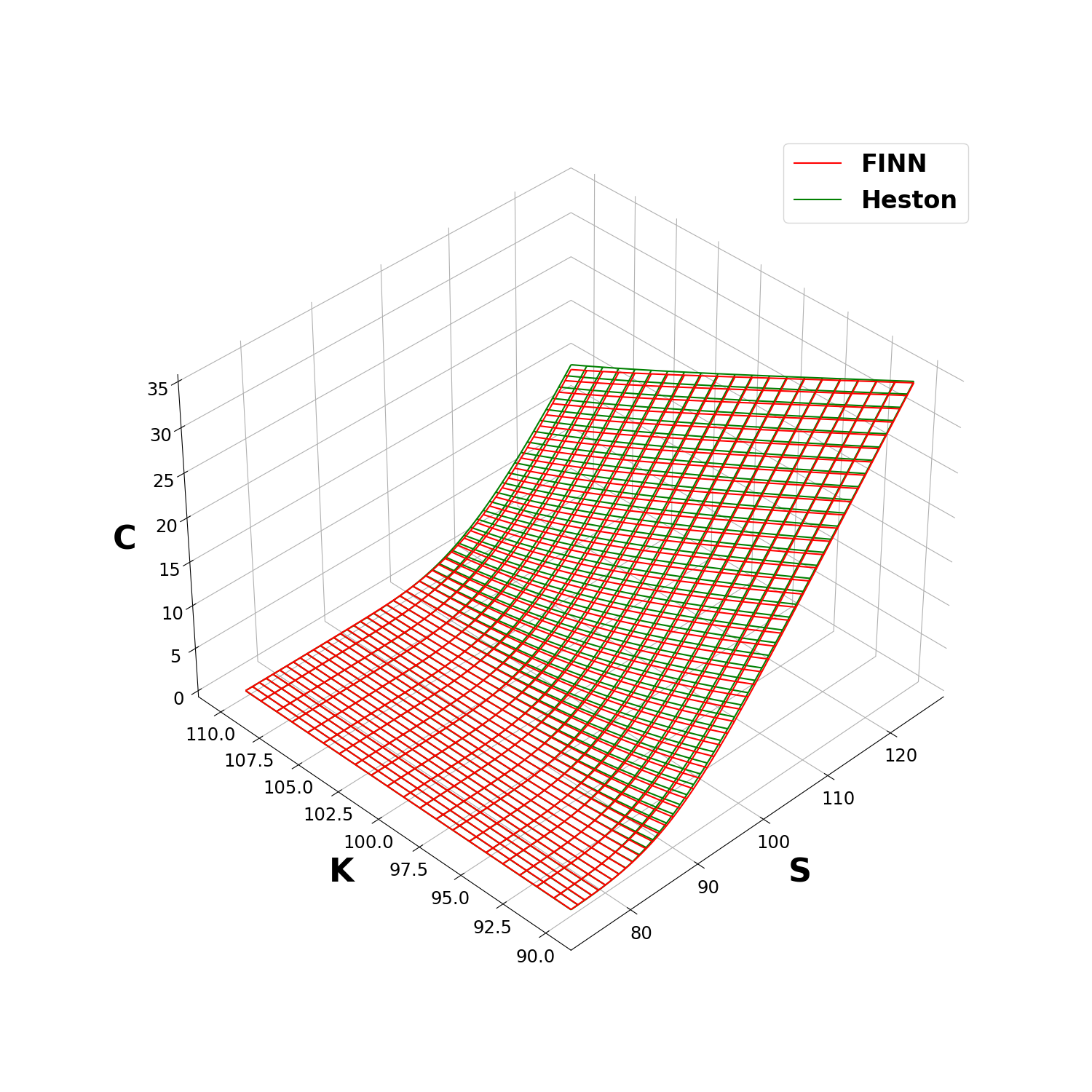} &
\includegraphics[width=0.48\linewidth]{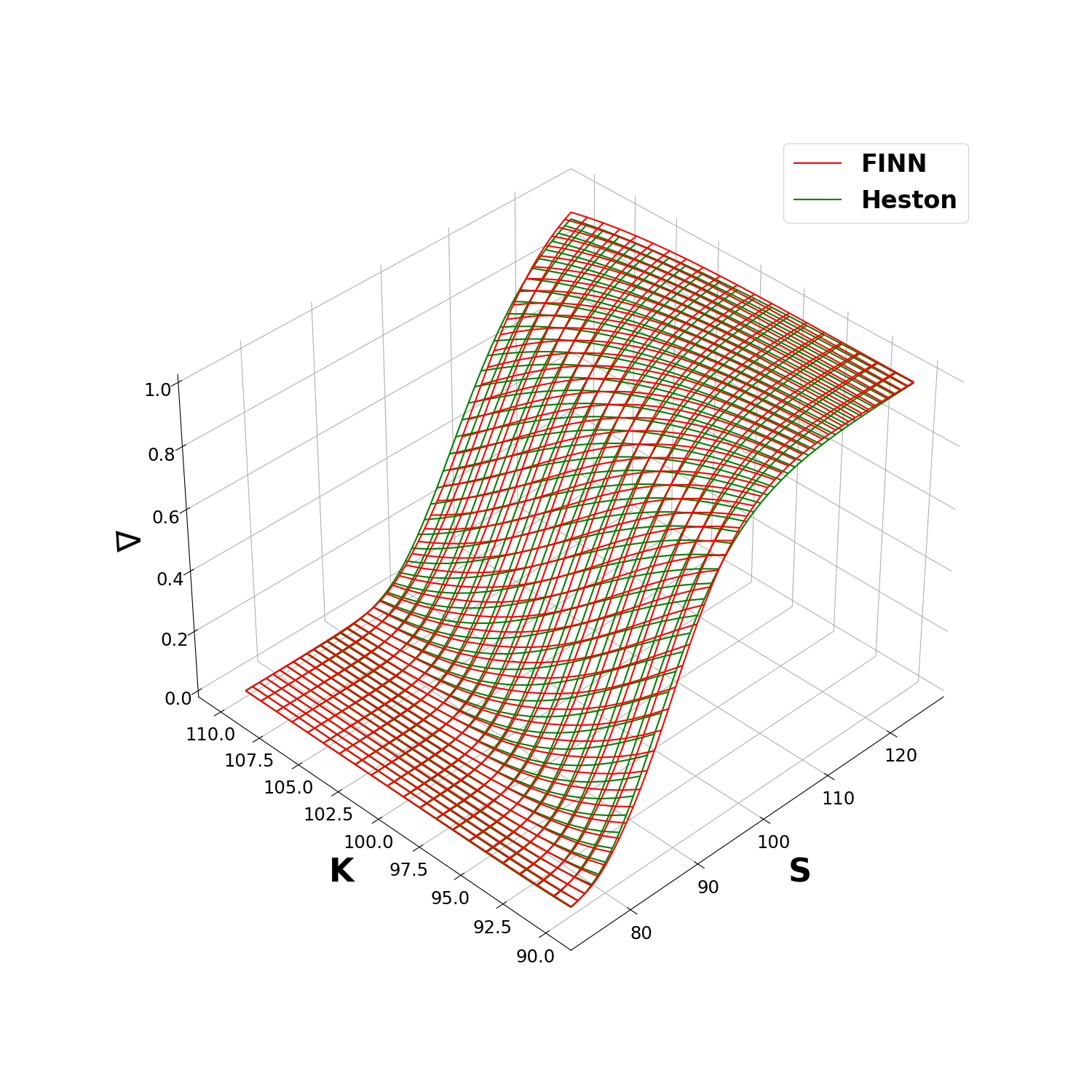} \\
{\footnotesize (a) Estimated call option price surface (Heston)} &
{\footnotesize (b) Estimated hedge ratio $(\Delta)$ surface (Heston)} \\
[2mm]
\includegraphics[width=0.48\linewidth]{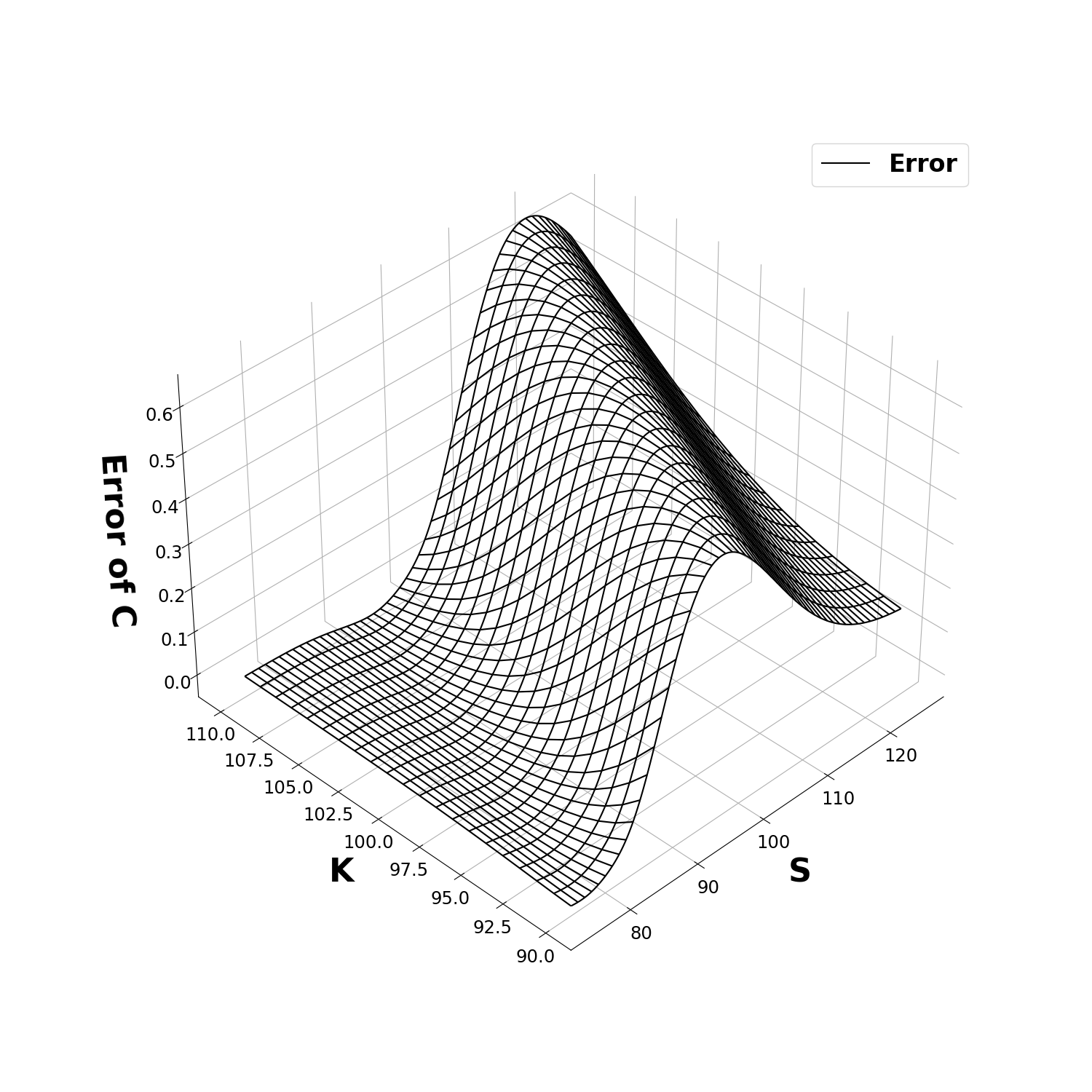} &
\includegraphics[width=0.48\linewidth]{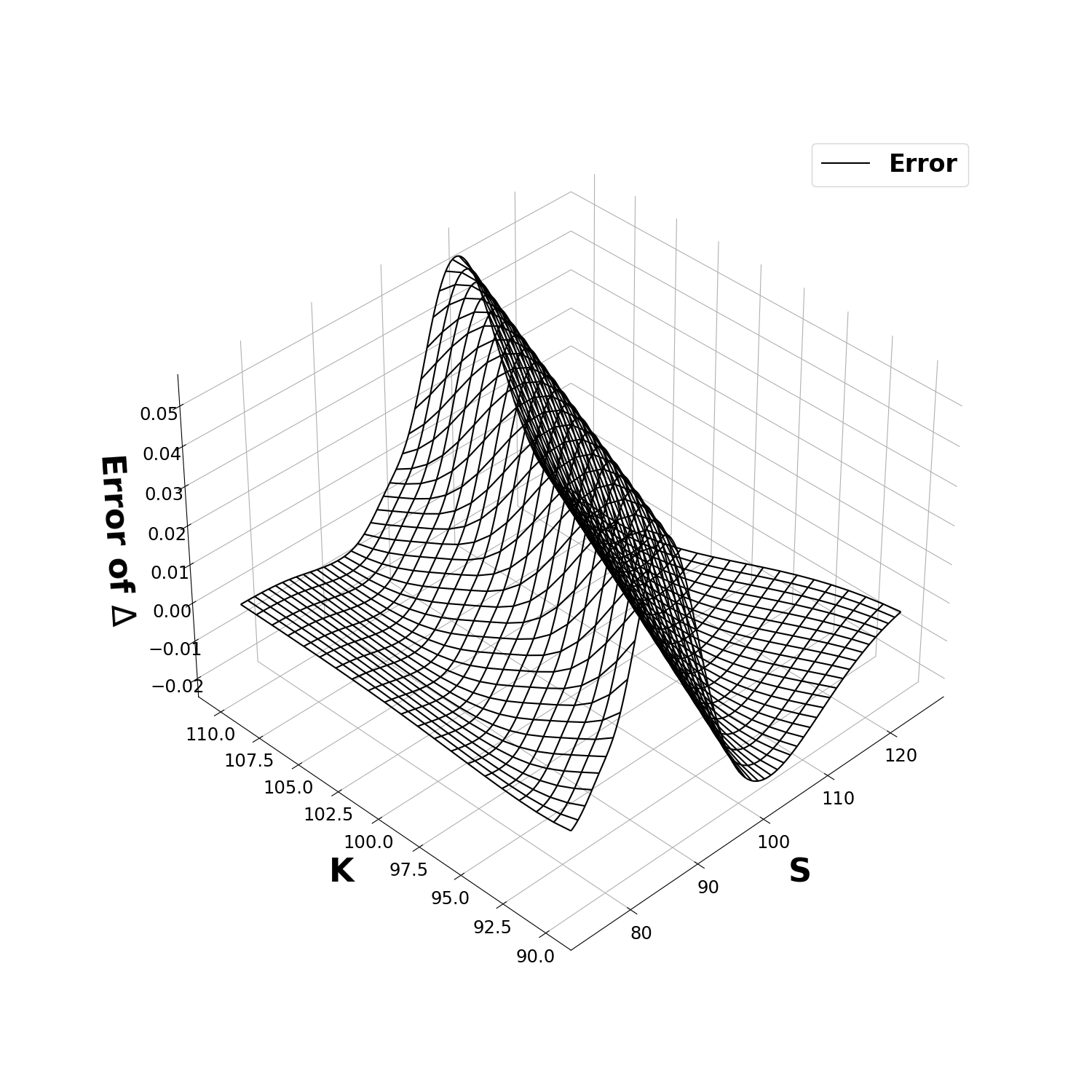} \\
{\footnotesize (c) Call price error (Heston)} &
{\footnotesize (d) Hedge ratio $(\Delta)$ error (Heston)} \\
\end{tabular}
}
\caption{Visual validation of FINN in the Heston model for a European call with volatility-of-volatility $\xi=0.15$ and $\tau=0.36$. Panels (a)--(b) show predicted and benchmark surfaces; panels (c)--(d) show errors, concentrated near the at-the-money region.}
\label{fig:heston_four_plots}
\end{figure}

Table~\ref{table_heston} reports pricing and delta errors across different volatility-of-volatility parameters and maturities. FINN maintains high accuracy despite the increased complexity of the underlying dynamics. Absolute pricing deviations remain small, corresponding to economically negligible relative errors for typical option values. Hedge ratio estimates are similarly stable, with delta errors remaining on the order of a few percent.

As expected, pricing errors increase with both volatility-of-volatility and time to maturity, reflecting greater uncertainty and nonlinearity in the pricing function. This degradation is gradual and systematic rather than abrupt, indicating that FINN adapts smoothly to deviations from constant volatility assumptions. Performance on put options exhibits the same qualitative patterns (Appendix~\ref{secA3}).

Figure~\ref{fig:heston_four_plots} visualizes FINN’s performance in the Heston setting. Predicted price and delta surfaces closely track the Monte Carlo benchmark, while residual errors are again concentrated in the at-the-money region where curvature is highest. This mirrors the behavior observed under GBM and confirms that FINN learns the relevant nonlinear structure even when volatility is stochastic.

Taken together, these baseline experiments establish that FINN (i) recovers classical no-arbitrage based pricing when model assumptions hold and (ii) remains accurate and stable when those assumptions are relaxed. This provides a foundation for evaluating no-arbitrage consistency and hedging performance in more demanding settings.

\subsection{Adherence to No-Arbitrage Principles: Emergent Put--Call Parity}
\label{sec:parity}

A key requirement for any European option pricing methodology is consistency with no-arbitrage relationships that link prices across contracts. Among these, put--call parity,
\begin{equation}
C(t,S;K,\tau) - P(t,S;K,\tau) = S - K e^{-r\tau},
\end{equation}
plays a fundamental role by enforcing internal coherence between call and put prices. FINN is not trained to satisfy this relationship explicitly: call and put options are learned independently through the same self-supervised hedging objective. As such, adherence to put--call parity provides a strong out-of-sample test of whether the model has internalized the economic structure of arbitrage-free pricing rather than simply fitting prices locally.

\begin{table}
\centering
{\begin{tabular}{cc|cc|cc}
\hline
\multirow{2}{*}{$\sigma$ or $\xi$} & \multirow{2}{*}{$\tau$} &
\multicolumn{2}{c|}{Parity Error (GBM)} &
\multicolumn{2}{c}{Parity Error (Heston)} \\
\cline{3-6}
 & & MAD & MSE & MAD & MSE \\
\hline
\multirow{7}{*}{0.125}
& 0.24 & 0.078 (0.024) & 0.011 (0.005) & 0.106 (0.070) & 0.026 (0.031) \\
& 0.28 & 0.081 (0.025) & 0.012 (0.005) & 0.113 (0.079) & 0.030 (0.038) \\
& 0.32 & 0.082 (0.027) & 0.012 (0.006) & 0.119 (0.089) & 0.034 (0.046) \\
& 0.36 & 0.083 (0.027) & 0.012 (0.006) & 0.124 (0.098) & 0.037 (0.053) \\
& 0.40 & 0.082 (0.027) & 0.012 (0.006) & 0.129 (0.106) & 0.040 (0.061) \\
& 0.44 & 0.082 (0.026) & 0.012 (0.005) & 0.132 (0.114) & 0.043 (0.069) \\
& 0.48 & 0.081 (0.025) & 0.011 (0.005) & 0.133 (0.122) & 0.045 (0.078) \\
\hline
\multirow{7}{*}{0.150}
& 0.24 & 0.103 (0.072) & 0.023 (0.031) & 0.101 (0.042) & 0.021 (0.013) \\
& 0.28 & 0.108 (0.071) & 0.024 (0.032) & 0.105 (0.045) & 0.022 (0.015) \\
& 0.32 & 0.112 (0.071) & 0.025 (0.033) & 0.108 (0.046) & 0.023 (0.016) \\
& 0.36 & 0.115 (0.072) & 0.026 (0.035) & 0.109 (0.048) & 0.023 (0.017) \\
& 0.40 & 0.117 (0.074) & 0.027 (0.037) & 0.109 (0.049) & 0.023 (0.017) \\
& 0.44 & 0.119 (0.076) & 0.028 (0.040) & 0.108 (0.049) & 0.022 (0.017) \\
& 0.48 & 0.120 (0.078) & 0.028 (0.043) & 0.106 (0.050) & 0.021 (0.017) \\
\hline
\multirow{7}{*}{0.175}
& 0.24 & 0.105 (0.038) & 0.022 (0.015) & 0.086 (0.027) & 0.015 (0.008) \\
& 0.28 & 0.111 (0.041) & 0.025 (0.017) & 0.093 (0.029) & 0.018 (0.010) \\
& 0.32 & 0.114 (0.042) & 0.027 (0.019) & 0.098 (0.031) & 0.019 (0.010) \\
& 0.36 & 0.115 (0.043) & 0.027 (0.019) & 0.101 (0.031) & 0.020 (0.010) \\
& 0.40 & 0.113 (0.043) & 0.026 (0.019) & 0.101 (0.031) & 0.020 (0.010) \\
& 0.44 & 0.111 (0.041) & 0.025 (0.018) & 0.099 (0.030) & 0.019 (0.009) \\
& 0.48 & 0.108 (0.040) & 0.024 (0.016) & 0.096 (0.029) & 0.017 (0.009) \\
\hline
\end{tabular}}
\captionsetup{width=0.8\linewidth}
\caption{FINN put--call parity error under GBM and Heston models. Entries report mean RMAD and RMSE with standard deviations in parentheses (across 10 different training seeds).}
\label{tab:put_call_parity_comparison}
\end{table}

We evaluate parity consistency by measuring the deviation of FINN-implied call and put prices from the theoretical parity condition across maturities, strikes, and volatility regimes. Table~\ref{tab:put_call_parity_comparison} reports the resulting errors under both the GBM and Heston data-generating processes.

In the Black--Scholes (GBM) environment, FINN exhibits strong parity consistency across all tested maturities and volatility levels. We use mean absolute deviations (MAD) and mean squared error (MSE) as our evaluation metrics, with additional details can be found at ~\ref{sec:measures}. MAD remain below $0.12$, with corresponding MSE below $0.03$. These deviations are stable across maturities, indicating that parity adherence is not a temporary effect of short-horizon training but persists across the option surface.

Under stochastic volatility (Heston), parity deviations increase slightly, reflecting the greater complexity of the pricing environment and the absence of a closed-form solution. Nevertheless, FINN continues to exhibit robust no-arbitrage behavior: parity errors remain below $0.13$ in mean absolute terms and below $0.05$ in mean squared terms across all configurations. This stability is important given that volatility is both latent and time-varying, and that the model is trained without any explicit parity constraint.

\begin{figure}
{\centering
\begin{tabular}{@{}cc@{}}
\includegraphics[width=0.48\linewidth]{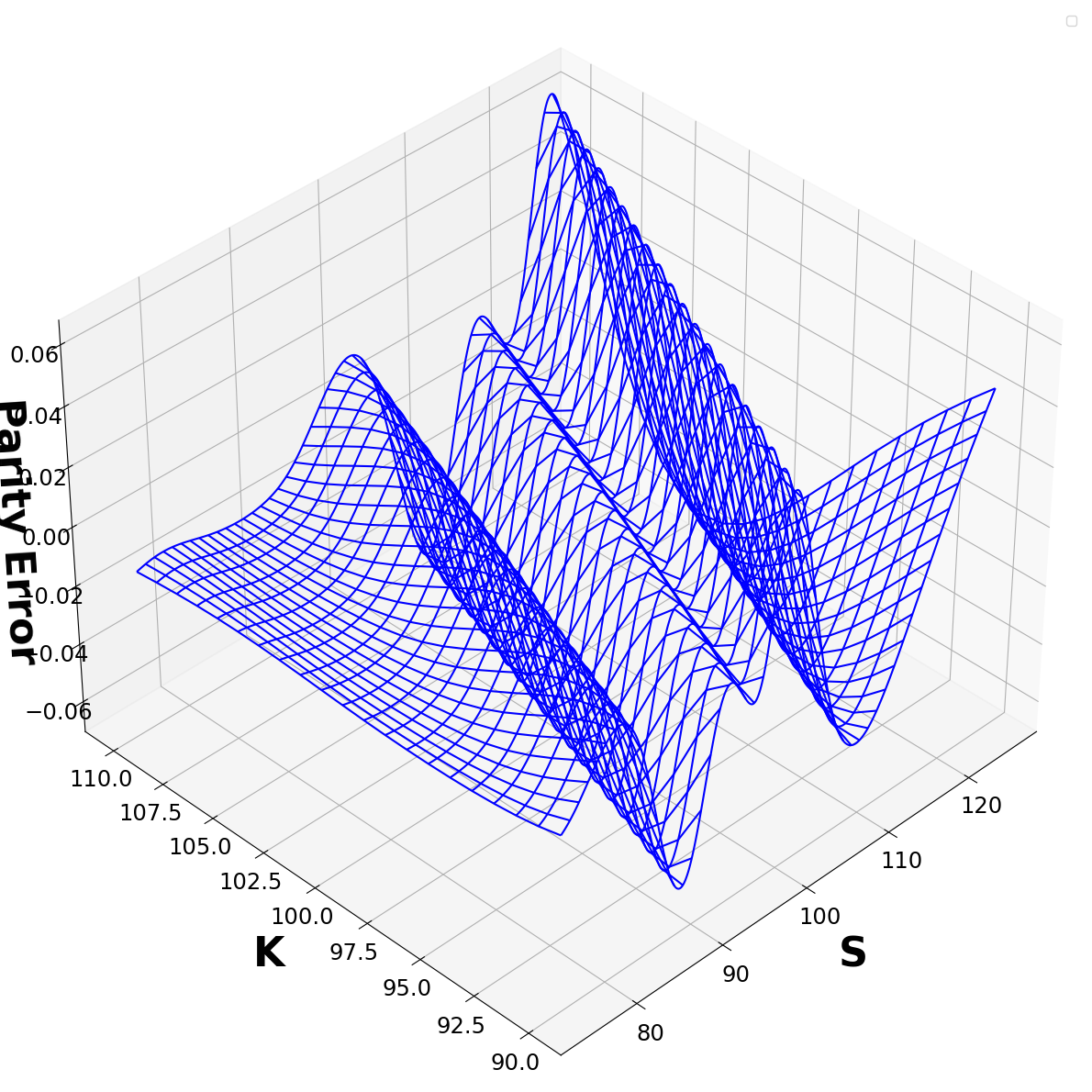} &
\includegraphics[width=0.48\linewidth]{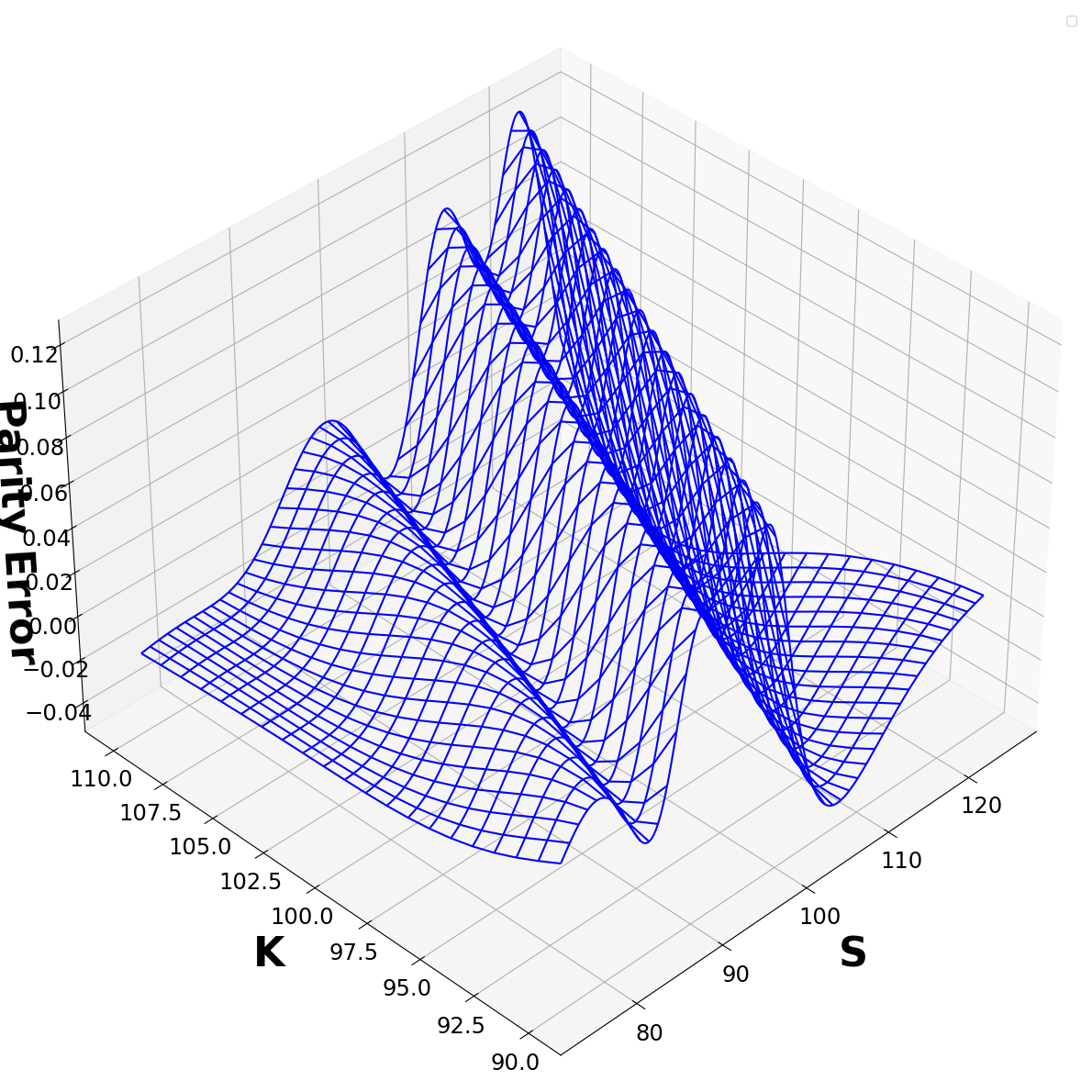} \\
{\footnotesize (a) Put--call parity error (GBM)} &
{\footnotesize (b) Put--call parity error (Heston)} \\
\end{tabular}
}
\caption{Spatial distribution of put-call parity errors for GBM and Heston models.}
\label{fig:parity_errors}
\end{figure}

Figure~\ref{fig:parity_errors} visualizes the spatial distribution of parity deviations across the state space. Errors are small and structured, with mild concentration in at-the-money regions where curvature is highest and learning is most challenging. In fact, there is no evidence of systematic parity violations or regime-dependent breakdowns.

Taken together, these results demonstrate that FINN learns no-arbitrage relationships endogenously through its replication-based training objective. The emergence of put--call parity, despite independent training and the absence of explicit constraints, provides strong empirical support for the claim that FINN internalizes the economic logic of arbitrage-free pricing rather than enforcing it mechanically.

\subsection{Application to Advanced Risk Management: Delta--Gamma Hedging}\label{sec:Delta-Gamma-Hedging}

We next evaluate the practical value and extensibility of the FINN framework in a setting that is central to professional risk management: delta--gamma hedging. Unlike delta-only strategies, delta--gamma hedging explicitly controls both first-order and second-order exposure to movements in the underlying asset, requiring accurate and stable estimation of higher-order Greeks. This setting therefore provides a strong test of whether FINN can support sophisticated hedging decisions beyond price prediction.

To this end, we augment the FINN loss function to construct a self-financing portfolio that simultaneously neutralizes both delta and gamma risk by incorporating a second, liquidly traded option as a hedging instrument. As described in Section~\ref{sec:Methodology}, the hedge ratios are determined endogenously from the network’s own learned sensitivities, without access to analytical Greeks.

\begin{table}
\centering
{\begin{tabular}{ccc|cc}
\hline
Vol $\sigma$ & Hedge TTM $\tau_h$ & Metric & RMAD at $\tau=0.24$ & RMAD at $\tau=0.48$ \\
\hline
\multirow{6}{*}{0.125} & \multirow{3}{*}{0.12} & Price $(C)$  & 0.003 (0.001) & 0.003 (0.002) \\
 &  & Delta $(\Delta)$ & 0.008 (0.002) & 0.007 (0.002) \\
 &  & Gamma $(\Gamma)$ & 0.059 (0.017) & 0.053 (0.013) \\
\cline{2-5}
 & \multirow{3}{*}{0.36} & Price $(C)$  & 0.003 (0.003) & 0.004 (0.004) \\
 &  & Delta $(\Delta)$ & 0.005 (0.002) & 0.005 (0.002) \\
 &  & Gamma $(\Gamma)$ & 0.046 (0.012) & 0.034 (0.009) \\
\hline
\multirow{6}{*}{0.150} & \multirow{3}{*}{0.12} & Price $(C)$  & 0.003 (0.002) & 0.003 (0.001) \\
 &  & Delta $(\Delta)$ & 0.005 (0.002) & 0.005 (0.002) \\
 &  & Gamma $(\Gamma)$ & 0.039 (0.010) & 0.030 (0.006) \\
\cline{2-5}
 & \multirow{3}{*}{0.36} & Price $(C)$  & 0.004 (0.003) & 0.004 (0.003) \\
 &  & Delta $(\Delta)$ & 0.005 (0.002) & 0.005 (0.002) \\
 &  & Gamma $(\Gamma)$ & 0.039 (0.007) & 0.027 (0.008) \\
\hline
\multirow{6}{*}{0.175} & \multirow{3}{*}{0.12} & Price $(C)$  & 0.003 (0.002) & 0.003 (0.002) \\
 &  & Delta $(\Delta)$ & 0.005 (0.002) & 0.004 (0.001) \\
 &  & Gamma $(\Gamma)$ & 0.033 (0.012) & 0.025 (0.008) \\
\cline{2-5}
 & \multirow{3}{*}{0.36} & Price $(C)$  & 0.002 (0.002) & 0.003 (0.003) \\
 &  & Delta $(\Delta)$ & 0.005 (0.005) & 0.005 (0.005) \\
 &  & Gamma $(\Gamma)$ & 0.031 (0.012) & 0.026 (0.015) \\

\hline
\end{tabular}}
\caption{Summary of FINN delta--gamma hedging performance of European call option. Entries report mean (standard deviation) across 10 different training seeds for the shortest ($\tau=0.24$) and longest ($\tau=0.48$) maturities; full results are in Appendix~\ref{app:full_gamma_results}.}
\label{table_gamma_summary}
\end{table}

\begin{figure}
{\centering
\begin{tabular}{@{}ccc@{}}
\includegraphics[width=0.32\linewidth]{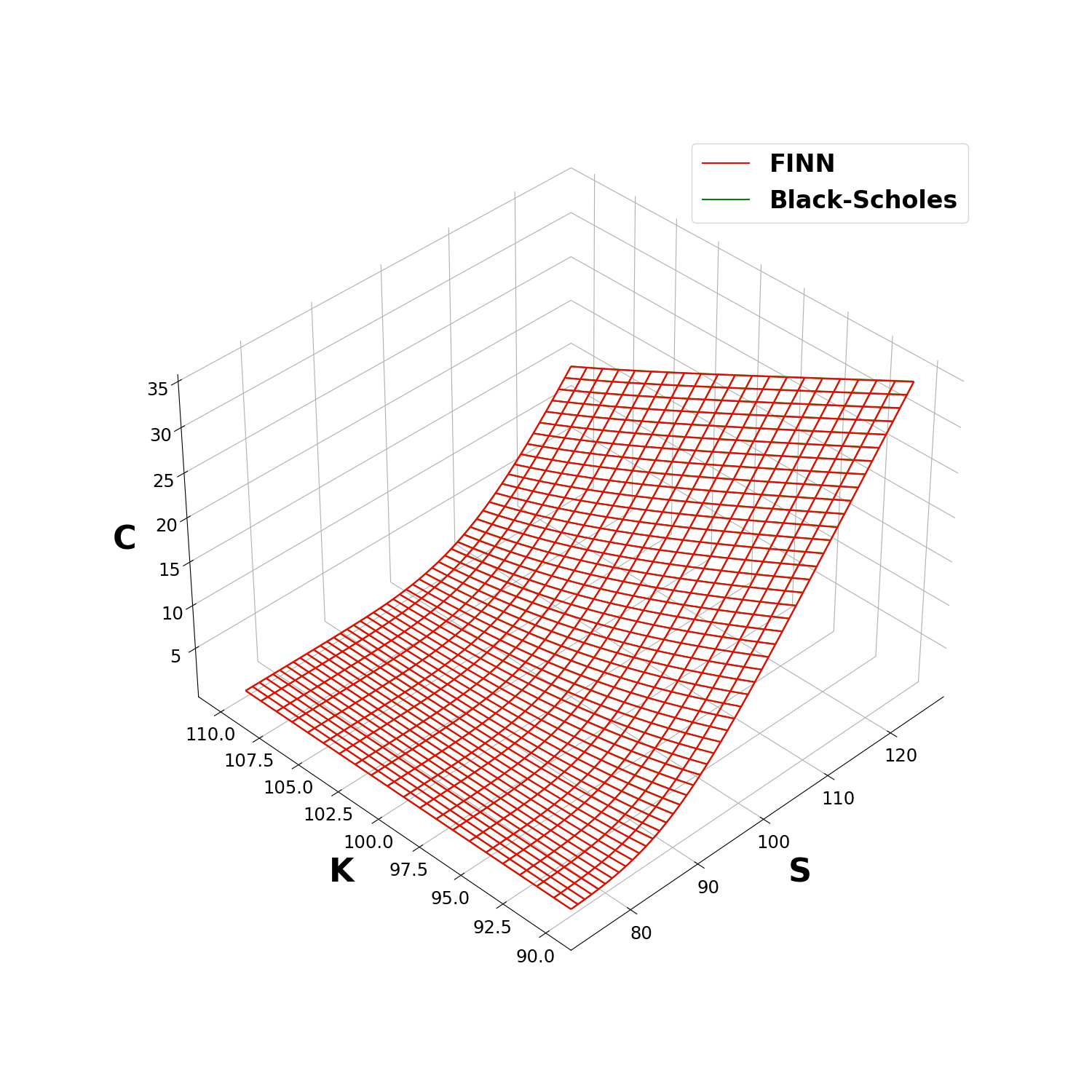} &
\includegraphics[width=0.32\linewidth]{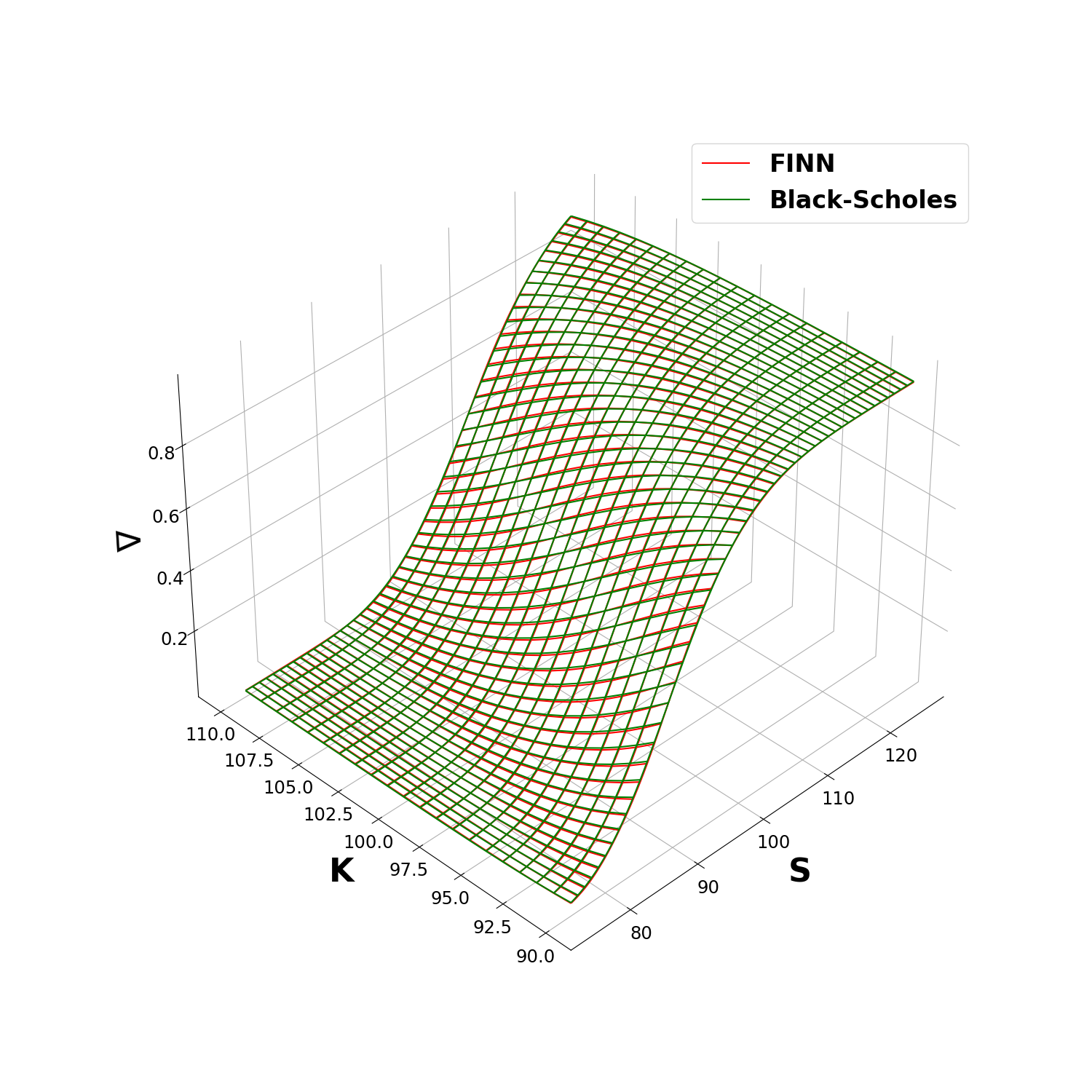} &
\includegraphics[width=0.32\linewidth]{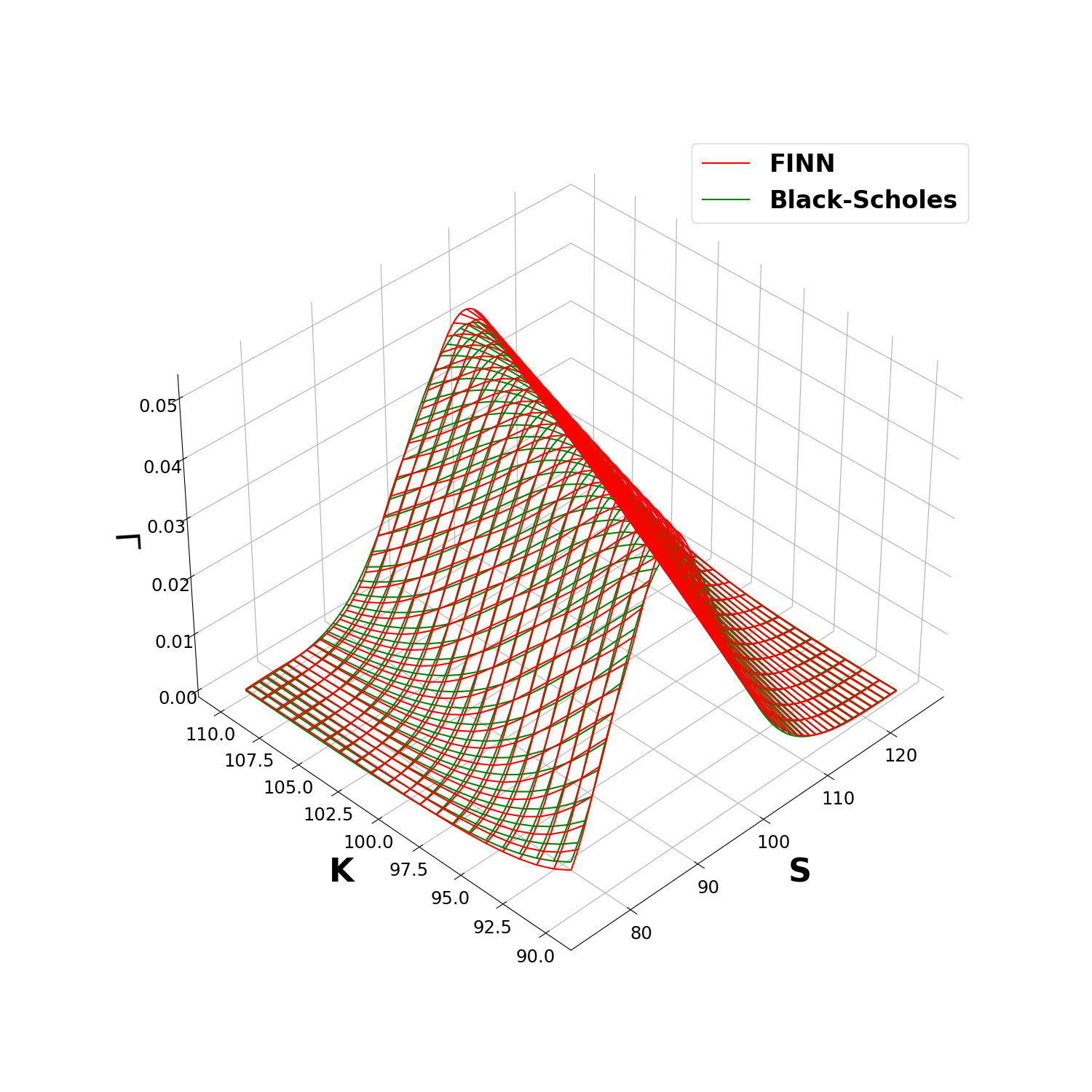} \\
{\footnotesize (a) Estimated call price} &
{\footnotesize (b) Estimated hedge ratio $(\Delta)$} &
{\footnotesize (c) Estimated gamma $(\Gamma)$} \\
[2mm]
\includegraphics[width=0.32\linewidth]{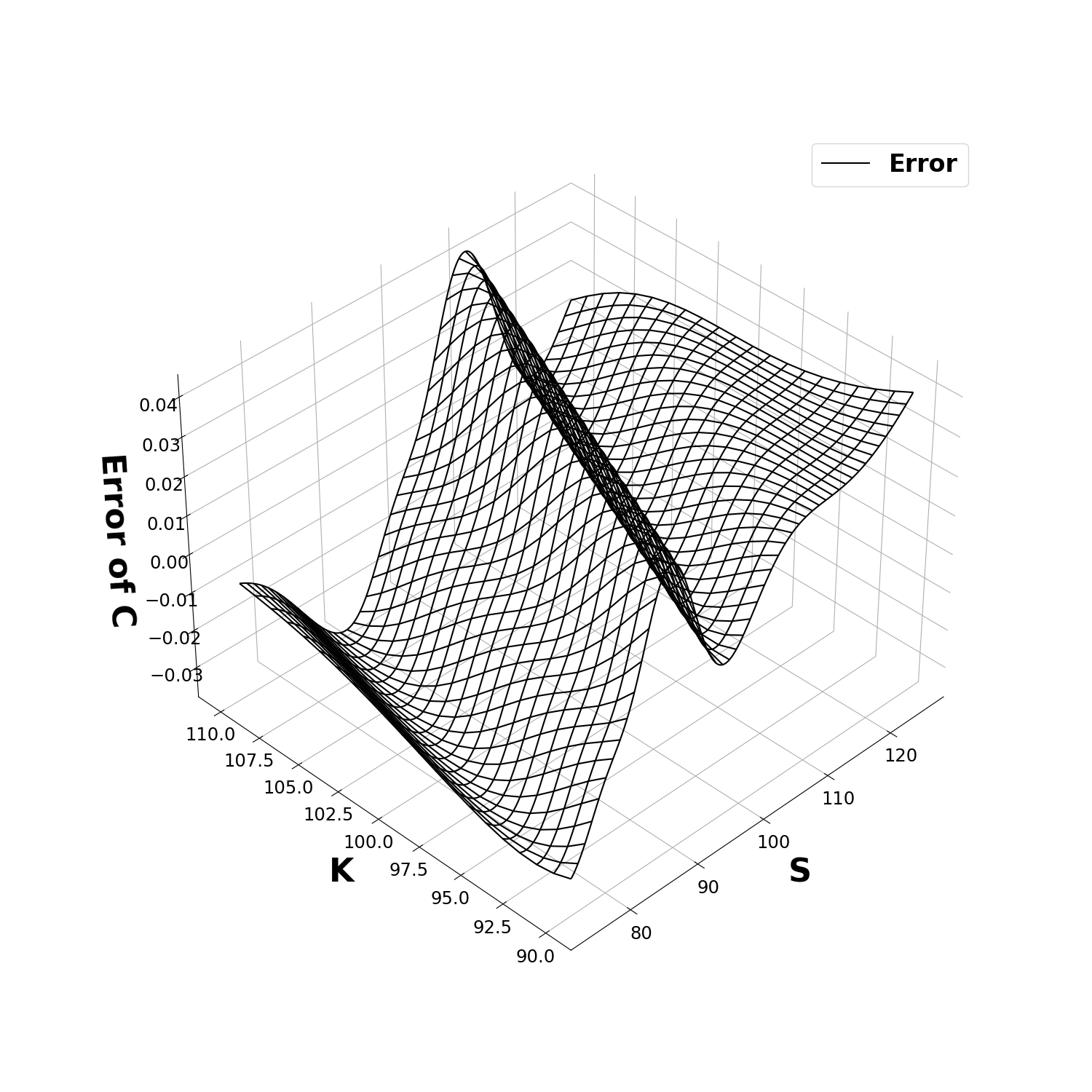} &
\includegraphics[width=0.32\linewidth]{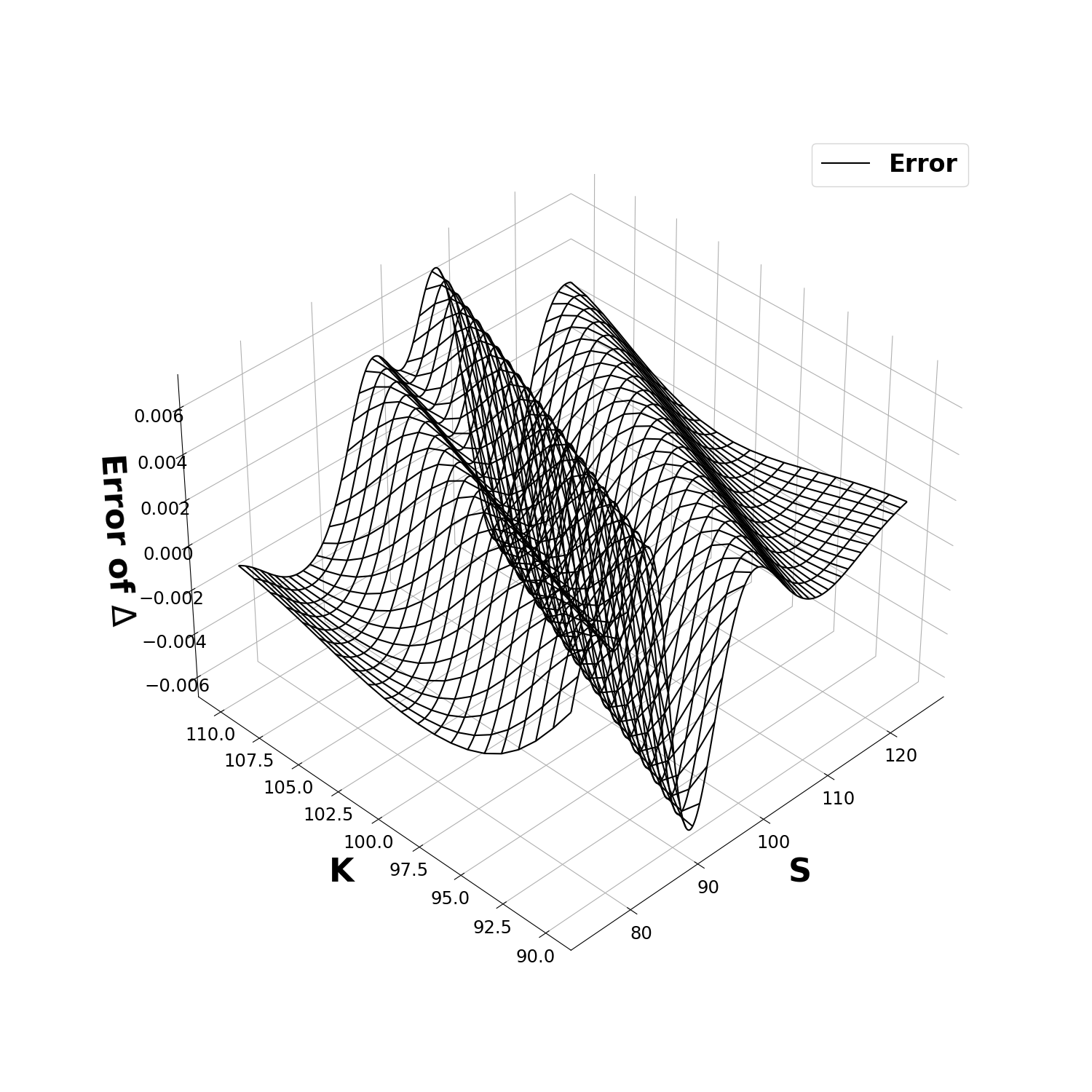} &
\includegraphics[width=0.32\linewidth]{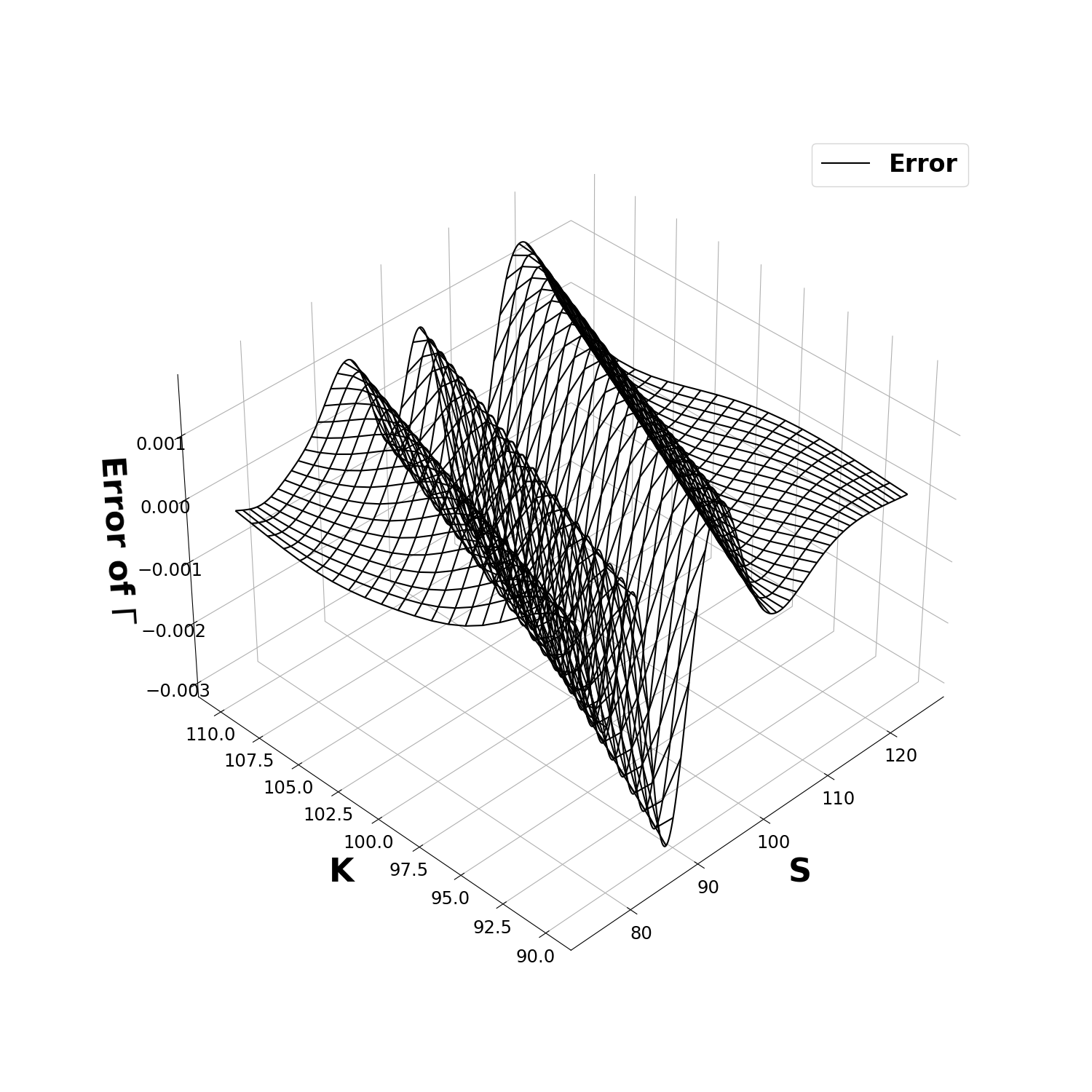} \\
{\footnotesize (d) Price error} &
{\footnotesize (e) Delta error} &
{\footnotesize (f) Gamma error} \\
\end{tabular}
}
\caption{GBM ($\sigma=0.15$, $\tau=0.36$): FINN surfaces for price, delta, and gamma (top row) and the corresponding errors (bottom row).}
\label{fig:gbm_gamma_six_plots}
\end{figure}

Table~\ref{table_gamma_summary} summarizes the resulting performance across volatility regimes and maturities. The results show a substantial improvement over delta-only hedging. Across all configurations, the relative mean absolute deviation (RMAD) for option prices remains extremely small, typically below $0.04$, representing an order-of-magnitude reduction compared to the delta-only results reported in Table~\ref{finn_gbm}. At the same time, FINN achieves highly accurate estimation of second-order risk: the RMAD for gamma remains consistently low across all scenarios, indicating that the model learns curvature with good precision.

An important practical consideration in delta--gamma hedging is the choice of hedging instrument. The table shows that FINN’s performance is robust to this choice: using a near-maturity ($\tau_h=0.12$) versus longer-maturity ($\tau_h=0.36$) option as the gamma hedge results in only minor variations in error. This robustness shows that FINN’s gains are not driven by fine-tuned instrument selection but by the theory-informed structure of the learning objective itself. Incorporating additional financial structure directly into the loss function therefore translates into tangible improvements in hedging accuracy.

Figure~\ref{fig:gbm_gamma_six_plots} provides a visual confirmation of these results. The top panels show that the FINN-implied surfaces for the option price, delta, and gamma are nearly indistinguishable from the analytical Black--Scholes benchmarks. The bottom panels report the corresponding errors, which are uniformly small and concentrated in the at-the-money region where curvature is highest and hedging is most challenging. The ability of FINN to recover not only prices and deltas, but also second-order sensitivities with high fidelity, highlights its suitability as a unified framework for pricing and advanced risk management.

These findings demonstrate that FINN naturally extends beyond pricing to support multi-Greek hedging strategies. By embedding replication-based principles directly into the training objective, FINN delivers economically meaningful improvements in risk control, rather than simply improved statistical fit. So, the training objective can be augmented to enforce additional Greeks or sensitivities tailored to the risk requirements of a given downstream task, ensuring alignment with practical risk-management objectives.

\subsection{Application to American Options}
\label{sec:american_option_hedging}
American options introduce an optimal stopping feature through early exercise. In continuous time, the value process is the Snell envelope of the discounted payoff, and in the PDE formulation it solves a free-boundary variational inequality \cite{shreve2004stochastic}. For a non-dividend-paying equity under geometric Brownian motion (GBM), American calls coincide with European calls, while American puts exhibit a real early-exercise premium and a non-trivial exercise boundary. This makes the American put a natural one-dimensional benchmark: high-accuracy reference solutions are available via finite-difference (FD) methods, and both the free boundary and local curvature are sensitive to modeling and numerical errors. In this setting, the objective is not only to match prices, but also to recover deltas and an economically coherent exercise boundary at a nonzero risk-free rate. This provides a meaningful test of FINN’s self-supervised, hedging-based training when an inequality constraint $g \ge f$ must hold everywhere.

We consider an American put written on an underlying $S$ following risk-neutral GBM
\begin{align}
dS_t = r S_t\,dt + \sigma S_t\,dW_t, \quad t \in [0,T].
\end{align}
with constant volatility $\sigma>0$, risk-free rate $r>0$, strike $K$, and payoff $f(S) = (K-S)^+$ at maturity $T$. Let $g(t,S)$ denote the time-$t$ American option value. In the PDE formulation, $g$ solves the variational inequality
\(
\max\Bigl\{
  \partial_t g(t,S)
  + \tfrac{1}{2}\sigma^2 S^2\,\partial_{SS} g(t,S)
  + rS\,\partial_S g(t,S)
  - rg(t,S),
  \; f(S) - g(t,S)
\Bigr\}
= 0.
\)
with terminal condition $g(T,S) = f(S)$ and appropriate boundary conditions as $S \downarrow 0$ and $S \uparrow \infty$. The continuation and stopping regions are
\(
\mathcal{C} = \{ (t,S) : g(t,S) > f(S)\},  
\mathcal{S} = \{ (t,S) : g(t,S) = f(S)\},
\)
separated by a free boundary $S^\ast(t)$ such that immediate exercise is optimal when $S \le S^\ast(t)$. In our experiments, we fix the strike price at $K = 100$, vary the volatility parameter $\sigma$ and the time-to-maturity $\tau$, and keep the risk-free interest rate $r$ fixed.

The American FINN architecture is identical to the European case; only the loss is modified to encode early exercise. Paths of $(S_t)_{t=0,\dots,T}$ are generated under the risk-neutral GBM dynamics on a fixed time grid. For each path and time step $t$, the network produces $g^\theta(t,S_t)$ and, via automatic differentiation, the hedge ratio
\(
\Delta_t^\theta = \frac{\partial g^\theta}{\partial S}(t,S_t).
\)
The one-step hedged P\&L residual is
\(
R_t^\theta
=
\Delta_t^\theta\,(S_{t+\Delta t} - S_t)
-
\big(g^\theta(t+\Delta t, S_{t+\Delta t}) - g^\theta(t, S_t)\big),
\)
which coincides with the European FINN residual in the continuation region.

To restrict the hedging loss to states where continuation should hold, we introduce a smooth gate $\gamma_\eta(t,S_t)$ that approximates the indicator $\mathbf{1}_{\{g^\theta(t,S_t) > f(S_t)\}}$ as introduced in equation ~\eqref{eq:american_loss} in Section ~\ref{sec:Methodology}. Concretely,
\(
\gamma_\eta(t,S_t)
=
\sigma\!\big(\eta\,(g^\theta(t,S_t) - f(S_t))\big),
\)
where $\sigma(\cdot)$ is the logistic sigmoid and $\eta>0$ controls steepness. This gate is close to $0$ when the network output is at or below the payoff (candidate stopping region) and close to $1$ in the continuation region, while remaining differentiable near the free boundary. The hedging component of the loss uses the gated residual $\gamma_\eta(t,S_t)\,R_t^\theta$.

The American constraint $g \ge f$ is enforced through a quadratic penalty on violations. At each $(t,S_t)$ we define
\(
\Pi_t^\theta = \big(\max\{f(S_t) - g^\theta(t,S_t),\,0\}\big)^2
\)
and add $\lambda \,\Pi_t^\theta$ to the loss, where $\lambda>0$ is a tuning parameter. This term is active only when the network falls below the payoff surface, pushing $g^\theta$ back above $f$ and thus shaping the early-exercise region. The overall objective is the empirical expectation of the gated hedging residual squared plus the scaled penalty term, summarized in equation ~\eqref{eq:american_loss}. We use the same optimizer, learning-rate schedule, and mini-batching regime as in the European experiments, choosing $(\lambda,\eta)$ once and holding them fixed across all $(\sigma,\tau)$ configurations.

As a benchmark, we employ a Crank–Nicolson finite-difference method combined with a penalty approach to compute American put prices. The finite-difference grid is refined iteratively until further refinement leads to changes in the computed prices below a prescribed tolerance. The resulting price and delta surfaces are then treated as the ground truth benchmarks.

\begin{figure}
{\centering
\begin{tabular}{@{}cc@{}}
\includegraphics[width=0.48\linewidth]{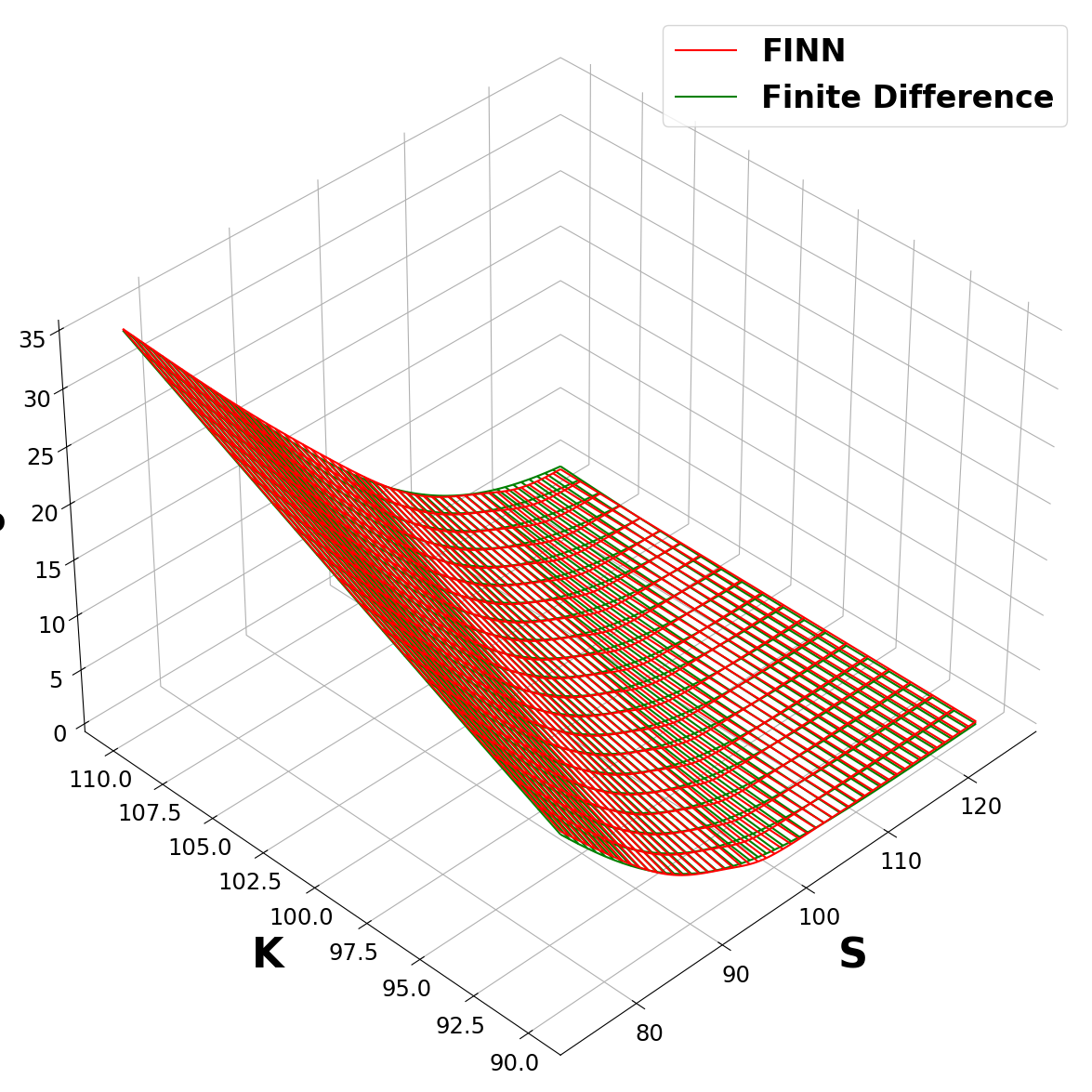} &
\includegraphics[width=0.48\linewidth]{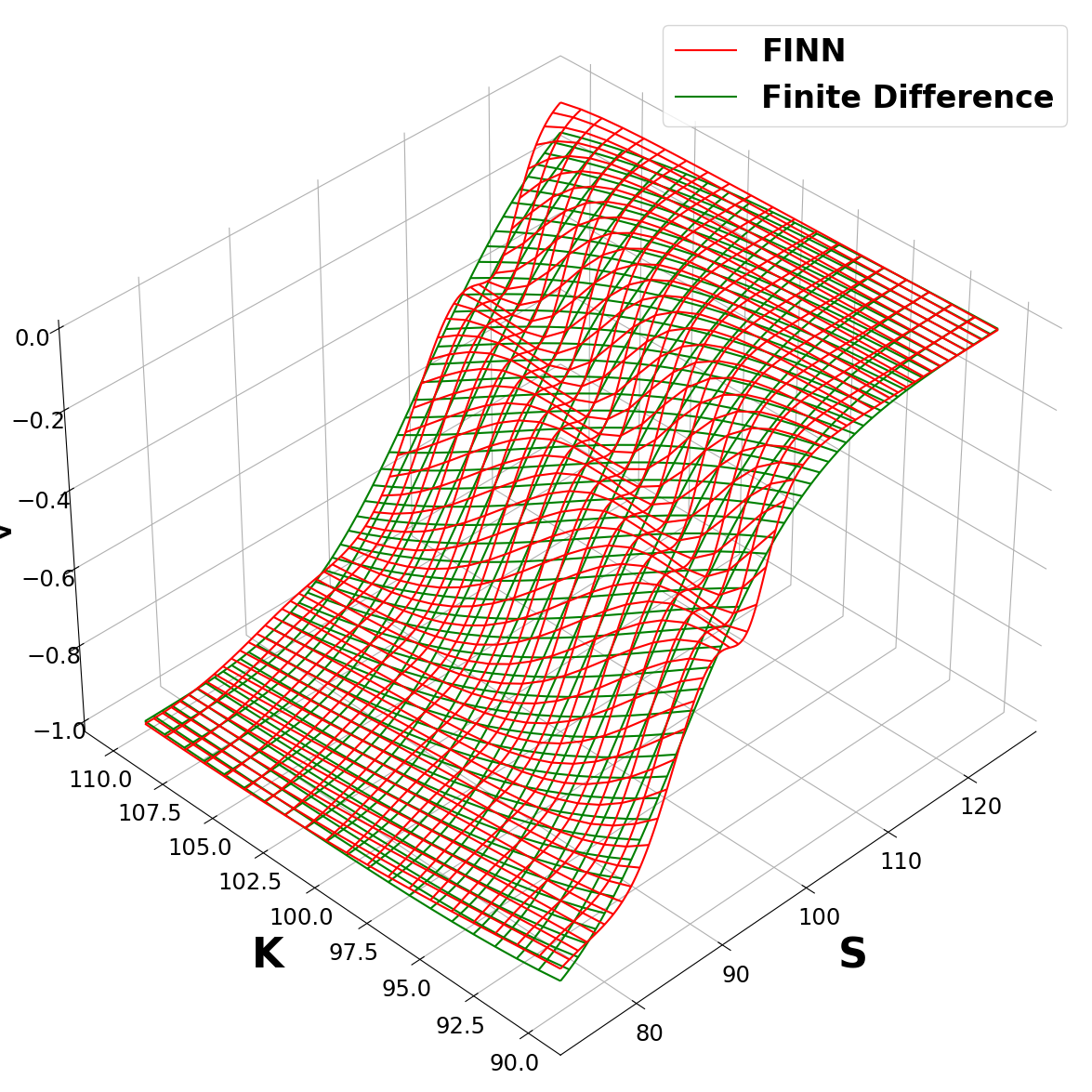} \\
{\footnotesize (a) Estimated American put price surface} &
{\footnotesize (b) Estimated hedge ratio $(\Delta)$ surface} \\
[2mm]
\includegraphics[width=0.48\linewidth]{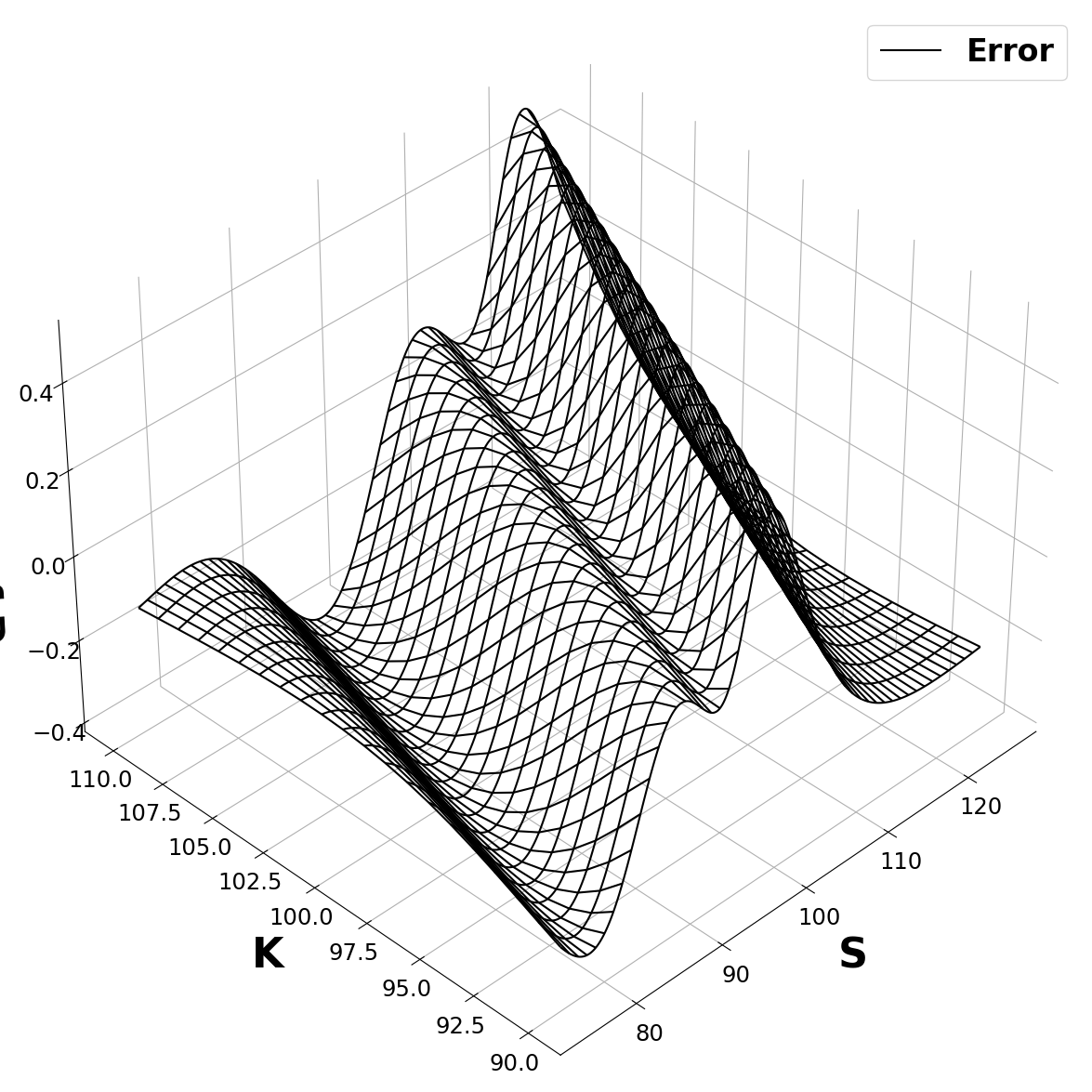} &
\includegraphics[width=0.48\linewidth]{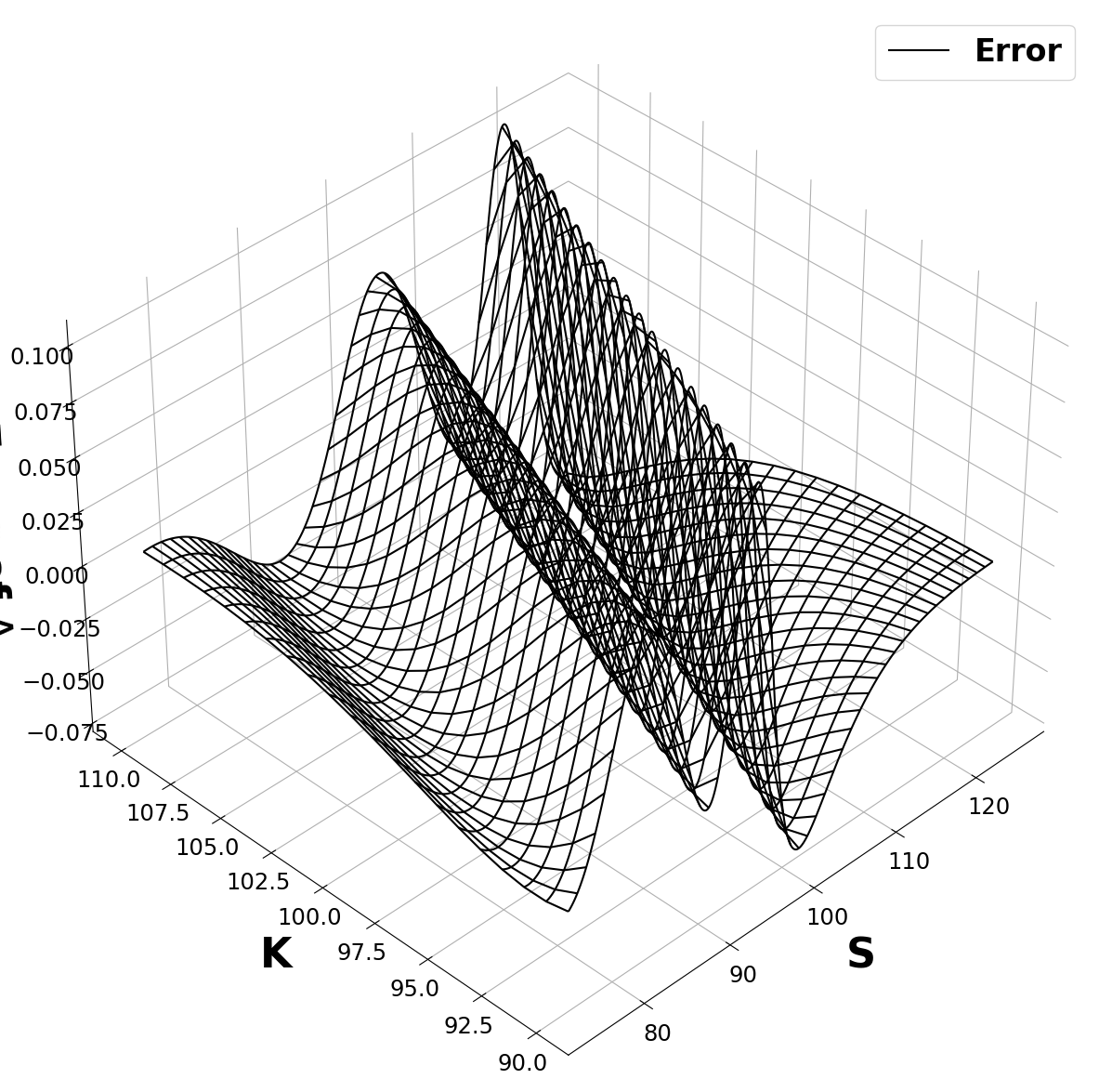} \\
{\footnotesize (c) Price error (FINN minus FD)} &
{\footnotesize (d) Delta error (FINN minus FD)} \\
\end{tabular}
}
\caption{Visual validation of FINN against the finite-difference (FD) benchmark for an American put under GBM ($\sigma=0.15$, $\tau=0.36$). FINN matches both the price and delta surfaces closely; residual errors concentrate near the at-the-money region and around the exercise boundary. The penalty parameter is $\lambda=10$.\label{fig:american_four_plots}}
{}
\end{figure}

Figure~\ref{fig:american_four_plots} compares FINN to the finite difference benchmark for a representative configuration ($\sigma=0.15$, $\tau=0.36$, $\lambda=10$). Panels~(a)--(b) show that the price and delta surfaces produced by FINN closely track the finite difference solution over the state space. Panels~(c)--(d) report the corresponding errors. As expected, the largest discrepancies concentrate near the at-the-money region and close to the exercise boundary, where curvature is highest and the exercise decision is most sensitive. The error surfaces do not exhibit systematic drift across $S$ and remain localized to these challenging regions.

\newpage

\begin{figure}
{\centering
\includegraphics[width=0.9\linewidth]{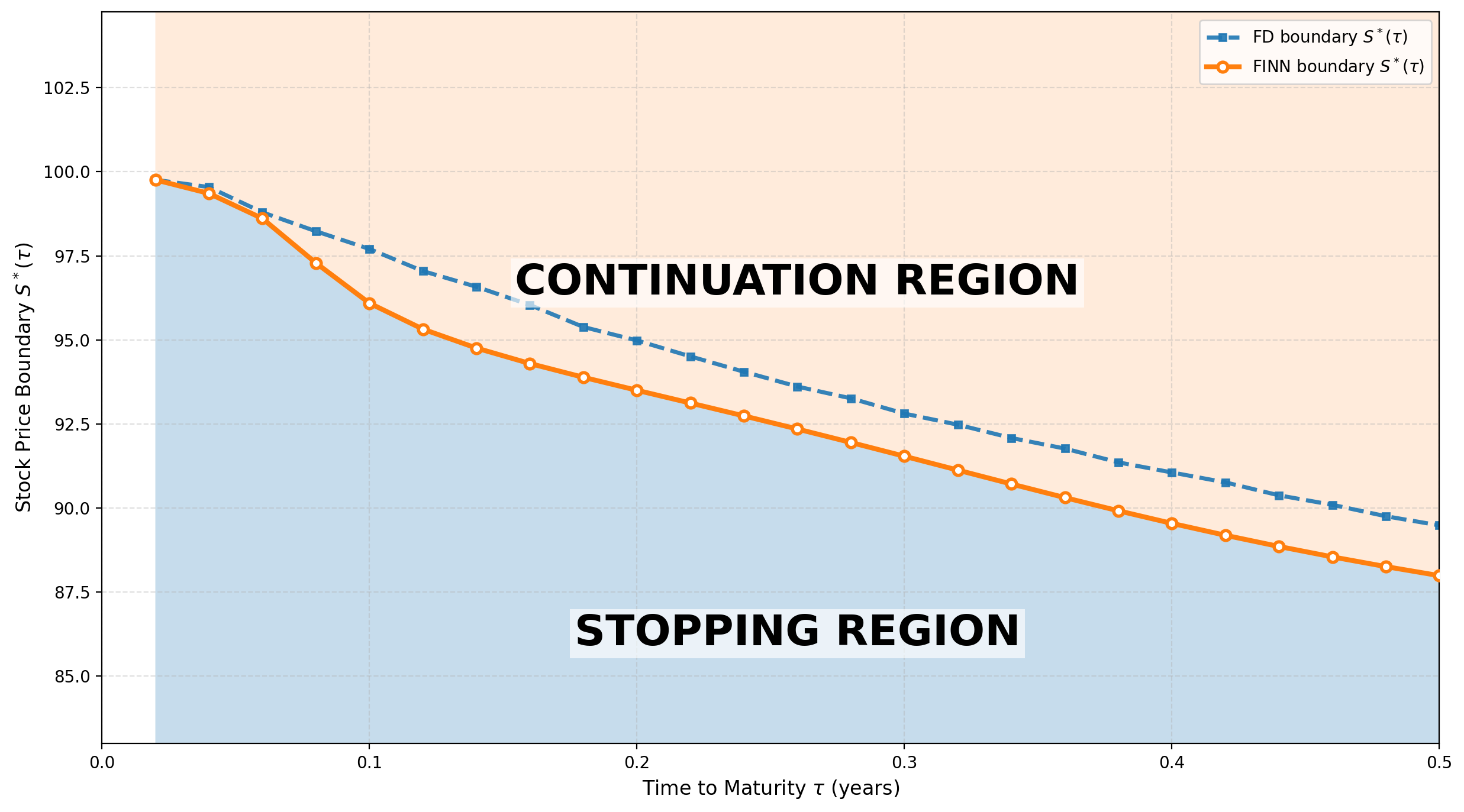}
}
\caption{American put early-exercise boundary estimated by FINN and a finite-difference (FD) benchmark. The FINN model is trained for \textbf{American puts} with constant volatility $\sigma=0.15$, risk-free rate $r=0$, and dividend yield $q=0$. The figure plots the optimal stopping boundary $S^*(\tau)$ as a function of time to maturity $\tau$. The dashed curve corresponds to the FD solution, while the solid curve shows the FINN-implied boundary extracted from the learned value function. Shaded regions indicate the stopping region ($S \le S^*(\tau)$) and the continuation region ($S > S^*(\tau)$) inferred by FINN. FINN closely tracks the FD boundary across maturities, with small and smooth deviations near short maturities.}
\label{fig:american_put_ex_bound}
\end{figure}

The learned early-exercise boundary extracted from FINN is shown in Figure \ref{fig:american_put_ex_bound} for $\sigma=0.15$ and $K=100$. The boundary tracks the finite difference benchmark closely over the full time interval, with mild smoothing caused by the sigmoid gate. This indicates that the combination of hedging-based training and a simple payoff-violation penalty can recover a coherent free boundary without explicitly solving a variational inequality.

\begin{table}
\centering
{\begin{tabular}{cc|cc|cc}
\hline
\multirow{2}{*}{$\sigma$} & \multirow{2}{*}{TTM $\tau$} &
\multicolumn{2}{c|}{Put Option Price $(P)$} &
\multicolumn{2}{c}{Hedge Ratio $(\Delta)$} \\
\cline{3-6}
 & & RMAD & RMSE & RMAD & RMSE \\
\hline
\multirow{7}{*}{0.125}
 & 0.240 & 0.049 (0.024) & 0.041 (0.020) & 0.078 (0.027) & 0.088 (0.030) \\
 & 0.280 & 0.052 (0.027) & 0.044 (0.023) & 0.081 (0.030) & 0.090 (0.035) \\
 & 0.320 & 0.054 (0.029) & 0.046 (0.024) & 0.084 (0.034) & 0.092 (0.039) \\
 & 0.360 & 0.057 (0.030) & 0.048 (0.026) & 0.087 (0.037) & 0.094 (0.042) \\
 & 0.400 & 0.059 (0.032) & 0.050 (0.028) & 0.090 (0.039) & 0.097 (0.044) \\
 & 0.440 & 0.061 (0.033) & 0.052 (0.029) & 0.094 (0.041) & 0.101 (0.046) \\
 & 0.480 & 0.062 (0.034) & 0.053 (0.030) & 0.098 (0.043) & 0.104 (0.047) \\
\hline
\multirow{7}{*}{0.150}
 & 0.240 & 0.036 (0.012) & 0.029 (0.009) & 0.074 (0.016) & 0.081 (0.022) \\
 & 0.280 & 0.036 (0.013) & 0.029 (0.010) & 0.074 (0.017) & 0.080 (0.022) \\
 & 0.320 & 0.036 (0.014) & 0.030 (0.011) & 0.075 (0.016) & 0.080 (0.021) \\
 & 0.360 & 0.036 (0.015) & 0.030 (0.011) & 0.076 (0.015) & 0.080 (0.020) \\
 & 0.400 & 0.036 (0.015) & 0.030 (0.012) & 0.079 (0.015) & 0.082 (0.019) \\
 & 0.440 & 0.036 (0.016) & 0.030 (0.013) & 0.082 (0.015) & 0.084 (0.019) \\
 & 0.480 & 0.036 (0.017) & 0.031 (0.013) & 0.085 (0.015) & 0.087 (0.019) \\
\hline
\multirow{7}{*}{0.175}
 & 0.240 & 0.030 (0.011) & 0.025 (0.010) & 0.066 (0.025) & 0.070 (0.028) \\
 & 0.280 & 0.028 (0.012) & 0.025 (0.010) & 0.061 (0.025) & 0.066 (0.029) \\
 & 0.320 & 0.028 (0.010) & 0.025 (0.010) & 0.060 (0.025) & 0.064 (0.030) \\
 & 0.360 & 0.027 (0.011) & 0.024 (0.011) & 0.060 (0.025) & 0.063 (0.029) \\
 & 0.400 & 0.029 (0.010) & 0.026 (0.011) & 0.061 (0.026) & 0.064 (0.028) \\
 & 0.440 & 0.039 (0.026) & 0.034 (0.021) & 0.064 (0.024) & 0.067 (0.025) \\
 & 0.480 & 0.041 (0.025) & 0.036 (0.019) & 0.066 (0.024) & 0.068 (0.024) \\
\hline
\end{tabular}}
\caption{American put option FINN prediction RMAD/RMSE results with a finite-difference benchmark for different $\sigma$ values. Entries report mean (standard deviation) across seeds.}
\label{tab:finn_results_put_rmad_rmse}
\end{table}

Table~\ref{tab:finn_results_put_rmad_rmse} summarizes relative errors (RMAD/RMSE) of prices and deltas across volatility levels and maturities, averaged across random seeds. Table~\ref{tab:finn_results_put_nmad_mse} reports the corresponding normalized errors (NMAD/MSE) in the Appendix. Overall, FINN achieves stable accuracy across configurations: price errors remain small in relative terms and delta errors remain well controlled throughout the grid. The slight reduction in performance compared to the European case reflects the added complexity of the free boundary and the sharp change introduced by the exercise decision. Nevertheless, the resulting deltas remain sufficiently accurate to produce a smooth and economically meaningful.

Overall, these experiments show that the American FINN objective~\eqref{eq:american_loss} yields a coherent approximation of both the American value function and the associated optimal stopping rule in a controlled one-dimensional GBM setting, while preserving the self-supervised, replication-based nature of the training procedure.

\subsection{Robustness and Stress Tests}\label{sec:robustness-tests}
Here we further assess the robustness of FINN-based hedging strategies under volatility regime shifts and transaction costs, conditions under which classical delta hedging is known to perform poorly. To isolate the effect of the hedge ratio itself, we work in a geometric Brownian motion (GBM) setting for all experiments, but deliberately introduce systematic mismatches between the volatility used to compute hedge ratios and the volatility governing realized price dynamics and option valuation.

For each test year $Y\in\{2008,2011,2020\}$, we compute a ``training'' volatility $\sigma_{\text{train}}$ as the realized annualized volatility of daily log returns of S\&P 500 from January~1990 up to the end of year $Y-1$, and a ``test'' volatility $\sigma_{\text{test}}$ as the realized annualized volatility within year $Y$. The training volatility is used to define the model under which deltas are computed, while the test volatility is used to drive the simulated stock paths and to compute option values along the simulated paths. This design ensures that any differences in performance come from the choice of delta, rather than from differences in pricing models or valuation methods.

We deliberately choose the years 2008, 2011, and 2020 because they correspond to distinct market conditions. The years 2008 and 2020 are crisis financial years with sharp volatility spikes and large drawdowns, in which $\sigma_{\text{test}}$ is much higher than the long-run $\sigma_{\text{train}}$. The year 2011 is a moderate post-crisis environment with higher but less extreme volatility. Thus, the stress tests probe how the hedging schemes behave under systematic mismatches between $\sigma_{\text{train}}$ and $\sigma_{\text{test}}$ in both severe crises and a moderately stressed regime.

It is important to note that these results do not contradict the earlier finding that FINN recovers the Black–Scholes hedge under idealized conditions. In settings where volatility is correctly specified and markets are frictionless, the Black–Scholes delta is optimal and FINN converges to the same hedge. The stress tests considered here deliberately violate these assumptions through volatility misspecification, discrete rebalancing, and transaction costs. Under these conditions, we observe that FINN’s replication-based training objective yields hedge ratios that are more robust to model error, leading to improved hedging performance.

\begin{table}
\centering
\caption{Stress-test regimes and realized volatilities. For each test year $Y$, $\sigma_{\text{train}}$ is computed from 1990 up to $Y\!-\!1$, while $\sigma_{\text{test}}$ is computed within year $Y$.\label{tab:stress_test_sigmas}}
{\begin{tabular}{rlllrr}
\hline
Test year & Regime & Train start & Train end & $\sigma_{\text{train}}$ & $\sigma_{\text{test}}$ \\
\hline
2008 & Crisis   & 1990-01-01 & 2007-12-31 & 0.158 & 0.410 \\
2011 & Moderate & 1990-01-01 & 2010-12-31 & 0.186 & 0.233 \\
2020 & Crisis   & 1990-01-01 & 2019-12-31 & 0.174 & 0.347 \\
\hline
\end{tabular}}
{}
\end{table}

For each test year, we simulate independent stock price paths under a geometric Brownian motion (GBM) model with drift $\mu$ and volatility fixed at $\sigma_{\text{test}}$. The initial stock price is set to $S_0 = 100$. On each simulated path, we assume our portfolio begins with a short position in one at-the-money European call option with strike $K = S_0$ and an initial time to maturity of 90 trading days. The hedging is carried out over a 21-day period (approximately one month). During this window, the hedge is rebalanced once per trading day. The risk-free interest rate is set to $r = 0$, and the option position is kept unchanged during the experiment; no additional option trading takes place. All hedging trades are executed using the underlying stock.

We consider two alternative sources for the hedge ratios. Under the \emph{BSM--Delta} strategy, the hedge ratio is given by the Black--Scholes delta computed using a GBM model with volatility set to $\sigma_{\text{train}}$. Under the \emph{FINN--Delta} strategy, the hedge ratio is produced by a FINN model trained at the same volatility level $\sigma_{\text{train}}$ and evaluated at the current state $(S_t, K, \tau_t)$. In both strategies, the option value used for profit-and-loss (P\&L) accounting is the Black--Scholes price computed with volatility $\sigma_{\text{test}}$. This design isolates the role of the delta itself. The two strategies differ only in how the hedge ratio $\Delta_t$ is obtained, either from the Black--Scholes formula or from FINN, while all other aspects of pricing, trading, and valuation are held constant.

The hedge is implemented using a self-financing portfolio composed of stock and cash. Let $\alpha_t$ denote the number of shares held immediately after rebalancing on day $t$, and let $c_t$ represent the associated cash account. When the hedge position changes from $\alpha_t$ to $\alpha_{t+1}$, we buy or sell $\alpha_{t+1}-\alpha_t$ shares at price $S_{t+1}$. A proportional transaction cost $\delta \in \{0, 0.01, 0.02\}$ is applied to the traded amount. The cash account therefore evolves according to
\(
c_{t+1}
=
c_t
- (\alpha_{t+1}-\alpha_t)\, S_{t+1}
- \delta\,\bigl|\alpha_{t+1}-\alpha_t\bigr|\, S_{t+1}.
\)
No transaction costs are applied to the option position, as the option is shorted once at initiation and is not re-traded during the hedging horizon; transaction costs therefore arise only from rebalancing in the underlying stock. The total portfolio value at time $t$ is denoted by $\Pi_t$ and is therefore given by
\(
\Pi_t
=
- V^{\text{call}}_t
+ \alpha_t S_t
+ c_t,
\)
where $V^{\text{call}}_t$ denotes the market value of the short call computed with volatility $\sigma_{\text{test}}$. The realized profit and loss over the hedging period is defined as $\Pi_T - \Pi_0$, with $T = 21$.

For each test year and each level of transaction costs, we simulate $50{,}000$ independent GBM price paths using the testing year's volatility $\sigma_{test}$ and apply both the BSM--Delta and FINN--Delta hedging strategies. This produces a distribution of realized profit and loss (P\&L) outcomes for each experimental setting. To summarize these outcomes, we report the sample mean and standard deviation of P\&L, which capture average performance and overall variability. In addition, we assess downside risk by focusing on the left tail of the P\&L distribution, which captures rare but severe losses that are most relevant for risk management and regulatory assessment. Specifically, the value-at-risk at level $\alpha$, denoted $\mathrm{VaR}_{\alpha}$, is defined as the $\alpha$-level lower quantile of P\&L (for example, $\mathrm{VaR}_{95}$ corresponds to the 5th percentile). The conditional value-at-risk, $\mathrm{CVaR}_{\alpha}$, is the expected P\&L conditional on outcomes falling below $\mathrm{VaR}_{\alpha}$. We report $\mathrm{VaR}_{95}$, $\mathrm{CVaR}_{95}$, $\mathrm{VaR}_{99}$, and $\mathrm{CVaR}_{99}$ alongside the mean and standard deviation. Comparing these statistics across test years, transaction cost levels, and the two delta specifications allows us to assess how differences in hedging performance and tail risk are driven by the choice of model (BSM versus FINN), rather than by changes in market volatility or trading frictions.

\begin{table}
{\begin{tabular}{lcllrrrrrr}
\hline
Year & Train End & TC & Model & Mean & Std & VaR$_{95}$ & CVaR$_{95}$ & VaR$_{99}$ & CVaR$_{99}$ \\
\hline
\multirow{6}{*}{2008 (Crisis)} & \multirow{6}{*}{2007}
& 0\% & BSM--Delta   & \textbf{-0.018} & 1.635 & -2.308 & -2.821 & -3.140 & -3.589 \\
& &     & FINN--Delta & -0.021 & \textbf{1.304} & \textbf{-1.852} & \textbf{-2.273} & \textbf{-2.539} & \textbf{-2.918} \\
\cline{3-10}
& & 1\% & BSM--Delta   & -1.462 & 1.974 & -4.310 & -4.958 & -5.372 & -5.919 \\
& &     & FINN--Delta  & \textbf{-1.294} & \textbf{1.554} & \textbf{-3.557} & \textbf{-4.082} & \textbf{-4.409} & \textbf{-4.883} \\
\cline{3-10}
& & 2\% & BSM--Delta   & -2.906 & 2.328 & -6.340 & -7.120 & -7.620 & -8.280 \\
& &     & FINN--Delta  & \textbf{-2.566} & \textbf{1.816} & \textbf{-5.271} & \textbf{-5.909} & \textbf{-6.294} & \textbf{-6.869} \\
\hline
\multirow{6}{*}{2011 (Moderate)} & \multirow{6}{*}{2010}
& 0\% & BSM--Delta   & \textbf{-0.002} & 0.277 & -0.479 & -0.627 & -0.723 & -0.851 \\
& &     & FINN--Delta & -0.008 & \textbf{0.232} & \textbf{-0.426} & \textbf{-0.551} & \textbf{-0.628} & \textbf{-0.734} \\
\cline{3-10}
& & 1\% & BSM--Delta   & -0.861 & 0.413 & -1.573 & -1.787 & -1.926 & -2.103 \\
& &     & FINN--Delta  & \textbf{-0.720} & \textbf{0.335} & \textbf{-1.308} & \textbf{-1.480} & \textbf{-1.587} & \textbf{-1.735} \\
\cline{3-10}
& & 2\% & BSM--Delta   & -1.719 & 0.554 & -2.677 & -2.958 & -3.144 & -3.368 \\
& &     & FINN--Delta  & \textbf{-1.431} & \textbf{0.446} & \textbf{-2.205} & \textbf{-2.432} & \textbf{-2.576} & \textbf{-2.767} \\
\hline
\multirow{6}{*}{2020 (Crisis)} & \multirow{6}{*}{2019}
& 0\% & BSM--Delta   & \textbf{-0.013} & 0.988 & -1.426 & -1.769 & -1.994 & -2.294 \\
& &     & FINN--Delta & -0.015 & \textbf{0.719} & \textbf{-1.086} & \textbf{-1.368} & \textbf{-1.562} & \textbf{-1.806} \\
\cline{3-10}
& & 1\% & BSM--Delta   & -1.232 & 1.220 & -3.037 & -3.486 & -3.773 & -4.159 \\
& &     & FINN--Delta  & \textbf{-1.060} & \textbf{0.891} & \textbf{-2.437} & \textbf{-2.800} & \textbf{-3.048} & \textbf{-3.356} \\
\cline{3-10}
& & 2\% & BSM--Delta   & -2.450 & 1.464 & -4.663 & -5.222 & -5.590 & -6.050 \\
& &     & FINN--Delta  & \textbf{-2.104} & \textbf{1.072} & \textbf{-3.800} & \textbf{-4.245} & \textbf{-4.542} & \textbf{-4.924} \\
\hline
\end{tabular}}
\caption{Stress tests daily hedged P\&L with transaction cost $\delta\in\{0\%,1\%,2\%\}$. Train window 1990$\rightarrow Y\!-\!1$; test on volatility of the year $Y$. Hedge ratios computed under $\hat\sigma_{\text{train}}$; option market value computed under $\hat\sigma_{\text{test}}$.}
\label{tab:mc_gbm_delta_tc000_010_020}
\end{table}

Table~\ref{tab:mc_gbm_delta_tc000_010_020} summarizes the hedging performance of BSM--Delta and FINN--Delta across the three test years. When transaction costs are absent ($\delta=0$), the two strategies perform similarly. In all years, the average P\&L is close to zero for both approaches. While BSM--Delta shows a slightly less negative mean, FINN--Delta consistently exhibits lower variability and smaller downside risk, as reflected in its lower standard deviation and less negative VaR and CVaR values in 2008, 2011, and 2020. This indicates that even in frictionless settings, the FINN-based hedge delivers slightly more stable outcomes.

The differences become more pronounced once transaction costs are introduced. At both $\delta=1\%$ and $\delta=2\%$, FINN--Delta outperforms BSM--Delta in all three years, producing less negative average P\&L and clearly tighter left tails. For example, in the crisis year 2008 with a 2\% transaction cost, the mean loss decreases from $-2.91$ under BSM--Delta to $-2.57$ under FINN--Delta, and the 95\% CVaR improves from $-7.12$ to $-5.91$. Similar improvements are observed in 2020, where the mean loss at $c=2\%$ improves from $-2.45$ to $-2.10$, and the 95\% CVaR from $-5.22$ to $-4.25$. Even in the more moderate 2011 environment, FINN--Delta reduces both the average loss and tail risk at higher transaction costs.

Overall, these results suggest that the FINN-based delta is more robust in the presence of transaction costs and volatility changes. By avoiding reliance on a fixed parametric model, FINN produces smoother hedge adjustments that limit excessive trading. As a result, it achieves comparable performance in stable markets while providing meaningful reductions in trading losses and downside risk during periods of high volatility and market frictions. Furthermore, Appendix ~\ref{app:robustness_paths} provides additional detailed sample paths analysis for each of the year 2008, 2011, and 2020, which illustrates the reason why FINN--Delta is able to outperforms BSM-Delta: in adverse scenarios BSM--Delta tends to rebalance more aggressively and maintain larger absolute stock positions, leading to higher trading volume and transaction costs, while FINN--Delta produces smoother hedge ratios and therefore tighter P\&L tails.

\subsection{Implied Volatility Surfaces: Structural versus Learned Pricing Operators}
\label{sec:iv_surfaces_hybrid_vs_finn}

A persistent challenge in modeling implied volatility (IV) surfaces is separating structural features of option prices from transient market states. Empirically, the shape of the volatility smile; its skew, convexity, and term structure\footnote{The term structure of implied volatility refers to how implied volatility varies with time to maturity, that is, how volatility differs across short- and long-maturity options.}; evolves slowly, while its overall level can shift rapidly in response to changes in market conditions and risk sentiment\footnote{Risk sentiment refers to investors’ overall willingness to bear risk, often reflected in shifts between risk-taking and risk-averse behavior.}.This distinction is particularly transparent in stochastic volatility models such as Heston. In that setting, the instantaneous variance $v_0$ represents a rapidly varying latent state, whereas the remaining parameters,
\(
\Theta_{\text{struct}}=\{\kappa,\theta,\sigma,\rho\},
\)
where $\kappa$ denotes the speed of mean reversion of variance, $\theta$ the long-run variance level, $\sigma$ the volatility of variance, and $\rho$ the correlation between asset returns and variance innovations govern the geometry of the volatility surface and are commonly assumed to be stable over time. Standard practice calibrates all parameters jointly on historical data and freezes them thereafter. However, such static calibrations often generalize poorly out of sample, largely because $v_0$ decorrelates much faster than the structural parameters, meaning that shocks to the instantaneous variance dissipate on a much shorter time scale than changes in $\Theta_{\text{struct}}$.

All implied volatility surfaces in this section are built using daily end-of-day prices of European options from standard market SPY data sources. We keep only actively traded options with sufficient liquidity, reasonable bid--ask spreads, and maturities within a predefined range. Implied volatilities are then obtained by numerically inverting the Black--Scholes formula, after discarding contracts that violate basic economic pricing rules.

Motivated by these observations, we adopt a hybrid learning perspective for IV surface modeling. Drawing an analogy with transfer learning, we view the training period as a pre-training phase during which the model learns the stable geometric structure of the volatility surface. In the testing period, the model then performs online inference to adapt its latent state to new market conditions. This viewpoint naturally leads to a calibration strategy that decomposes the parameter space into time-invariant structural components and time-varying state variables.

Within this framework, we consider two complementary modeling approaches. The first extends the classical Heston model by explicitly separating the calibration of long-run structural parameters from the dynamic updating of the latent state, namely the instantaneous variance, which is adjusted to reflect current market conditions. The second approach abandons parametric stochastic dynamics altogether and instead employs a learned pricing operator (FINN) that directly maps the current market state to option prices and sensitivities. By doing so, FINN implicitly captures the geometry of the IV surface without assuming a specific diffusion process. Comparing these two approaches allows us to disentangle the sources of out-of-sample pricing errors. In particular, it enables us to assess whether such errors arise primarily from structural misspecification inherent in parametric models, or from limitations in the functional representation used to approximate the pricing relationship.

\subsubsection{Heston Model Calibration}
\label{subsec:hybrid_heston}
We adopt a hybrid calibration strategy for the Heston model that mirrors the structural--state decomposition described above. Here, we learn time-invariant parameters governing the long-run geometry of the implied-volatility surface on a long historical window, while allowing the instantaneous variance to adapt dynamically out of sample. We first initialize the model under the physical measure $\mathbb{P}$ using spot returns from 2010--2016 by matching empirical moments, which provides an economically grounded starting point and mitigates convergence to implausible parameter values. Building on this initialization, we calibrate the structural parameter set $\Theta_{\text{struct}}=\{\kappa,\theta,\sigma,\rho\}$ under the risk-neutral measure $\mathbb{Q}$ on the training option sample via least-squares fitting of implied volatilities, with the risk-free rate $r$ and dividend yield $q$ fixed from market proxies. In the out-of-sample period, the structural set $\Theta_{\text{struct}}^*$ is held fixed and the latent instantaneous variance $v_0^t$ is updated daily to match the observed implied-volatility cross-section, enabling rapid adaptation to changing market conditions while preserving the learned long-run surface geometry.

\subsubsection{FINN as a Learned Pricing Operator}
\label{subsec:finn_pricing_operator}
While the Heston framework provides a parametric and economically interpretable decomposition between structural parameters and latent volatility states, the FINN model approaches the implied volatility surface from a fundamentally different perspective. Rather than explicitly specifying a stochastic volatility law, FINN learns a nonlinear pricing operator that maps normalized contract characteristics to option prices, implicitly encoding the geometry of the volatility surface. Specifically, FINN is trained to approximate the pricing functional: $\mathcal{P} : (m, T) \mapsto C$ where the inputs consist of dimensionless invariants, forward moneyness $m = S / (K e^{-rT})$ and time-to-maturity $T$, and the output is a normalized option price. This normalization removes scale effects associated with the underlying price and strike, encouraging the network to learn structural regularities of the pricing surface rather than memorizing asset-specific levels.

From a representation-learning viewpoint, the FINN model can be interpreted as learning a low-dimensional manifold on which option prices lie. The network weights encode a time-invariant pricing geometry, analogous to the structural parameter set $\Theta_{\text{struct}}$ in the Heston model, but without imposing an explicit parametric form. In this sense, FINN replaces analytically specified dynamics with a data-driven approximation of the market's pricing operator. The trained network performs a forward pass to produce a dimensionless price prediction, which is then rescaled by the strike $K$ and adjusted to satisfy basic no-arbitrage conditions by enforcing a lower bound given by the option’s intrinsic value (i.e., the immediate exercise payoff). This step ensures that the learned pricing map remains economically admissible while preserving flexibility in the learned pricing surface.

Although FINN is trained directly on the underlying prices rather than implied volatilities, implied volatility remains the most natural quantity for comparing models and market data. To obtain an implied volatility surface from FINN, we invert the Black--Scholes pricing formula using the prices predicted by the network. Specifically, for each strike–maturity pair $(K, T)$, we solve
\(
    C_{\text{BS}}(\sigma; S, K, r, T) = C_{\text{FINN}}(K, T)
\)
for the volatility parameter $\sigma$ using a numerical root-finding procedure. Repeating this inversion across the full grid of strikes and maturities yields a FINN-implied volatility surface that can be directly compared to both market-implied and Heston-implied volatilities.

From a conceptual perspective, this inversion step is closely related to latent-state inference in the Heston framework. While Heston explicitly infers the instantaneous variance $v_0^t$ given fixed structural parameters, FINN implicitly infers volatility through the interaction of its learned representation with the Black--Scholes operator. In both cases, implied volatility emerges as a derived state variable, not a primitive modeling input.

Viewed through a unifying lens, Heston and FINN represent two complementary approaches to the same problem. Heston imposes a physics-inspired stochastic structure and performs online inference over a latent volatility state, whereas FINN learns a flexible, nonparametric pricing operator whose geometry is fixed after training. Comparing market-implied, Heston-implied, and FINN-implied volatility surfaces therefore allows us to disentangle:
\begin{itemize}
    \item errors arising from structural misspecification (parametric bias in Heston),
    \item errors arising from representation bias (function class limitations in FINN), and
    \item actual market non-stationarities that challenge both paradigms.
\end{itemize}
\noindent This unified comparison forms the basis for the empirical analysis that follows. 

\subsubsection{Empirical Comparison: Market vs.\ Hybrid Heston vs.\ FINN Implied Volatility Surfaces}
\label{subsec:iv_empirical_comparison}
We compare the out-of-sample implied volatility (IV) surfaces generated by the hybrid Heston latent-state adaptation framework and by the FINN learned pricing operator against market-implied volatilities. The evaluation focuses on both aggregate accuracy and stability across trading dates. Errors are measured directly in implied volatility units and defined as: 
\(
\mathrm{IV\ Error} = \widehat{\mathrm{IV}} - \mathrm{IV}_{\text{Market}},
\)
so that negative values correspond to systematic underestimation of market-implied volatility.

\begin{figure}
{\centering
\includegraphics[width=0.92\textwidth]{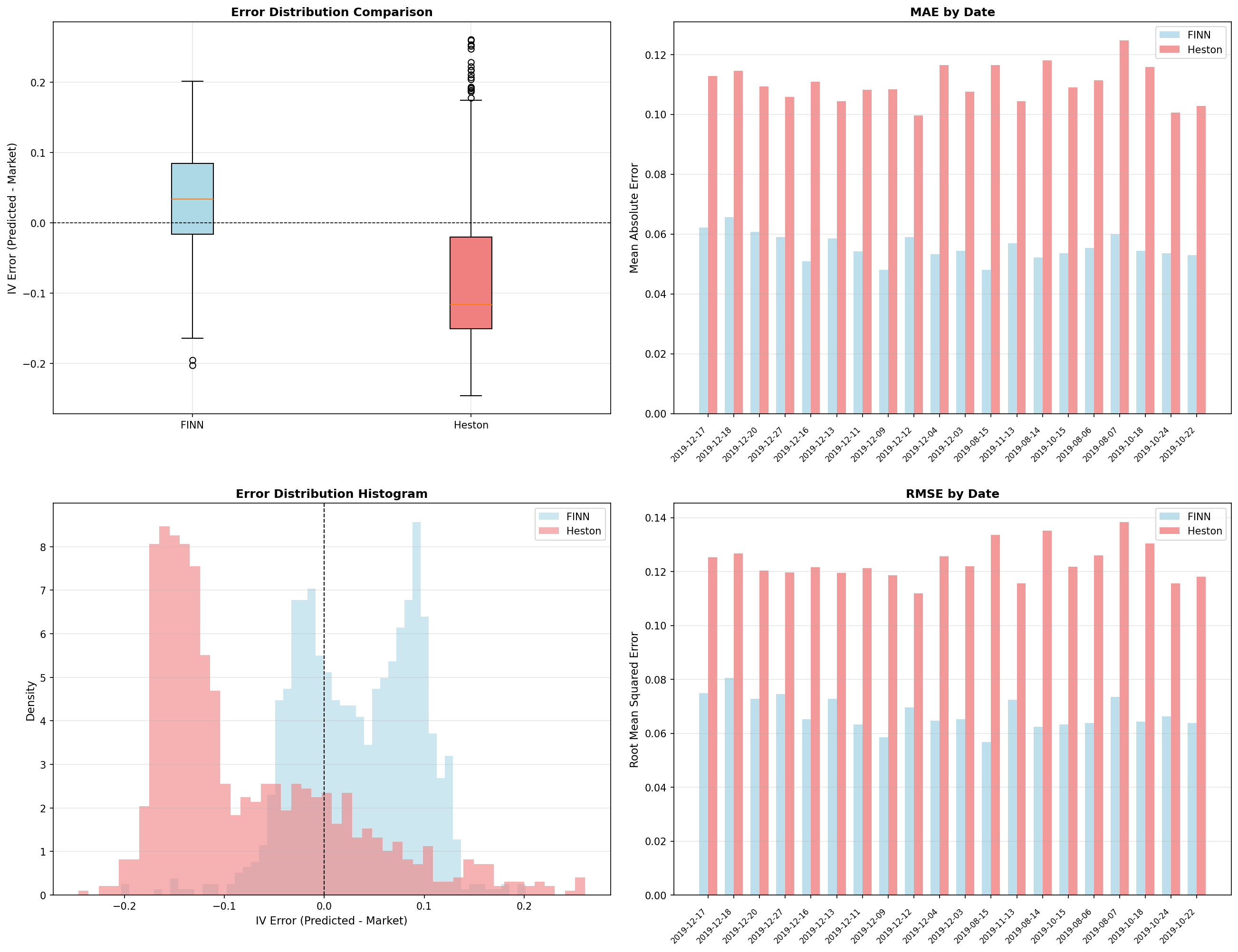}
}
\caption{Out-of-sample error summary. The boxplot (left) and histogram (bottom right) compare the distribution of IV errors ($\widehat{\mathrm{IV}}-\mathrm{IV}_{\text{Market}}$). Heston errors are heavily skewed negative, whereas FINN errors are centered near zero. The bar charts report daily MAE and RMSE, showing consistent FINN outperforming other methods.}
\label{fig:iv_error_summary}
\end{figure}

Figure~\ref{fig:iv_error_summary} summarizes the distribution of implied volatility errors and their stability across evaluation dates. We additionally provide the metrics we used in this section in Appendix section ~\ref{sec:measures}. Across the full out-of-sample period, FINN consistently outperforms the hybrid Heston framework in all reported accuracy metrics. Specifically, FINN reduces the mean absolute error (MAE), from $0.1153$ to $0.0456$, corresponding to a $60.5\%$ improvement, and the root mean squared error (RMSE) from $0.1218$ to $0.0562$, a $53.9\%$ reduction. The mean absolute percentage error (MAPE) decreases from $59.08\%$ to $26.95$, representing a $54.4\%$ improvement. In economic terms, an MAE reduction of approximately seven implied-volatility points corresponds to a meaningful improvement in option valuation accuracy and hedging sensitivity.

Importantly, the per-date bar charts in Figure~\ref{fig:iv_error_summary} indicate that these gains are persistent across the entire test period rather than being driven by a small subset of dates. This temporal stability suggests that FINN’s superior performance reflects improved generalization rather than episodic overfitting.

\begin{table}
\centering
{\begin{tabular}{lccc}
\hline
Model & MAE & RMSE & MAPE \\
\hline
FINN & 0.0456 & 0.0562 & 26.95\% \\
Hybrid Heston & 0.1153 & 0.1218 & 59.08\% \\
\hline
Relative improvement (FINN vs.\ Heston) & 60.5\% & 53.9\% & 54.4\% \\
\hline
\end{tabular}}
\caption{Out-of-sample IV accuracy (FINN vs.\ Heston). Errors are computed as $\widehat{\mathrm{IV}}-\mathrm{IV}_{\text{Market}}$ across all valid options in the evaluation set ($N=543$).}
\label{tab:iv_oos_metrics}
\end{table}

The magnitude of the baseline error observed for the hybrid Heston model is informative. Because the structural parameters $\Theta_{\text{struct}}$ are frozen after the 2010--2016 training period, this experiment directly probes the stationarity assumption underlying the Heston specification. The resulting error levels indicate that the effective structural curvature of the market-implied volatility surface is not fully time-invariant over multi-year horizons. In contrast, the FINN learned pricing operator exhibits substantially greater robustness, generalizing effectively to the 2017 regime despite being trained on the same historical window.

\begin{figure}
{\centering
\includegraphics[width=0.92\textwidth]{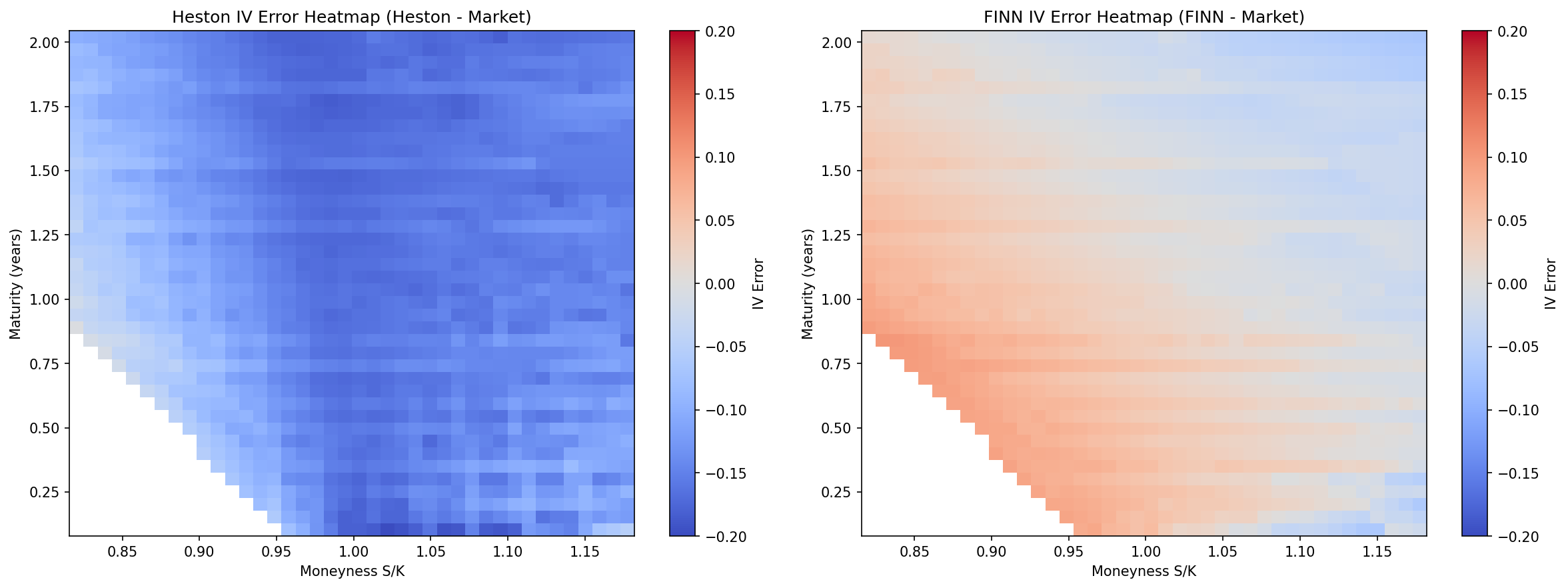}
}
\caption{Implied volatility error heatmaps over the $(S/K,T)$ grid. Left: hybrid Heston latent-state adaptation (Heston $-$ Market). Right: FINN learned pricing operator (FINN $-$ Market). The deep blue region in the Heston panel illustrates the systematic negative bias resulting from structural non-stationarity.}
\label{fig:iv_error_heatmaps}
\end{figure}

While the aggregate metrics in Table~\ref{tab:iv_oos_metrics} establish a clear and robust performance gap between FINN and the hybrid Heston framework, they do not reveal where in the state space these differences arise. In particular, option pricing errors are known to exhibit strong heterogeneity across moneyness and maturity, reflecting variations in curvature, skew, and tail sensitivity of the implied volatility surface. To better understand the geometric and economic structure of the residual errors, we next examine in Figure~\ref{fig:iv_error_heatmaps} the spatial distribution of implied volatility deviations over the $(S/K,\,T)$ domain. This heatmap analysis provides a localized diagnostic of model performance, highlighting whether errors are concentrated in regions that are most relevant for risk management and hedging.

Figure~\ref{fig:iv_error_heatmaps} visualizes implied-volatility errors across the $(S/K,\,T)$ domain. The hybrid Heston model exhibits a pronounced and widespread negative bias, indicating systematic underestimation of market-implied volatility over most strikes and maturities. This pattern persists despite daily re-estimation of the instantaneous variance $v_0^t$, suggesting that the historically calibrated structural parameters, in particular the mean-reversion speed $\kappa$ and volatility-of-volatility $\sigma$, impose curvature constraints that are misaligned with the 2017 market smile. 

\begin{figure}
{\centering
\includegraphics[width=0.90\linewidth]{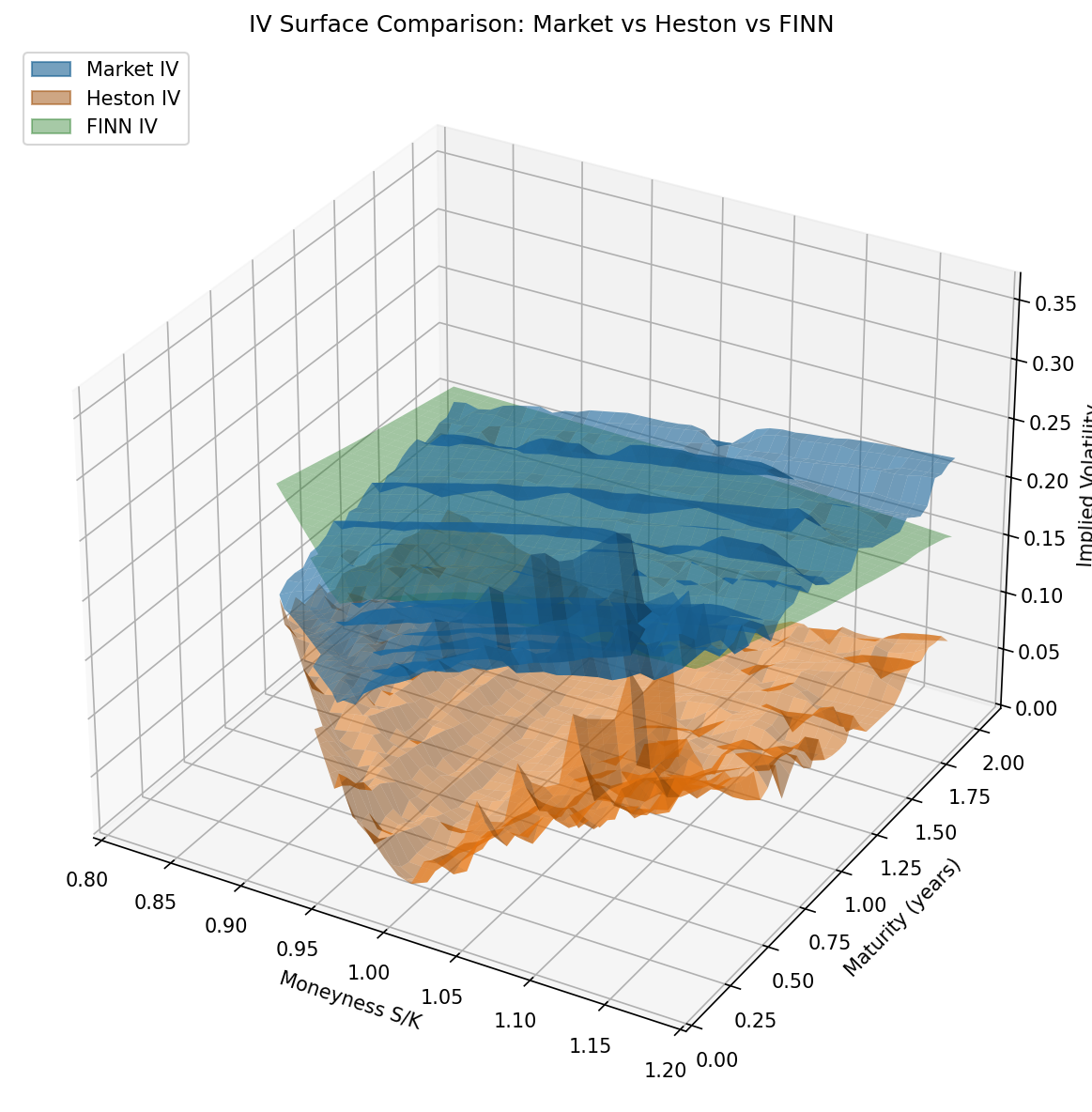}
}
\caption{IV surface overlay (Market vs.\ Heston vs.\ FINN). The Heston surface fails to capture the market level, illustrating the limitations of static structural parameters. FINN reconstructs the surface geometry.}
\label{fig:iv_surface_overlay}
\end{figure}

In contrast, FINN errors are centered closer to zero and display a weaker spatial structure. Residual deviations are localized and smaller in magnitude, indicating that the learned pricing operator adapts more flexibly across both moneyness and maturity dimensions. This difference in error geometry suggests that FINN captures the dominant shape of the implied volatility surface more robustly, while the remaining discrepancies likely reflect transient market effects rather than systematic structural bias. To complement the error-based analysis, Figures~\ref{fig:iv_surface_overlay} presents a direct visualization of the implied volatility surfaces. Here, we overlay the market, hybrid Heston, and FINN surfaces on a common grid. The hybrid Heston surface lies consistently below the market surface across much of the domain, indicating that adjustment through the instantaneous variance $v_0$ alone is insufficient to recover the observed volatility level. By contrast, the FINN surface closely follows the market surface in both level and slope across moneyness and maturity, reproducing the overall geometry of the implied volatility surface.

Taken together, these findings support the decomposition underlying our framework. First, structural misspecification in parametric models can lead to persistent pricing biases that are not eliminated by daily latent-state updates. Second, the FINN pricing operator mitigates representation bias, resulting in lower and more stable implied-volatility reconstruction errors. Finally, the remaining residual structure in FINN errors appears limited and localized, suggesting that it reflects actual market non-stationarities rather than systematic model deficiencies. 

\subsection{Learning Prices Where Option Markets Do Not Exist}
\label{sec:real_assets_no_options}

The preceding sections established FINN’s ability to recover arbitrage-consistent option prices and hedging sensitivities in controlled environments, including simulated dynamics and actively traded option markets. A natural next question is whether the framework can be deployed in settings where no options market exists at all. In such cases, traditional calibration-based approaches are infeasible, as neither implied volatilities nor liquid option prices are available.

This setting is practically important. Many financial instruments lack listed options at early stages of their lifecycle or remain permanently outside organized options markets. Examples include newly launched exchange-traded funds (ETFs), niche sector funds with limited liquidity, over-the-counter (OTC) securities, and specialized structured products that require internal valuation of contingent claims. In these contexts, practitioners must often rely on ad hoc assumptions or proxy instruments to assess derivative prices and risk.

In this section, we demonstrate that FINN can be trained directly on historical price data of the underlying asset, without specifying a stochastic differential equation or estimating model parameters. The resulting pricing operator produces arbitrage-consistent option values and Greeks, providing a principled, data-driven approach to initializing option prices for instruments without an existing derivatives market.

\begin{figure}[t]
{%
\centering
\includegraphics[width=\textwidth]{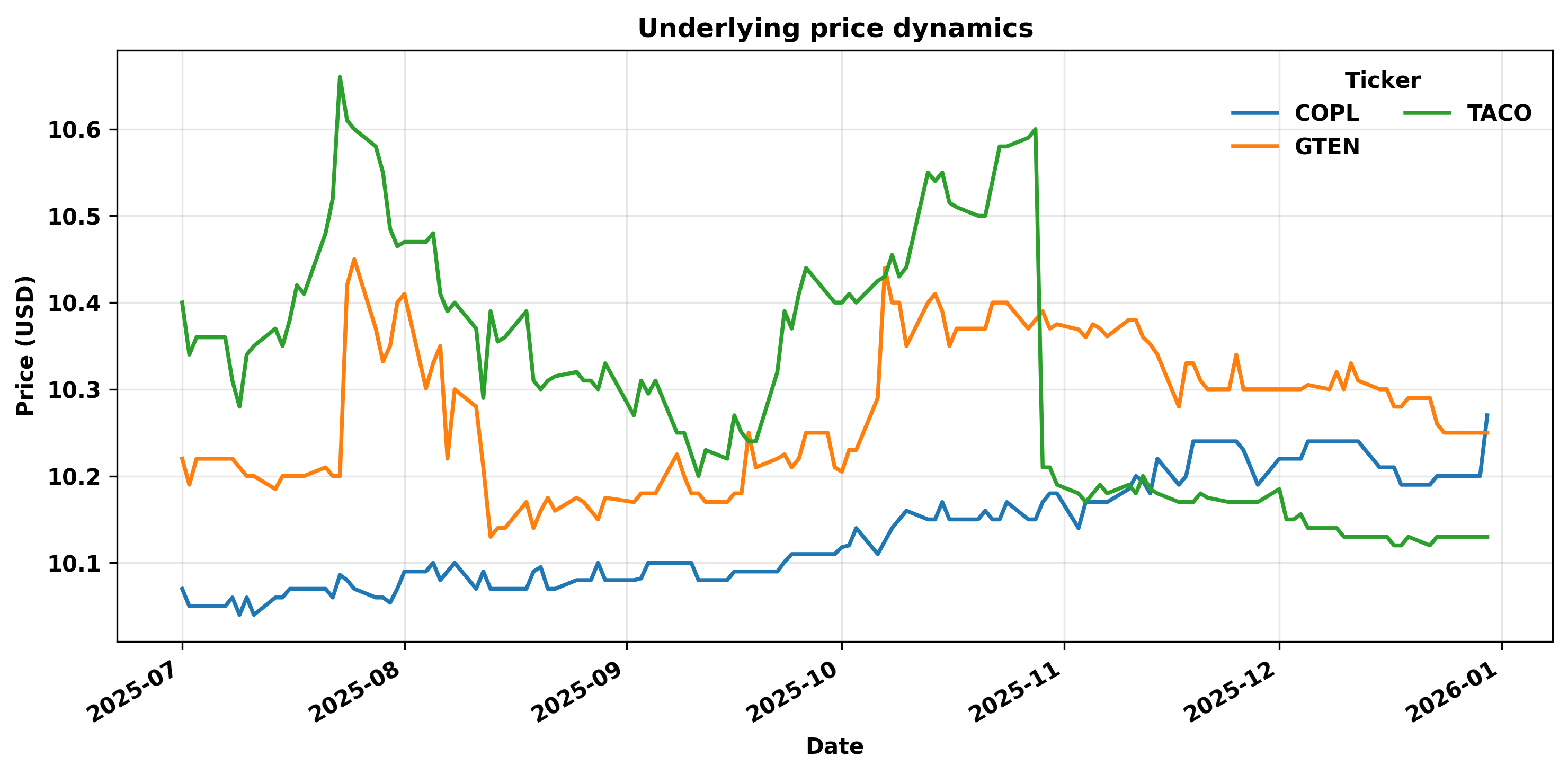}
}
\caption{%
Underlying price dynamics for the newly issued instruments \texttt{TACO}, \texttt{GTEN}, and \texttt{COPL}.The sample starts on 2025-07-01. }
\label{fig:finn_new_instruments_analysis}
\end{figure}

Here we consider a collection of recently issued financial instruments for which no actively traded listed options are available. Specifically, we study three tickers 
\[
\texttt{TACO},\ \texttt{GTEN},\ \texttt{COPL}
\]
all of which were introduced in 2025 and have limited historical price records of approximately 120 trading days, starting 2025-07-01. The short data history and absence of derivative markets make these assets unsuitable for standard option pricing methodologies based on implied volatility extraction or parametric calibration. Figure~\ref{fig:finn_new_instruments_analysis} displays the corresponding underlying price dynamics for these three assets.

In contrast to the simulation-based experiments presented earlier, FINN is trained directly on observed market data for each instrument, without imposing a parametric stochastic model for the underlying asset. Let $\{S_t\}_{t=1}^{T}$ denote the realized price trajectory of a given asset over the available historical window. This path serves as the unique source of information for training, and no synthetic data are generated.

This setting differs from the SDE-based experiments along three important dimensions.

\begin{enumerate}
    \item \textbf{Absence of SDE simulation.}  
    Rather than generating artificial price paths from a calibrated diffusion model, FINN is trained on the empirical trajectory itself. This removes simulation bias and ensures that training reflects the realized temporal structure of returns.

    \item \textbf{No parametric model calibration.}  
    We do not estimate or assume parameters from a specific stochastic process (e.g., GBM, Heston, or jump--diffusion). As a result, the approach avoids model risk associated with misspecification of the data-generating process and eliminates dependence on auxiliary calibration procedures.

    \item \textbf{Learning market-implied dynamics.}  
    By operating directly on observed prices, the network internalizes the effective dynamics implied by the data, including time-varying volatility, discontinuities, and regime changes. These features are absorbed implicitly through the replication-based training objective rather than being encoded in advance through a parametric law.
\end{enumerate}

As a result, this design positions FINN as a nonparametric pricing operator learned entirely from market data. The resulting option prices and sensitivities reflect the empirical behavior of the underlying asset while remaining guided by the no-arbitrage structure imposed through the hedging loss.

\subsubsection{Experiment Setup.}

Training batches are constructed by resampling local one-step price transitions from the observed historical trajectory. For each mini-batch, indices $i \in \{1,\dots,T-1\}$ are drawn uniformly, and the corresponding transition $(S_i,S_{i+1})$ is used in the replication loss. For each sampled index, we generate a hypothetical option contract by sampling
\begin{align}
\label{eq:sampling}
    S &= S_i, \quad S' = S_{i+1}, \\
    K &= S / m, \quad m \sim \mathcal{U}(0.5,\,2.0), \\
    \tau &\sim \mathcal{U}\!\left(\tfrac{1}{250},\,1\right),
\end{align}
where $m$ denotes moneyness and $\tau$ the time to maturity (in years). Uniform sampling in $m$ ensures balanced coverage across in-, at-, and out-of-the-money regions, while the maturity range spans short- to medium-term horizons commonly observed in listed option markets.

This procedure generates a diverse set of synthetic contracts $(K,\tau)$ while grounding the training signal entirely in realized market transitions. No synthetic price dynamics are introduced; all randomness arises from resampling observed transitions and contract characteristics. As a result, FINN learns a pricing operator that reflects empirical price behavior while enforcing no-arbitrage through the replication-based objective.

All instruments share the same FINN architecture and training protocol described in Sections~\ref{FINN-Parameterization_Self-Supervision} and~\ref{sec:design}. The model is a fully connected network with three hidden layers of widths $[256,512,256]$ and $\tanh$ activations. Inputs are normalized forward moneyness $S/(K e^{-r\tau})$ and time to maturity $\tau$ according to the sampling rule in ~\ref{eq:sampling}, and the output is a scalar option price with a softplus activation to ensure positivity. Training minimizes the self-supervised delta-hedging replication loss in~\eqref{eq:loss_discrete}.

Each model is trained for 250 epochs with 1{,}000 mini-batches per epoch (batch size 10{,}000) using the Adam optimizer. The learning rate follows an exponential decay schedule with initial value $5\times10^{-4}$ (decay rate $0.96$ with decay steps set to 100{,}000). The risk-free rate is fixed at $r=0$. No asset-specific hyperparameter tuning is performed, highlighting the robustness and transferability of the FINN framework across different assets.

\subsubsection{Results.}

FINN is trained directly on the realized spot-price histories of three newly issued instruments (\texttt{TACO}, \texttt{GTEN}, \texttt{COPL}), using data starting from 2025-07-01. For each ticker, separate models are trained for European calls and puts under the same self-supervised delta-hedging objective. In all cases, optimization is numerically stable: training losses decrease smoothly over epochs, with no signs of divergence or instability despite the relatively short sample length (approximately 120 trading days).

Out-of-sample hedging losses are uniformly small across instruments and option types. Averaged across tickers, the converged test loss is on the order of $10^{-6}$, indicating that the learned pricing operators track observed price changes with very small hedging discrepancies. Training behavior and final loss levels are consistent across all three instruments.

\begin{figure}[t]
    \centering

    \begin{subfigure}[t]{0.30\textwidth}
        \centering
        \includegraphics[width=\linewidth]{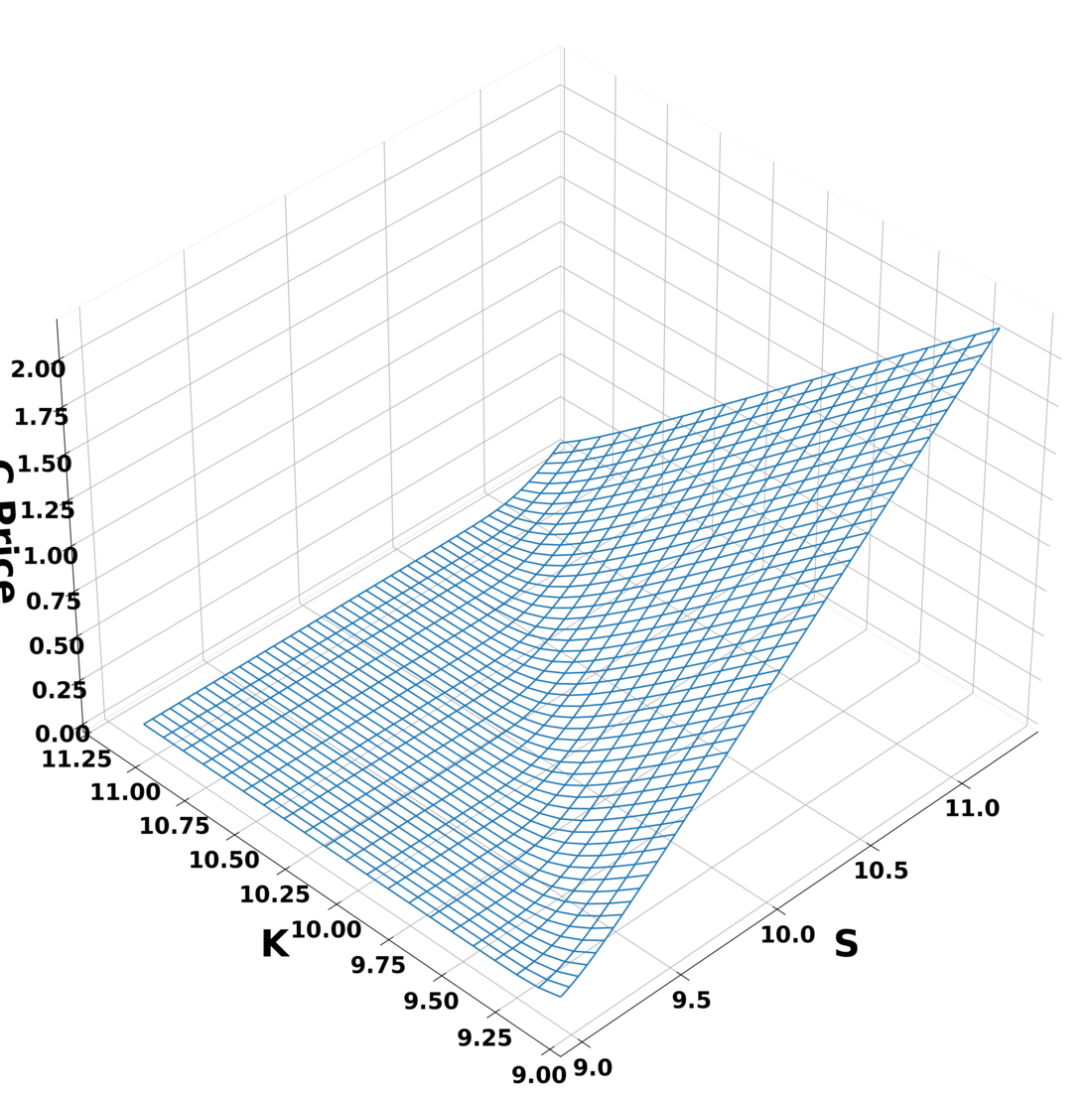}
        \caption{\texttt{COPL} Price}
        \label{fig:copl_call_price_surface}
    \end{subfigure}\hfill
    \begin{subfigure}[t]{0.30\textwidth}
        \centering
        \includegraphics[width=\linewidth]{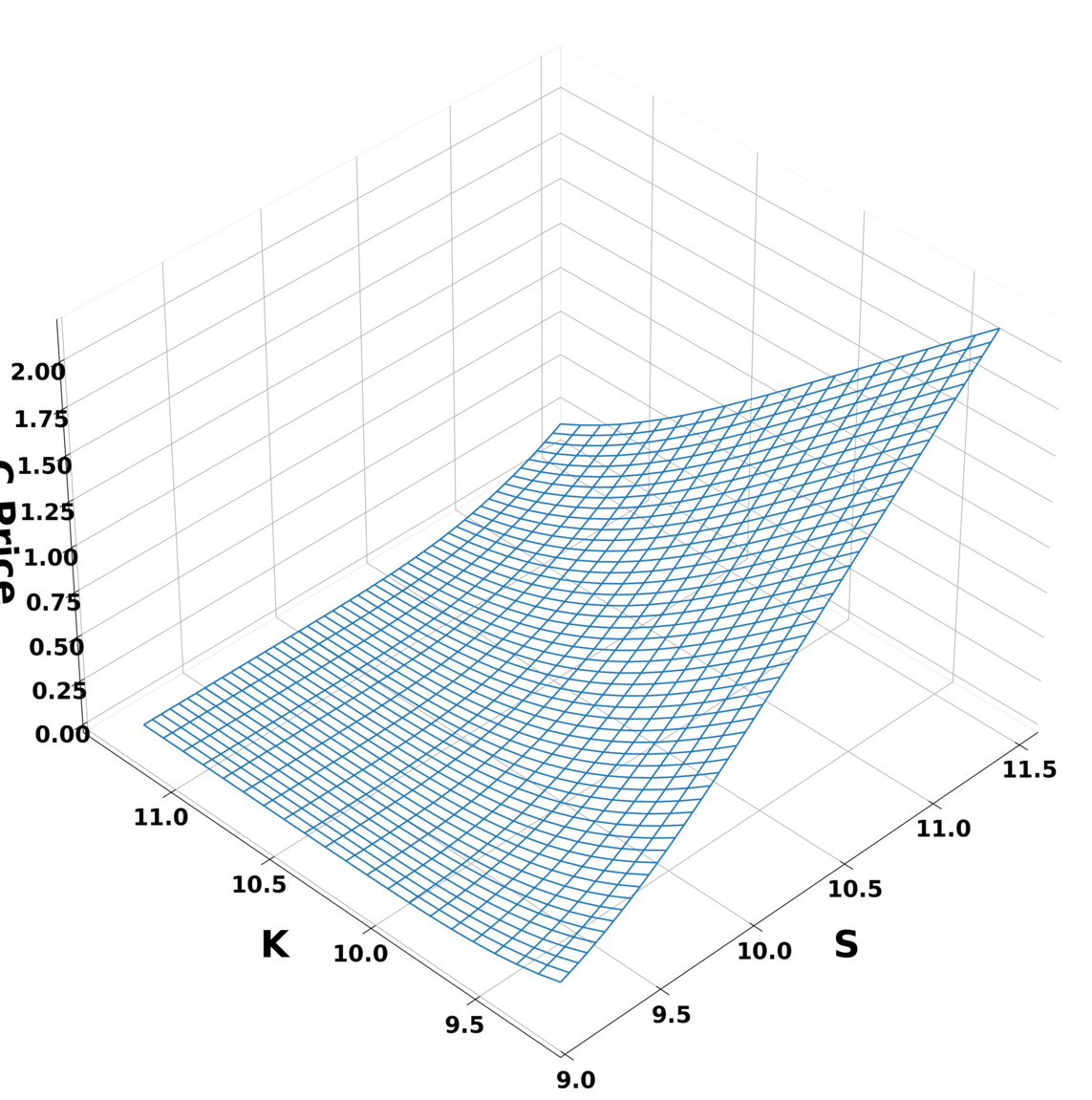}
        \caption{\texttt{GTEN} Price}
        \label{fig:gten_call_price_surface}
    \end{subfigure}\hfill
    \begin{subfigure}[t]{0.30\textwidth}
        \centering
        \includegraphics[width=\linewidth]{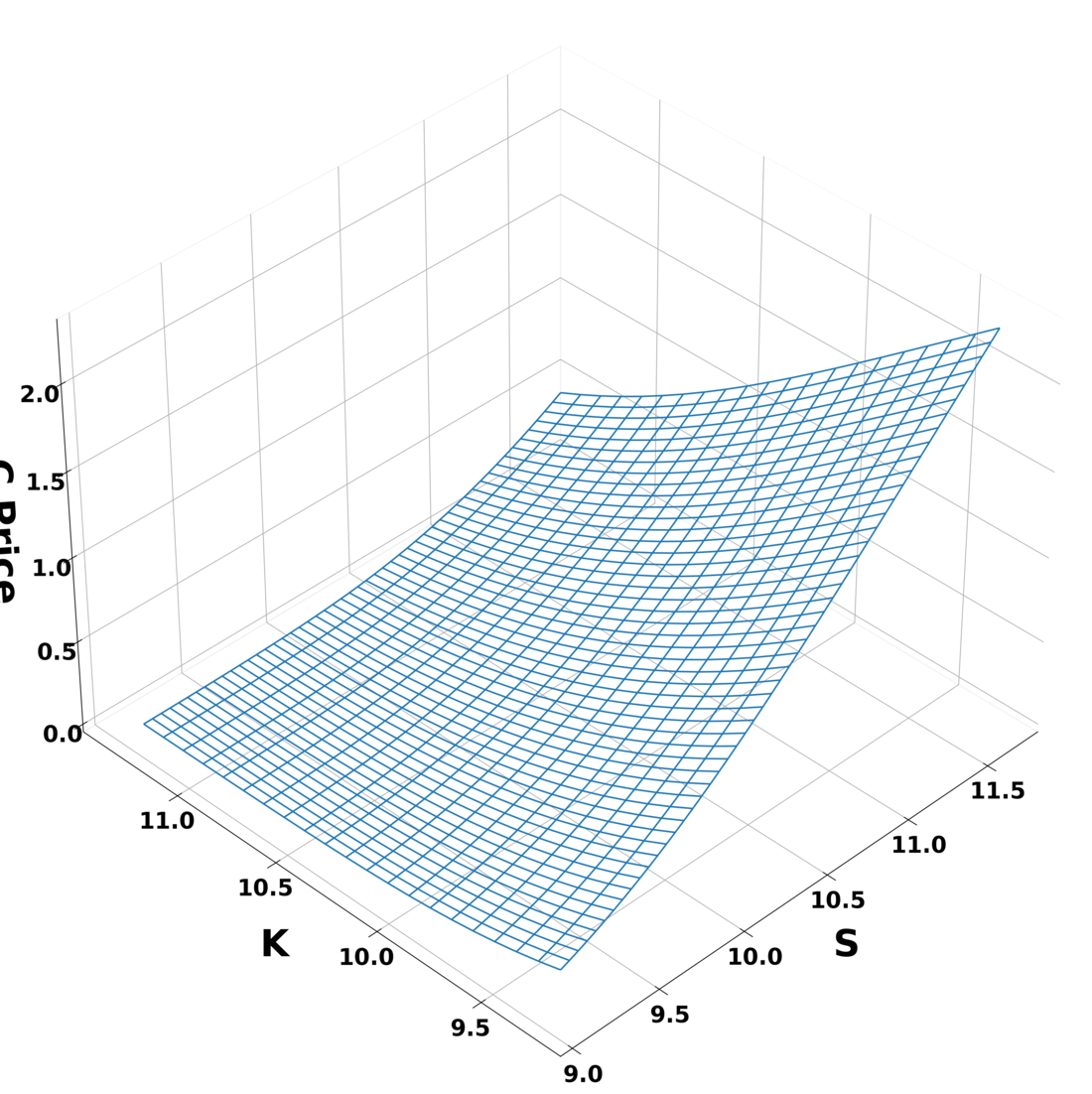}
        \caption{\texttt{TACO} Price}
        \label{fig:taco_call_price_surface}
    \end{subfigure}

    \vspace{0.6em}

    \begin{subfigure}[t]{0.3\textwidth}
        \centering
        \includegraphics[width=\linewidth]{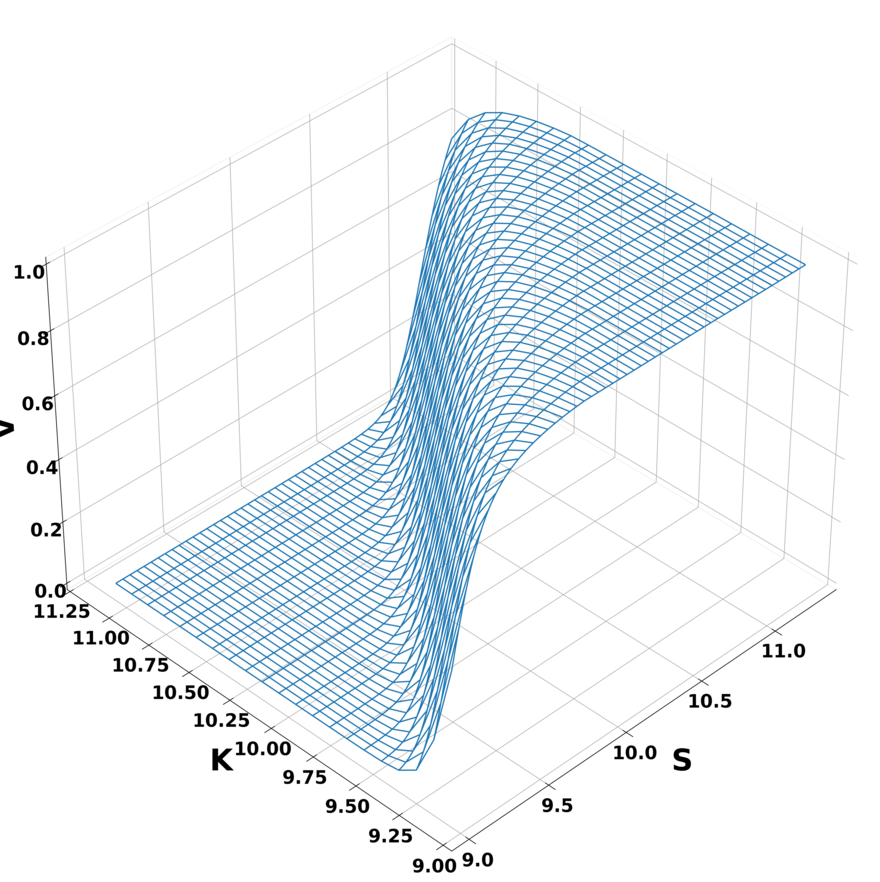}
        \caption{\texttt{COPL} $\Delta_{call}$}
        \label{fig:copl_call_delta_surface}
    \end{subfigure}\hfill
    \begin{subfigure}[t]{0.30\textwidth}
        \centering
        \includegraphics[width=\linewidth]{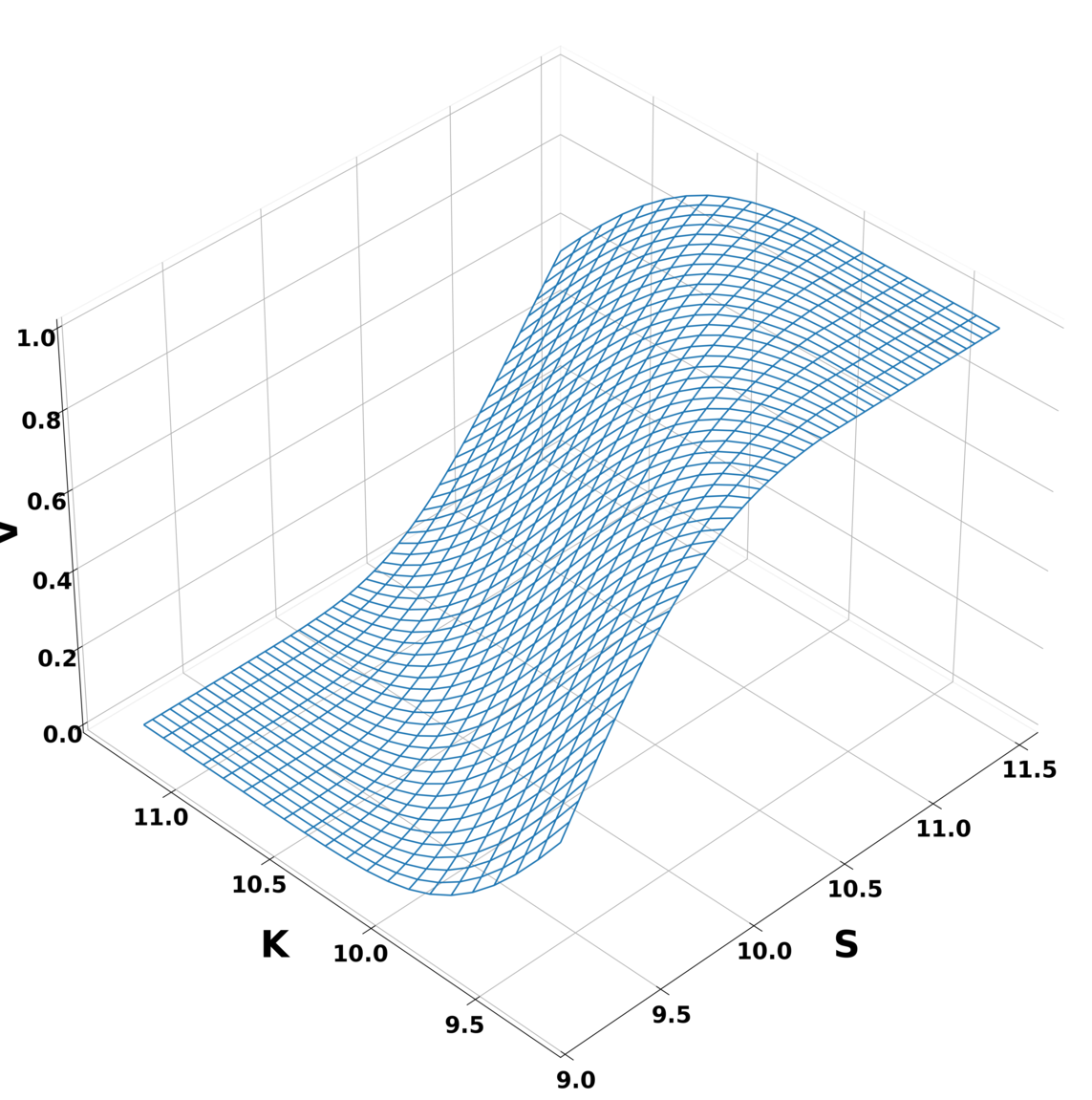}
        \caption{\texttt{GTEN} $\Delta_{call}$}
        \label{fig:gten_call_delta_surface}
    \end{subfigure}\hfill
    \begin{subfigure}[t]{0.30\textwidth}
        \centering
        \includegraphics[width=\linewidth]{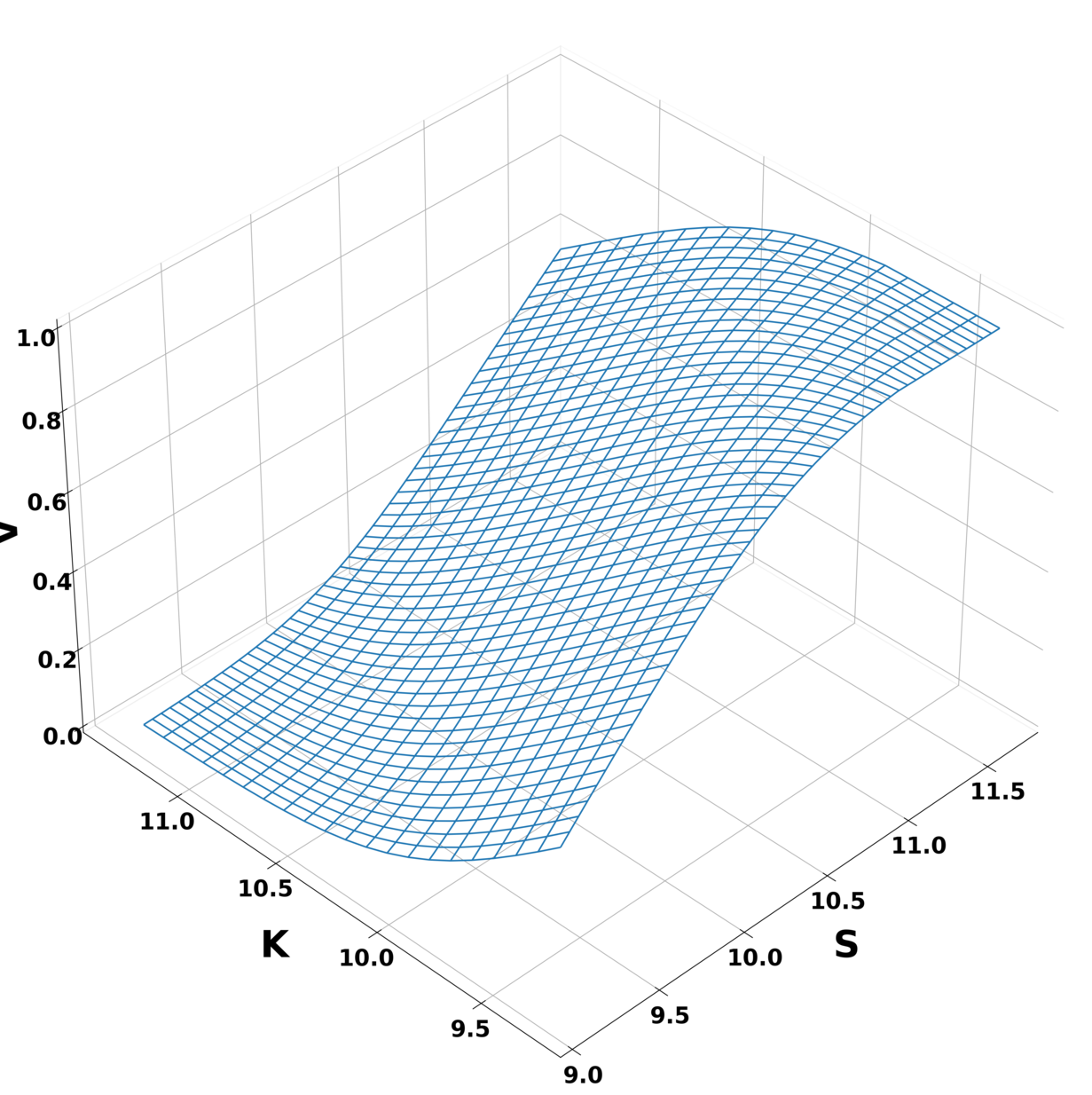}
        \caption{\texttt{TACO} $\Delta_{call}$}
        \label{fig:taco_call_delta_surface}
    \end{subfigure}

    \caption{FINN-implied European Call option surfaces for instruments without listed options (TTM $=0.36$). Top row: Call option prices. Bottom row: corresponding option deltas.}
    \label{fig:call_price_delta_surfaces_all}
\end{figure}

Figure~\ref{fig:call_price_delta_surfaces_all} reports the FINN-implied call price and delta surfaces (the corresponding put surfaces are provided in the Appendix; see Appendix~\ref{app:ipo}). For each instrument, the trained network is evaluated on a grid of forward moneyness $S/(Ke^{-rT})$ and time to maturity $T$, matching the experimental setup used in previous sections. The most recent observed spot price is used as the current underlying reference level for surface evaluation.

All reported surfaces are generated without access to any listed option prices and without assuming a parametric model for the underlying dynamics. The same training protocol is applied across instruments; observed differences in the resulting surfaces reflect differences in the underlying realized spot-price paths used during training.

\subsubsection{Discussion.}

This experiment demonstrates that FINN can produce economically coherent call and put price and delta surfaces even when no listed options exist for the underlying asset. The key takeaway is that FINN operates as a data-driven pricing operator: it learns a smooth mapping from $(S/(Ke^{-rT}), T)$ to option prices and sensitivities using only realized spot price transitions and a replication-based objective, without specifying or calibrating a parametric model for the underlying dynamics.

Across the three newly issued instruments, training is numerically stable and the learned surfaces exhibit the qualitative properties expected under no-arbitrage. Call prices are increasing and convex in $S$ with deltas confined to $[0,1]$, while put prices are decreasing in $S$ with deltas confined to $[-1,0]$. These shape properties arise naturally from the hedging-based loss together with the correct payoff behavior as $T \to 0$, rather than from explicit surface constraints or option price supervision.

From a practical standpoint, FINN delivers a complete and differentiable pricing surface over strikes and maturities, enabling downstream tasks such as stress testing, scenario analysis, and rapid valuation of customized contracts. Once trained, pricing new $(K,T)$ pairs is effectively instantaneous via a single forward pass, making the approach well suited as an initialization tool in settings where classical implied-volatility calibration is infeasible.

Overall, these results indicate that FINN provides a practical and economically grounded framework for initializing option prices, valuing bespoke derivatives, and supporting structured product design in markets where traditional option data or calibration methods are unavailable.

\section{Conclusion}
\label{sec:conclusion}

This paper introduced the \emph{Finance-Informed Neural Network (FINN)}, a principled learning framework for option pricing and hedging that resolves a long-standing trade-off between economic structure and statistical flexibility. By replacing price-based supervision with a self-supervised replication-consistency objective, FINN embeds no-arbitrage and dynamic hedging principles directly into the learning process. As a result, the model learns not only option prices, but also economically meaningful sensitivities that are directly usable for risk management.

From a theoretical perspective, we showed that minimizing the replication residual is equivalent, in the continuous-time limit, to enforcing the standard arbitrage-free pricing operator. Under mild regularity and approximation assumptions, FINN is consistent: vanishing replication loss implies convergence of both prices and Greeks to the arbitrage-free solution on compact domains. In incomplete markets, where prices are not uniquely defined, FINN naturally selects an economically admissible pricing functional determined by self-financing replication error minimization, providing a clear and operational interpretation of the learned solution.

Empirically, FINN performs robustly across a wide range of settings. In classical Black--Scholes environments, it accurately recovers analytical prices and hedge ratios. In stochastic volatility settings, including the Heston model, FINN remains stable and accurate despite the absence of closed-form Greeks. Fundamental no-arbitrage relationships, such as put--call parity, emerge endogenously without being imposed as explicit constraints. Beyond pricing, FINN supports advanced risk-management tasks: delta--gamma hedging based on FINN delivers improved hedging performance and reduced tail risk under volatility misspecification and transaction costs, particularly in stressed market regimes.

A key empirical insight of this work is that FINN learns a stable and transferable \emph{pricing geometry}. When applied to implied volatility surface reconstruction, FINN captures the structural shape of the surface more robustly than parametric stochastic volatility models whose calibrated parameters are assumed to be time-invariant. Comparisons with Heston calibration reveal that a learned pricing operator can adapt more effectively out of sample, reducing systematic bias while preserving no-arbitrage consistency. This highlights the potential of FINN as a flexible alternative to traditional calibration pipelines in option markets.

More generally, this capability extends beyond settings where option markets exist. We show that FINN can be trained directly on historical spot price data to learn arbitrage-consistent option prices and sensitivities for assets without listed options. In such environments, FINN does not rely on implied volatilities, parametric stochastic dynamics, or calibration to option prices. Instead, it constructs a pricing operator grounded in observed price transitions and replication consistency, enabling principled initialization of option prices for newly issued assets, illiquid securities, and bespoke financial instruments where traditional option pricing methodologies are infeasible.

Conceptually, FINN reframes option pricing as the problem of learning a \emph{pricing operator} rather than fitting prices. By aligning machine learning with the economic logic of replication and risk control, the framework shifts the focus from predictive accuracy alone to decision relevance, robustness, and governance. This perspective is particularly important in financial environments where model outputs directly inform trading, capital allocation, and regulatory reporting.

Several directions for future research naturally follow. First, extending FINN to multi-asset and high-dimensional settings would enable pricing and hedging of complex derivatives and portfolios. Second, incorporating jumps, stochastic interest rates, or limits on borrowing and financing into the replication objective would further reflect real market conditions. Third, integrating FINN with online or continual learning schemes could allow dynamic adaptation to evolving market conditions while preserving arbitrage discipline. Finally, fundamental connections between FINN and equilibrium-based asset pricing models may shed light on how learned pricing operators relate to investor preferences and market microstructure.

Overall, FINN provides a scalable and economically grounded approach to option pricing that bridges theory-driven finance and modern machine learning. By embedding financial principles directly into the learning objective, it offers a practical path toward trustworthy, interpretable, and robust AI systems for financial decision-making in practice.

\appendix

\section{Proofs of Main Results}
\label{app:proofs}

\subsection*{Proof of Proposition \ref{prop:existence}}

\proof{Proof.}
The proof proceeds in two steps: first, establishing the probabilistic representation via the Fundamental Theorem of Asset Pricing, and second, establishing the analytical regularity required for the existence of hedging sensitivities.

\textit{Probabilistic Representation (Existence and Uniqueness).}
We assume the market model satisfies Assumption~\ref{ass:regularity_main} (completeness and absence of arbitrage). By the First Fundamental Theorem of Asset Pricing \citep{delbaen1994general}, the absence of arbitrage implies the existence of an equivalent local martingale measure (ELMM), denoted by $\mathbb{Q}$, under which the discounted asset price processes are local martingales. Furthermore, the assumption of market completeness implies that this measure $\mathbb{Q}$ is unique \citep{harrison1981martingales}.

Consequently, for any European contingent claim with a square-integrable payoff $f(S_T)$, the unique arbitrage-free price process $g^*(t, S_t)$ is given by the conditional expectation of the discounted payoff under $\mathbb{Q}$:
\begin{equation}
    g^*(t,S_t) = \mathbb{E}^{\mathbb{Q}}\left[e^{-\int_t^T r_u du} f(S_T) \mid \mathcal{F}_t\right].
\end{equation}
This establishes the existence and uniqueness of the pricing functional in the probabilistic sense.

\textit{Analytical Regularity ($C^{1,2}$ Smoothness).}
To establish the smoothness required for the existence of Delta ($\partial_s g^*$) and Gamma ($\partial_s^2 g^*$), we invoke the Feynman-Kac representation theorem. Under Assumption~\ref{ass:regularity_main}, specifically that the coefficients of the underlying SDE (drift $\mu$ and diffusion $\sigma$) satisfy Lipschitz continuity and linear growth conditions, and that the payoff function $f$ satisfies polynomial growth constraints, the conditional expectation $g^*(t,s)$ is identified as the unique viscosity solution to the associated linear parabolic Partial Differential Equation (PDE):
\begin{equation}
    \frac{\partial g^*}{\partial t} + \mathcal{L}g^* - r g^* = 0,
\end{equation}
subject to the terminal condition $g^*(T,s) = f(s)$, where $\mathcal{L}$ is the infinitesimal generator of the underlying process under $\mathbb{Q}$.

From standard theory on parabolic differential equations \citep{friedman1975stochastic}, if the coefficients are sufficiently smooth and the operator is uniformly elliptic on the compact domain $D$, interior regularity estimates guarantee that $g^*(t, \cdot)$ is continuously differentiable once in time and twice in space. Thus, $g^* \in C^{1,2}(D)$. This smoothness ensures that the sensitivities $\Delta_{g^*} := \partial_s g^*$ and $\Gamma_{g^*} := \partial_s^2 g^*$ are well-defined continuous functions, validating the target gradients for the Finance-Informed Neural Network.
\endproof

\subsection*{Proof of Proposition \ref{prop:residual_pde}}

\proof{Proof.}
Let the underlying asset price $S_t$ follow a generic diffusion process on the probability space $(\Omega, \mathcal{F}, \mathbb{Q})$:

\begin{equation}
    dS_t = \mu(t, S_t) S_t dt + \sigma(t, S_t) S_t dW_t,
\end{equation}

where $W_t$ is a standard Brownian motion under the risk-neutral measure $\mathbb{Q}$, and $\mu, \sigma$ satisfy Assumption~\ref{ass:regularity_main}.

Let $g^\theta(t, S_t)$ be the pricing function approximated by the neural network. Assuming $g^\theta \in C^{1,2}$ (which is satisfied by the smooth activation functions of the network), we apply Itô's Lemma to expand the differential $dg^\theta$:

\begingroup
\small
\begin{equation}
    dg^\theta = \left( \frac{\partial g^\theta}{\partial t} + \mu S_t \frac{\partial g^\theta}{\partial s} + \frac{1}{2}\sigma^2 S_t^2 \frac{\partial^2 g^\theta}{\partial s^2} \right) dt + \sigma S_t \frac{\partial g^\theta}{\partial s} dW_t.
    \label{eq:ito_expansion}
\end{equation}
\endgroup

Now, consider the self-financing replicating portfolio $\Pi_t$ consisting of $\Delta_t$ units of the underlying asset and the remainder in a risk-free bank account growing at rate $r$. The dynamics of this portfolio are given by:

\begin{equation}
    d\Pi_t = \Delta_t dS_t + r(\Pi_t - \Delta_t S_t) dt.
\end{equation}

Substituting the dynamics of $dS_t$ and assuming the portfolio value tracks the model price ($\Pi_t = g^\theta(t, S_t)$) at the start of the interval, we obtain:

\begin{equation}
    d\Pi_t = \left( \Delta_t \mu S_t + r g^\theta - r \Delta_t S_t \right) dt + \Delta_t \sigma S_t dW_t.
    \label{eq:portfolio_dynamics}
\end{equation}

The \textit{replication residual} in continuous time is defined as the instantaneous difference between the change in the model price and the change in the hedging portfolio: $\mathcal{R}_t dt := dg^\theta - d\Pi_t$. Subtracting \eqref{eq:portfolio_dynamics} from \eqref{eq:ito_expansion} and grouping terms by $dt$ and $dW_t$:

\begingroup
\small
\begin{align}
    dg^\theta - d\Pi_t &= \underbrace{\left[ \frac{\partial g^\theta}{\partial t} + r S_t \frac{\partial g^\theta}{\partial s} + \frac{1}{2}\sigma^2 S_t^2 \frac{\partial^2 g^\theta}{\partial s^2} - r g^\theta \right]}_{\text{Black-Scholes Operator } \mathcal{L}_{BS} g^\theta} dt \nonumber \\
    &\quad + (\mu - r)S_t \left( \frac{\partial g^\theta}{\partial s} - \Delta_t \right) dt \nonumber \\
    &\quad + \sigma S_t \left( \frac{\partial g^\theta}{\partial s} - \Delta_t \right) dW_t.
    \label{eq:residual_expansion}
\end{align}
\endgroup

Note that in the first bracket, the physical drift $\mu$ cancels out with the portfolio drift term if and only if $\Delta_t = \partial_s g^\theta$, leaving the risk-free rate $r$.

We now prove the equivalence:

\begin{itemize}
    \item $(\Rightarrow)$ \textbf{Residual Vanishes implies PDE:}
    If the replication residual is zero almost surely ($dg^\theta - d\Pi_t = 0$), then by the uniqueness of the Itô decomposition, both the drift (the $dt$ term) and the diffusion (the $dW_t$ term) must vanish independently.\\
    1. The vanishing of the diffusion term implies:
    $$ \sigma S_t \left( \frac{\partial g^\theta}{\partial s} - \Delta_t \right) = 0 \implies \Delta_t = \frac{\partial g^\theta}{\partial s}. $$
    This confirms the neural network learns the correct Delta hedging strategy.\\
    2. Substituting $\Delta_t = \partial_s g^\theta$ into the drift term eliminates the $(\mu-r)$ component. For the remaining drift to be zero, we requires:
    $$ \frac{\partial g^\theta}{\partial t} + r S_t \frac{\partial g^\theta}{\partial s} + \frac{1}{2}\sigma^2 S_t^2 \frac{\partial^2 g^\theta}{\partial s^2} - r g^\theta = 0. $$
    This is precisely the no-arbitrage Partial Differential Equation (Black-Scholes PDE).
    \item $(\Leftarrow)$ \textbf{PDE implies Residual Vanishes:}
    Conversely, if $g^\theta$ satisfies the pricing PDE and we select the hedging ratio $\Delta_t = \partial_s g^\theta$, then by direct substitution into \eqref{eq:residual_expansion}, both the $dt$ and $dW_t$ coefficients become zero, rendering the residual identically zero.
\end{itemize}

Thus, minimizing the replication error is mathematically equivalent to solving the pricing PDE.
\endproof

\subsection*{Proof of Theorem \ref{thm:consistency}}

\proof{Proof.}
\textit{Part (i): Density of Smooth Neural Networks.}
This result follows from the Universal Approximation Theorem for derivatives. While standard theorems guarantee uniform approximation of the function value, extensions to smooth activation functions (e.g., swish or tanh) guarantee simultaneous approximation of the function and its derivatives on compact sets. We refer to \citet{hornik1990universal} and \citet{pinkus1999approximation} for foundational results on the density of neural networks in Sobolev and $C^k$ spaces. Since $g^*$ represents a valid economic pricing surface with bounded derivatives on $D$, it lies within the hypothesis space of our network architecture.

\textit{Part (ii): Convergence to the Unique Solution.}
As established in Proposition \ref{prop:residual_pde}, the continuous-time limit of the replication residual is the Black-Scholes operator (or its stochastic volatility equivalent). As the time step $\Delta t \to 0$, the discrete loss function $\mathcal{L}(\theta)$ converges to the integral of the squared PDE residual.
Let $\bar{g}$ be a limit point of the sequence $\{g^{\theta_m}\}$. Since $\mathcal{L}(\theta_m) \to 0$, $\bar{g}$ must satisfy the homogeneous PDE $\mathcal{L}_{BS} \bar{g} = 0$ almost everywhere.
Given that the boundary conditions are fixed by the payoff function, and invoking the uniqueness of classical solutions for linear parabolic equations \citep{friedman1975stochastic}, it follows that $\bar{g} \equiv g^*$. Finally, the interior regularity of parabolic operators implies that convergence in the pricing norm entails convergence of the spatial derivatives \citep{ladyzhenskaya1968linear}, ensuring that the learned Delta $\partial_s g^{\theta_m}$ converges to the true Delta $\partial_s g^*$.
\endproof

\subsection*{Proof of Proposition \ref{prop:bs_equiv}}

\proof{Proof.}
Let the dynamics of the underlying asset $S_t$ under the risk-neutral measure $\mathbb{Q}$ be given by the geometric Brownian motion:

\begin{equation}
    S_t = r S_t dt + \sigma S_t dW_t.
\end{equation}

We assume the neural network approximation $g^\theta(t, S_t)$ is of class $C^{1,2}([0,T) \times \mathbb{R}_+)$. By It\^o's Lemma, the differential of the option price process is:

\begin{equation}
    dg^\theta = \left( \frac{\partial g^\theta}{\partial t} + \frac{1}{2}\sigma^2 S_t^2 \frac{\partial^2 g^\theta}{\partial s^2} \right) dt + \frac{\partial g^\theta}{\partial s} dS_t.
\end{equation}

(Note: We retain $dS_t$ in the expansion to facilitate cancellation in the portfolio dynamics).

Consider the portfolio $\Pi_t = -g^\theta(t, S_t) + \alpha_t S_t$. By the self-financing condition, the change in the portfolio value $d\Pi_t$ depends only on the changes in the constituent asset prices:

\begin{equation}
    d\Pi_t = -dg^\theta(t, S_t) + \alpha_t dS_t.
\end{equation}

Substituting the expression for $dg^\theta$ and collecting terms involving $dS_t$:

\begingroup
\small
\begin{equation}
    d\Pi_t = -\left( \frac{\partial g^\theta}{\partial t} + \frac{1}{2}\sigma^2 S_t^2 \frac{\partial^2 g^\theta}{\partial s^2} \right) dt + \left( \alpha_t - \frac{\partial g^\theta}{\partial s} \right) dS_t.
\end{equation}
\endgroup

By hypothesis, the hedging strategy is defined as $\alpha_t = \partial_s g^\theta(t, S_t)$. Consequently, the coefficient of the stochastic term $dS_t$ (and thus $dW_t$) vanishes, rendering the portfolio locally risk-free:

\begin{equation}
    d\Pi_t = -\left( \frac{\partial g^\theta}{\partial t} + \frac{1}{2}\sigma^2 S_t^2 \frac{\partial^2 g^\theta}{\partial s^2} \right) dt.
    \label{eq:portfolio_change}
\end{equation}

The proposition states that the portfolio earns the risk-free rate, i.e., $d\Pi_t = r \Pi_t dt$. Substituting the definition of $\Pi_t$:

\begin{equation}
    d\Pi_t = r \left( -g^\theta + S_t \frac{\partial g^\theta}{\partial s} \right) dt.
    \label{eq:risk_free_growth}
\end{equation}

Equating \eqref{eq:portfolio_change} and \eqref{eq:risk_free_growth}, and dividing by $dt$:

\begin{equation}
    -\frac{\partial g^\theta}{\partial t} - \frac{1}{2}\sigma^2 S_t^2 \frac{\partial^2 g^\theta}{\partial s^2} = -r g^\theta + r S_t \frac{\partial g^\theta}{\partial s}.
\end{equation}

Rearranging all terms to the left-hand side yields the fundamental Black-Scholes partial differential equation:

\begin{equation}
    \frac{\partial g^\theta}{\partial t} + r S_t \frac{\partial g^\theta}{\partial s} + \frac{1}{2}\sigma^2 S_t^2 \frac{\partial^2 g^\theta}{\partial s^2} - r g^\theta = 0.
\end{equation}

This establishes the logical equivalence: the portfolio grows at the risk-free rate if and only if the pricing function $g^\theta$ satisfies the Black-Scholes operator.

\endproof

\section{Geometric Brownian Motion Implementation}
\label{appendix:GBM}

To demonstrate FINN's versatility in option pricing across different underlying asset models, we first tested our framework using the Geometric Brownian Motion (GBM) model. While FINN can accommodate various stochastic processes and empirical market data, GBM provides an ideal starting point due to its analytical tractability and widespread use in financial modeling. The GBM model describes asset price dynamics through the following stochastic differential equation:
\begin{equation}
d S(t)=\mu S(t) d t+\sigma S(t) d W(t)
\end{equation}
where $\mu$ represents the expected return (drift), $\sigma$ denotes the asset's volatility, and $dW(t)$ represents the increment of a Wiener process, capturing random price fluctuations. For numerical implementation of GBM, we must transform the continuous-time stochastic differential equation into a discrete-time approximation suitable for computer simulation. We accomplish this using the Monte Carlo method with Euler-Maruyama discretization \cite{VomScheidt1994}, a fundamental numerical scheme for approximating solutions to stochastic differential equations. Given a time step $\Delta t$, the discretized equation takes the form \cite{VomScheidt1994}:
\begin{equation}
\label{eq:GBM_sim}
S_{t+\Delta t}=S_t \exp \left((\mu-\dfrac{1}{2} \sigma^2) \Delta t+\sigma \sqrt{\Delta t} Z_t\right)
\end{equation}
This equation provides a method to calculate the asset price at the next time step ($S_{t+\Delta t}$) given the current price ($S_t$). The exponential form ensures that asset prices remain positive, which is a crucial property of stock prices. The equation \ref{eq:GBM_sim} consists of two key components: a deterministic component $(\mu-\dfrac{1}{2} \sigma^2) \Delta t$ that adjusts for the drift ($\mu$) of the process, with the -(1/2)$\sigma^2$ term serving as a necessary correction to ensure the expected return matches the continuous-time GBM model; and a random component $\sigma \sqrt{\Delta t} Z_t$, where $Z_t$ represents a random draw from a standard normal distribution, with $\sqrt{\Delta t}$ scaling the random shock appropriately for the time step size and the volatility parameter $\sigma$ determining the magnitude of these random fluctuations.

This discretization allows us to generate sample paths by iteratively computing price changes over small time intervals. For example, to simulate a price path over one year with daily steps, we would set $\Delta t = 1/252$ (assuming 252 trading days) and repeatedly apply this equation, drawing a new random normal variable $Z_t$ at each step.

Under the assumptions of GBM, the Black-Scholes model provides analytical solutions for European option pricing. For a European call option $C$ (which gives the right to buy) and a put option $P$ (which gives the right to sell), the prices are given by:
\begingroup
\small
\begin{align}
C(S,K,t,T,\sigma,r)
&= S\,N(d_1)-K e^{-r(T-t)}\,N(d_2), \label{eq:bs_call}\\
P(S,K,t,T,\sigma,r)
&= K e^{-r(T-t)}\,N(-d_2)-S\,N(-d_1). \label{eq:bs_put}
\end{align}
\endgroup
These formulas incorporate several key parameters: the current stock price ($S$), strike price ($K$), current time ($t$), maturity time ($T$), volatility ($\sigma$), and risk-free rate ($r$). The function $N(\cdot)$ represents the cumulative distribution function of the standard normal distribution. The terms $d_1$ and $d_2$ are defined as:
\begingroup
\small
\begin{align}
d_1
&= \frac{\ln(S/K)+\left(r+\tfrac{1}{2}\sigma^2\right)(T-t)}{\sigma\sqrt{T-t}},
\label{eq:bs_d1}\\
d_2
&= d_1-\sigma\sqrt{T-t}.
\label{eq:bs_d2}
\end{align}
\endgroup
To manage the risk associated with option positions, traders rely on hedge ratios, particularly Delta ($\Delta$). Delta measures the rate of change in the option price with respect to changes in the underlying asset price, mathematically expressed as:
\begin{equation}
\Delta_{\text{call}}=\dfrac{\partial C}{\partial S},
\Delta_{\text{put}}=\dfrac{\partial P}{\partial S}
\end{equation}
To rigorously evaluate our FINN model's performance, we conduct extensive testing using thousands of simulated scenarios. We compare FINN's predictions against these Black-Scholes analytical solutions using two comprehensive metrics: Mean Absolute Deviation (MAD) and Mean Square Error (MSE). These metrics assess both the accuracy of option prices and hedge ratios. This systematic evaluation framework enables us to verify FINN's capability to learn and reproduce the fundamental pricing relationships that the Black-Scholes model captures mathematically.

\section{The Heston Stochastic Volatility Implementation}
\label{appendix:Heston}
The Heston stochastic volatility model \cite{heston1993} addresses fundamental limitations of the Black-Scholes framework by incorporating time-varying volatility dynamics. This extension enables the model to capture empirically observed market phenomena, particularly the non-constant nature of asset price volatility and the leverage effect—the negative correlation between asset returns and volatility changes. By allowing volatility to evolve according to its own stochastic process, the Heston model provides a more sophisticated mathematical framework for option pricing that aligns with observed market behavior.

In this model, both the asset price and its volatility evolve according to coupled stochastic processes. The asset price $S_t$ follows the process:
\begin{equation}
d S_t=\mu S_t d t+\sqrt{\nu_t} S_t d W_t^S
\end{equation}
What makes this model special is that the volatility itself ($\sqrt{\nu_t}$) follows an Ornstein-Uhlenbeck process:
\begin{equation}
d\sqrt{\nu_t}=-\theta \sqrt{\nu_t} d t+\xi d W^\nu
\end{equation}
Through application of Itô's lemma, we can show that the variance $\nu_t$ follows a Cox-Ingersoll-Ross (CIR) process \cite{CoxIngersollRoss1985}:
\begin{equation}
d \nu_t=\kappa(\theta-\nu_t)d t+\xi\sqrt{\nu_t} d W_t^\nu
\end{equation}
The two Wiener processes $W_t^S$ and $W_t^\nu$ are correlated with correlation coefficient $\rho$, allowing the model to capture the leverage effect observed in equity markets.
\begin{equation}
dW_t^S dW_t^\nu = \rho dt
\end{equation}
This equation shows how the two Brownian motions are correlated, with $\rho = -0.7$ in our implementation. This negative correlation captures the leverage effect: when asset prices decrease, volatility tends to increase, and vice versa.

Overall, the Heston model is characterized by five key parameters:
\begin{itemize}
    \item Initial variance ($\nu_0$): The starting point for the variance process
    \item Long-term variance ($\theta$): The level around which the variance tends to oscillate
    \item Mean reversion rate ($\kappa$): How quickly the variance returns to its long-term average
    \item Volatility of volatility ($\xi$): How much the variance itself fluctuates (volvol)
    \item Correlation ($\rho$): Between asset returns and variance changes
\end{itemize}

To ensure the variance remains positive, the parameters must satisfy \cite{little_heston}:
\begin{equation}
2 \kappa \theta > \xi^2
\end{equation}
Under risk-neutral pricing, the European call option value is:
\begin{equation}
C_0=S_0\cdot \Pi_1- e^{-rT} K \cdot \Pi_2
\end{equation}
where $\Pi_1$ and $\Pi_2$ are risk-neutral probabilities computed through characteristic functions:
\begingroup
\small
\begin{align}
\Pi_1
&= \frac{1}{2}+\frac{1}{\pi}\int_{0}^{\infty}
\Re\!\left[
\frac{e^{-i w \ln K}\,\Psi_{\ln S_T}(w-i)}
{i w\,\Psi_{\ln S_T}(-i)}
\right]dw, \label{eq:pi1_cf}\\
\Pi_2
&= \frac{1}{2}+\frac{1}{\pi}\int_{0}^{\infty}
\Re\!\left[
\frac{e^{-i w \ln K}\,\Psi_{\ln S_T}(w)}
{i w}
\right]dw. \label{eq:pi2_cf}
\end{align}
\endgroup

The option price in the Heston model requires computing a characteristic function $\Psi$, which encapsulates the probabilistic behavior of future asset prices under stochastic volatility. This characteristic function is constructed using several interrelated auxiliary functions, each playing a specific role in capturing different aspects of the model's dynamics. The complete expression for the characteristic function takes the form \cite{heston1993}:
\begin{equation}
\Psi_{l n S_T}(w)= e ^{\Big[C(t, w) \cdot \theta + D(t, w) \cdot \nu_0+i \cdot w \cdot \ln \big(S_0 \cdot e ^{r t}\big)\Big]}
\end{equation}
where $C(t,w)$ captures the long-term variance effects, $D(t,w)$ incorporates the current variance level, and the exponential term with $\ln(S_0 \cdot e^{rt})$ accounts for the deterministic drift in asset prices.

In our experimental implementation, we used carefully chosen parameters that reflect typical market conditions:
\begingroup
\small
\begin{align}
r &= 0, && \text{(interest rate)} \notag\\
dt &= \frac{1}{250}, && \text{(Euler discretization step)} \notag\\
\xi &\in \{0.125,\,0.15,\,0.175\}, && \text{(vol-of-vol)} \notag\\
\rho &= -0.7, && \text{(correlation)} \notag\\
\kappa &= 1.25, && \text{(mean-reversion speed)} \notag\\
\theta &= 0.0225, && \text{(long-run variance)} \notag\\
\nu_0 &= 0.0225, && \text{(initial variance).} \notag
\end{align}
\endgroup
These parameter choices ensure realistic market dynamics while maintaining numerical stability in our simulations.

\newpage
\section{Pseudo Code}\label{pseudo_code}
\begin{algorithm}
\caption{FINN: Delta and Delta--Gamma Hedging for European Options}\label{alg:finn_hedge}
\vskip6pt
\begin{algorithmic}[1]
\Require Underlying prices $S_t,S_{t+1}$; strike $K$; time-to-maturity $\tau_t,\tau_{t+1}$; rate $r$.
\Require ATM hedge instrument values $C_{\mathrm{ATM},t},C_{\mathrm{ATM},t+1},\Delta_{\mathrm{ATM},t},\Gamma_{\mathrm{ATM},t}$ (only if Delta--Gamma hedging).
\Require FINN pricing operator $g^\theta(\cdot)$.
\State \textbf{Inputs:} $X_t \gets \bigl(S_t/(K e^{-r\tau_t}),\,\tau_t\bigr)$; \;
$X_{t+1} \gets \bigl(S_{t+1}/(K e^{-r\tau_{t+1}}),\,\tau_{t+1}\bigr)$.
\State \textbf{Price:} $C_t \gets g^\theta(X_t)$; \; $C_{t+1} \gets g^\theta(X_{t+1})$.
\State \textbf{Greeks:} $\Delta_t \gets \partial_S g^\theta(X_t)$; \; $\Gamma_t \gets \partial^2_{S} g^\theta(X_t)$.
\If{\textbf{Delta hedging}}
    \State $\Pi_t \gets -C_t + \Delta_t S_t$; \quad $\Pi_{t+1} \gets -C_{t+1} + \Delta_t S_{t+1}$.
\Else \Comment{\textbf{Delta--Gamma hedging using ATM instrument}}
    \State $\beta_t \gets \Gamma_t/\Gamma_{\mathrm{ATM},t}$; \quad $\alpha_t \gets \Delta_t-\beta_t\Delta_{\mathrm{ATM},t}$.
    \State $\Pi_t \gets -C_t + \alpha_t S_t + \beta_t C_{\mathrm{ATM},t}$.
    \State $\Pi_{t+1} \gets -C_{t+1} + \alpha_t S_{t+1} + \beta_t C_{\mathrm{ATM},t+1}$.
\EndIf
\State \textbf{Objective:} minimize sample average approximation (SAA) of the mean squared hedging residual $\mathbb{E}\!\left[(\Pi_{t+1}-e^{r(\tau_t-\tau_{t+1})}\Pi_t)^2\right]$ over $\theta$.
\end{algorithmic}
\end{algorithm}

\section{Numerical Example of a Network Forward Pass}
\label{app:forward_pass}
To demonstrate the practical implementation of the trained neural network, we provide a step-by-step numerical example. The network architecture consists of two hidden layers with 50 neurons each and a softplus activation on the output layer, which produces a normalized option price ($C/K$). The input vector at time t ($I_t$), is two-dimensional, consisting of the moneyness ratio ($S/K$) and the time to maturity ($T$).

\subsection{Example of Loss Calculation}
We first illustrate how FINN computes and minimizes the delta-hedging loss for a European call option in a GBM environment.

\paragraph{Initial Market Conditions:}
\begin{itemize}
    \item Current stock price ($S_t$): \$120
    \item Strike price ($K$): \$100
    \item Risk-free rate ($r$): 0\%
    \item Time to maturity ($\tau$): 0.24 years
\end{itemize}

\noindent Consider a scenario where the stock price increases by 1 unit, such that $S_{t+1}=\$121$. Let the FINN model, $g^\theta$, estimate a hedge ratio $\Delta = \partial_S g^\theta_t(I_t) = 0.5$. The change in the value of the stock position is therefore $\Delta \cdot (S_{t+1}-S_t) = 0.5 \cdot (121-120) = 0.5$.

If the FINN model outputs a price of $g^\theta_t(I_t) = 10$ at time $t$ and $g^\theta_{t+1}(I_{t+1}) = 11$ at time $t+1$, the change in the option's value is $1$. The squared hedging error for this single step is:
\(
\text{Loss} = [\Delta \cdot (S_{t+1}-S_t) - (g^\theta_{t+1}(I_{t+1}) - g^\theta_t(I_t))]^2 = [0.5 - 1]^2 = 0.25.
\)
The training objective is to minimize the expectation of this squared error over many such steps, as shown in the one-period framework in Figure \ref{fig:two_period_framework}.

\begin{figure}[H]
    \begin{center}
    \begin{tikzpicture}[
    node distance=1.6cm, 
    scale=0.7, 
    every node/.style={
        transform shape, 
        draw, 
        circle, 
        minimum size=0.7cm,
        align=center
    }, 
    every path/.style={
        draw, 
        -{Latex[length=1.8mm]},
        shorten >=2pt,
        shorten <=2pt
    }
]
\node (S_t) at (0,0) {$S_t=120$};
\node[right=2cm, above of=S_t] (S_t1_up) {$S_{t+1}=121$};
\node[right=2cm, below of=S_t] (S_t1_down) {$S_{t+1}=119$};
\draw (S_t) -- (S_t1_up);
\draw (S_t) -- (S_t1_down);

\node[right=4cm of S_t] (I_t) {$I_t(\theta)=10$};
\node[right=2cm, above of=I_t] (I_t1_up) {$I_{t+1}(\theta)=11$};
\node[right=2cm, below of=I_t] (I_t1_down) {$I_{t+1}(\theta)=9$};
\node[rectangle, below=0.8cm of I_t] (Delta_t) {$\Delta_t=0.5$};
\draw (I_t) -- (I_t1_up);
\draw (I_t) -- (I_t1_down);

\node[right=4cm of I_t] (L_t) {$L_t(\theta)$};
\node[right=2cm, above of=L_t] (L_t1_up) {$L_{t+1}(\theta)$};
\node[right=2cm, below of=L_t] (L_t1_down) {$L_{t+1}(\theta)$};
\draw (L_t) -- (L_t1_up);
\draw (L_t) -- (L_t1_down);

\node[below=0.8cm of L_t, draw=none] (Loss)
{$L_\theta(t)=\sqrt{(S_{t+1}-S_t)\Delta_t-(I_{t+1}-I_t)^2}$};
\end{tikzpicture}

    \end{center}
    \vspace{-60pt}
    \caption{A one-period illustrative example of the loss calculation.}
    \label{fig:two_period_framework}
\end{figure}
\noindent Figure \ref{fig:two_period_framework} illustrates the loss calculation in a simple one-step scenario. At time $t+1$, the stock price can either move up or down, leading to two potential outcomes for the value of the option. For each outcome, the FINN model predicts a new option price (for instance, increasing to \$11 or decreasing to \$9). The loss is then calculated as the squared hedging error: the mismatch between the change in value of the hedged stock position and the change in the option's predicted price.

\subsection{Example of Price Calculation}
We now demonstrate a full forward pass using a trained two-layer neural network over 250 epochs under the following conditions \href{https://github.com/lixuanze/FINN/blob/main/Demo.ipynb}{\textcolor{blue}{\underline{[Demo]]}}}:
\begin{itemize}
    \item Geometric Brownian Motion environment
    \item Volatility ($\sigma$): 0.125
    \item Initial stock price ($S_0$): \$100
\end{itemize}

\paragraph{Network Architecture: }
\begin{itemize}
    \item Two hidden layers, each containing 50 neurons with $\tanh$ activation.
    \item Output layer producing normalized option prices ($C/K$) with a \textit{softplus} activation.
\end{itemize}
The network takes a two-dimensional input vector $x = [S/K, T]^T$. The layer-by-layer transformations are given by:
\begin{align}
    a^{[1]} &= \tanh(W^{[1]}x + b^{[1]}) \in \mathbb{R}^{50} \\
    a^{[2]} &= \tanh(W^{[2]}a^{[1]} + b^{[2]}) \in \mathbb{R}^{50} \\
    \frac{C}{K} &= \text{softplus}(W^{[3]}a^{[2]} + b^{[3]})
\end{align}

\paragraph{Numerical Computation:}
We validate the network's performance on an in-the-money European call option where the moneyness ratio is $S/K = 1.1$ (indicating the option is 10\% in-the-money) and the time to maturity is $T = 0.36$ years (approximately 90 days). The input vector is $x = [1.1, 0.36]^T$.

\noindent The first layer's trained parameters are:
\begin{equation}
\begin{aligned}
W^{[1]}&=
\begin{bmatrix}
-0.374 & 0.131 \\
-0.514 & 0.110 \\
-0.506 & 0.360 \\
\vdots & \vdots \\
0.195 & -0.793 \\
-0.587 & -1.470
\end{bmatrix}
\in \mathbb{R}^{50 \times 2},\\
b^{[1]}&=
\begin{bmatrix}
-0.0161 \\
0.3584 \\
0.3024 \\
\vdots \\
0.0571 \\
0.4862
\end{bmatrix}
\in \mathbb{R}^{50}.
\end{aligned}
\end{equation}

\noindent The first layer's activation is computed as:
\begin{equation}
a^{[1]}=\tanh(W^{[1]}x + b^{[1]})
\end{equation}

\noindent Explicitly:
\begingroup
\small
\begin{equation}
\begin{aligned}
a^{[1]}
&=\tanh\Biggl(
\underbrace{
\begin{bmatrix}
-0.374 & 0.131 \\
-0.514 & 0.110 \\
-0.506 & 0.360 \\
\vdots & \vdots \\
0.195 & -0.793 \\
-0.587 & -1.470
\end{bmatrix}}_{W^{[1]}}
\underbrace{\begin{bmatrix} 1.1 \\ 0.36 \end{bmatrix}}_{x}
+
\underbrace{
\begin{bmatrix}
-0.0161 \\
0.3584 \\
0.3024 \\
\vdots \\
0.0571 \\
0.4862
\end{bmatrix}}_{b^{[1]}}
\Biggr)\\
&=
\begin{bmatrix}
-0.3633 \\
-0.1659 \\
-0.1242 \\
\vdots \\
-0.0137 \\
-0.5970
\end{bmatrix}.
\end{aligned}
\end{equation}
\endgroup

For the second layer transformation, with parameters:
\begin{equation}
\begingroup
\scriptsize
\setlength{\arraycolsep}{3pt}
\begin{aligned}
W^{[2]}
&=
\left[
\begin{array}{cccccc}
-0.364 & -0.158 & -0.227 & \cdots & 0.139 & -0.324 \\
 0.182 &  1.045 &  0.458 & \cdots & -0.468 &  0.960 \\
-0.091 & -0.046 &  0.099 & \cdots & 0.164 &  0.275 \\
\vdots & \vdots & \vdots & \ddots & \vdots & \vdots \\
-0.435 & -0.370 & -0.320 & \cdots & 0.357 & -0.390 \\
 0.395 &  0.165 &  0.373 & \cdots & -0.017 &  0.407
\end{array}
\right]
\\
&\in \mathbb{R}^{50\times 50}.
\end{aligned}
\endgroup
\end{equation}

\begin{equation}
b^{[2]} = \begin{bmatrix}
0.248 \\
0.183 \\
0.090 \\
\vdots \\
0.085 \\
-0.990
\end{bmatrix} \in \mathbb{R}^{50}
\end{equation}
The second layer's activation is:
\begin{equation}
a^{[2]} = \tanh(W^{[2]}a^{[1]} + b^{[2]}) =
\begin{bmatrix}
0.9999 \\
-0.9920 \\
-0.1724 \\
\vdots \\
-0.9999 \\
-0.9999
\end{bmatrix}
\end{equation}

The output layer parameters are:
\begin{equation}
\begingroup
\small
\begin{aligned}
W^{[3]}
&=
\begin{bmatrix}
-0.372 & -0.221 & -0.327 & \cdots & 0.726 & 0.280
\end{bmatrix}
\\
&\in \mathbb{R}^{1\times 50}.
\end{aligned}
\endgroup
\end{equation}

\begingroup
\begin{equation}
b^{[3]}=[-0.0965].
\end{equation}
\endgroup

The final normalized option price is computed as:
\begingroup
\small
\begin{equation}
\frac{C}{K}
=
\text{softplus}\!\bigl(W^{[3]}a^{[2]}+b^{[3]}\bigr)
=
\text{softplus}(0.1039)
=
0.1039.
\end{equation}
\endgroup

\paragraph{Result Interpretation:}
The trained FINN model produces a price of \$10.39 (computed as $0.1039 \times K = 0.1039 \times 100$), which differs by only \$0.01 from the analytical Black-Scholes price of \$10.38. Similarly, FINN's estimated Delta of 0.86, obtained through automatic differentiation, approximates the Black-Scholes Delta of 0.90 with reasonable accuracy. This detailed walkthrough demonstrates how the network architecture translates inputs into accurate, theoretically consistent option prices and hedge ratios.

\section{Additional Sample Paths Analysis for Stress Tests}
\label{app:robustness_paths}
\begin{figure}
{\includegraphics[width=\linewidth]{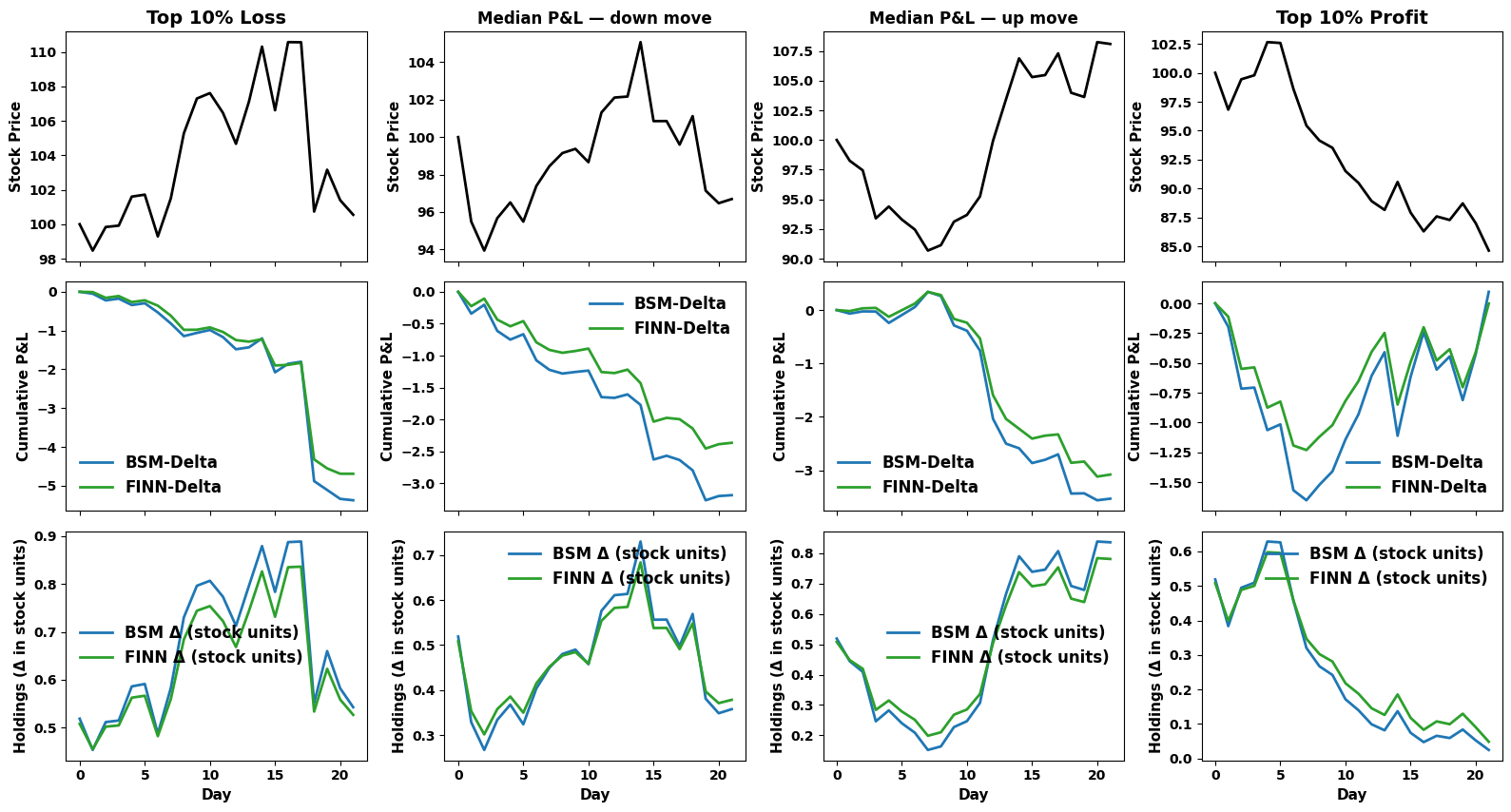}}
\caption{Representative portfolio hedging paths using volatility from the year of 2008 (crisis regime, $\delta=2\%$). Each panel shows a sample path of the stock price (top row), cumulative hedged P\&L (middle row), and delta holdings in stock units (bottom row) for BSM- and FINN-based delta hedging. The three panels on the left show a representative trajectory drawn from the worst 10\% of outcomes among 50{,}000 simulated scenarios; the two middle panels illustrate representative median-P\&L down- and up-move paths; and the rightmost panel shows a representative trajectory from the best 10\%.}
\label{fig:mc_gbm_paths_2008_tc02}
\end{figure}

\begin{figure}
{\includegraphics[width=\linewidth]{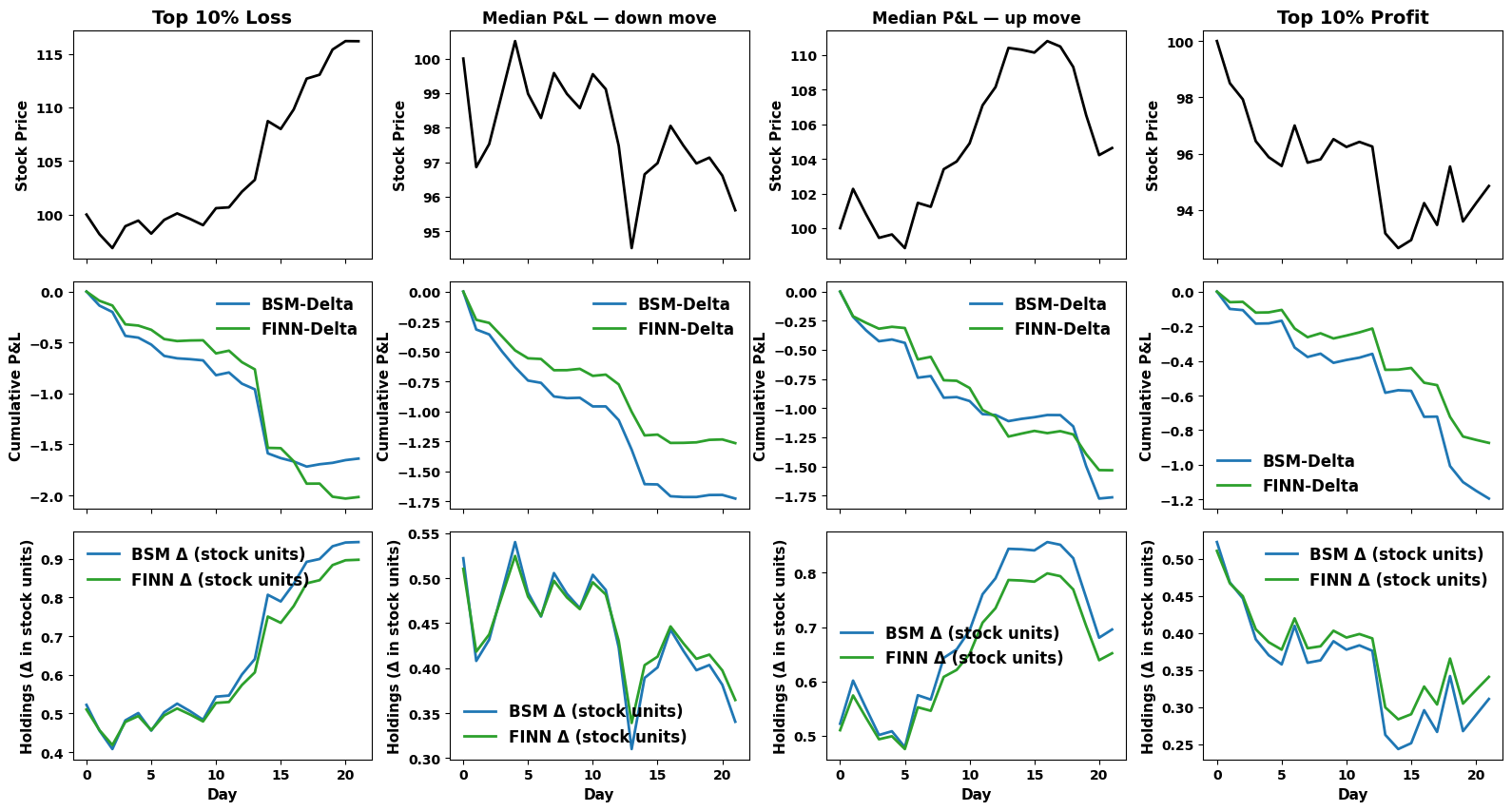}}
\caption{Representative portfolio hedging paths using volatility from the year of 2011 (moderate regime, $\delta=2\%$). Each panel shows a sample path of the stock price (top row), cumulative hedged P\&L (middle row), and delta holdings in stock units (bottom row) for BSM- and FINN-based delta hedging. The three panels on the left show a representative trajectory drawn from the worst 10\% of outcomes among 50{,}000 simulated scenarios; the two middle panels illustrate representative median-P\&L down- and up-move paths; and the rightmost panel shows a representative trajectory from the best 10\%.}
\label{fig:mc_gbm_paths_2011_tc02}
\end{figure}

\begin{figure}
{\includegraphics[width=\linewidth]{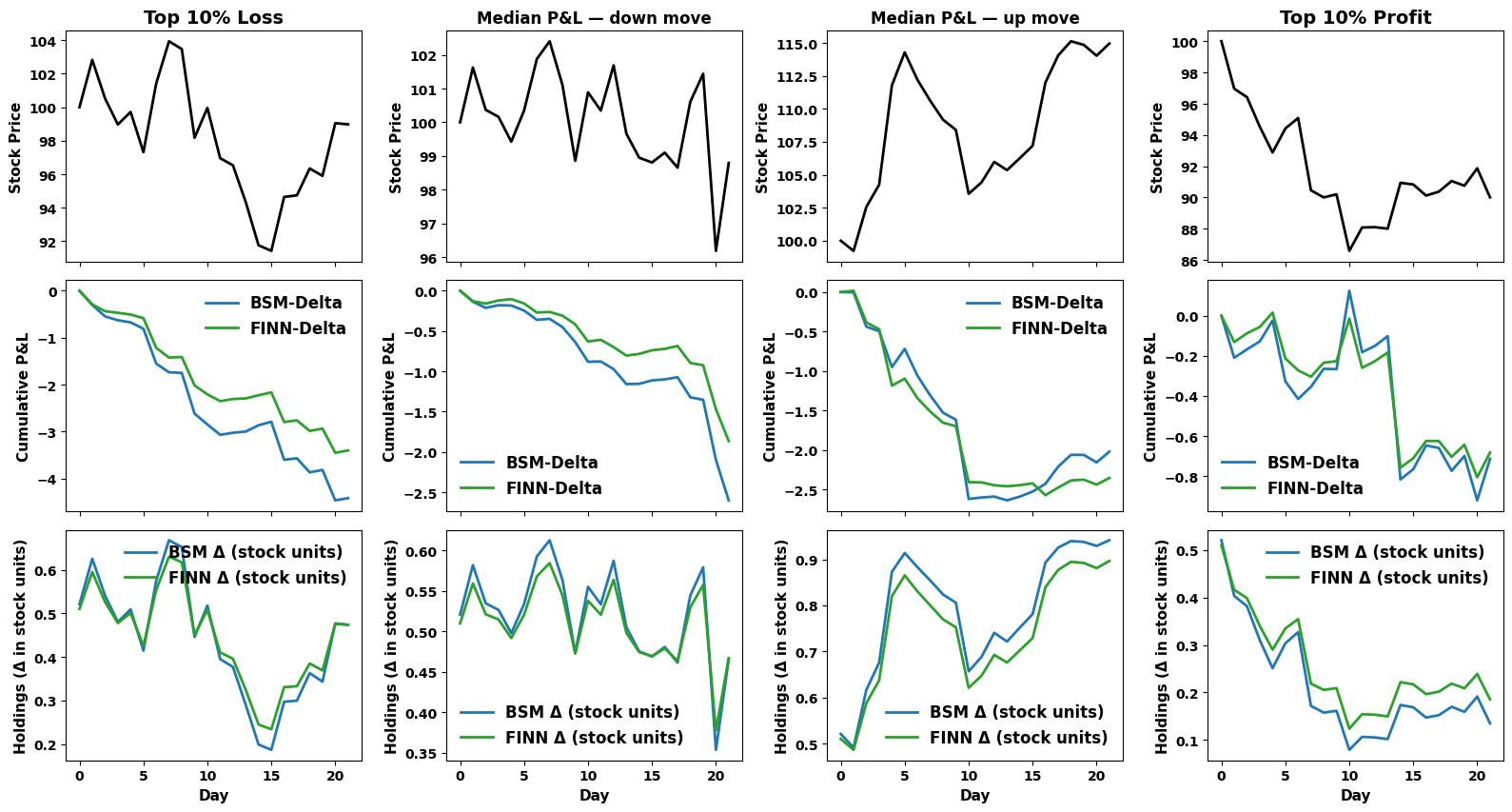}}
\caption{Representative portfolio hedging paths using volatility from the year of 2020 (crisis regime, $\delta=2\%$). Each panel shows a sample path of the stock price (top row), cumulative hedged P\&L (middle row), and delta holdings in stock units (bottom row) for BSM- and FINN-based delta hedging. The three panels on the left show a representative trajectory drawn from the worst 10\% of outcomes among 50{,}000 simulated scenarios; the two middle panels illustrate representative median-P\&L down- and up-move paths; and the rightmost panel shows a representative trajectory from the best 10\%.}
\label{fig:mc_gbm_paths_2020_tc02}
\end{figure}

The pathwise plots in Figures~\ref{fig:mc_gbm_paths_2008_tc02}--\ref{fig:mc_gbm_paths_2020_tc02} provide intuition for the aggregate results reported in the main text. In scenarios with large losses or median negative outcomes, the BSM--Delta strategy typically holds larger absolute stock positions and adjusts them more aggressively as prices decline. This leads to higher trading volume and, as a result, larger transaction costs.

\noindent In contrast, the FINN--Delta strategy generates smoother and generally smaller hedge positions, especially during crisis periods. As a consequence, the cumulative P\&L paths under FINN--Delta tend to remain consistently above those of BSM--Delta in adverse scenarios. In profitable paths, the two strategies behave similarly, indicating that both capture the primary directional risk of the option.

\noindent Overall, these pathwise comparisons support the conclusion that the FINN-based delta is more robust in the presence of transaction costs and volatility shocks. By not relying on a specific parametric model, FINN produces more conservative hedge adjustments that limit excessive trading. This allows it to perform comparably in stable market conditions while substantially reducing downside risk and trading losses during periods of increased volatility and market frictions.

\newpage
\subsection{Error Measures}
\label{sec:measures}
Let $y_i$ denote the benchmark quantity and $\hat y_i$ the corresponding FINN prediction at evaluation point $i$, and define the pointwise error
\(
e_i \;=\; y_i - \hat y_i.
\)
All metrics are computed over an index set $\mathcal{I}$ that excludes near-zero benchmark values to avoid unstable normalization. Concretely, for a user-chosen tolerance $\delta>0$, we set
\(
\mathcal{I} \;=\; \{\, i : |y_i| \ge \delta \,\},
\)
and we add a small constant $\varepsilon>0$ (e.g., $10^{-12}$) to denominators for numerical stability.

\subsubsection{Relative metrics.}
The primary metrics reported in the main text are the \emph{Relative Mean Absolute Deviation (RMAD)} and the \emph{Relative Root Mean Squared Error (RMSE)}:
\begin{equation}
\mathrm{RMAD}
\;=\;
\frac{\sum_{i\in\mathcal{I}} |e_i|}{\sum_{i\in\mathcal{I}} |y_i|+\varepsilon}.
\end{equation}
\begin{equation}
\mathrm{RMSE}
\;=\;
\sqrt{\frac{\sum_{i\in\mathcal{I}} e_i^2}{\sum_{i\in\mathcal{I}} y_i^2+\varepsilon}}.
\end{equation}

\subsubsection{Put-Call Parity Section Metrics}
Unlike the pricing and hedging benchmarks above, the put--call parity test is not evaluated against an external ``ground-truth'' price or hedge ratio. Instead, we measure the pointwise parity residual (in dollars),
\(
\varepsilon_i
\;=\;
\bigl(C_i - P_i\bigr)
-
\bigl(S_i - K_i e^{-r\tau_i}\bigr),
\)
computed over a test set of contracts. Since $\varepsilon_i$ is already an absolute arbitrage-violation measure with a direct financial interpretation, we summarize it using Mean Absolute Deviation (MAD) and Mean Squared Error (MSE):
\begin{equation}
\mathrm{MAD}
\;=\;
\frac{1}{N}\sum_{i=1}^{N}\bigl|\varepsilon_i\bigr|,
\qquad
\mathrm{MSE}
\;=\;
\frac{1}{N}\sum_{i=1}^{N}\varepsilon_i^2,
\end{equation}
where $N$ is the number of evaluated contracts. This choice keeps the parity residual comparable across GBM and Heston settings and preserves its interpretation as a dollar-valued no-arbitrage deviation.

\subsubsection{Additional Metrics.}
For implied-volatility (IV) experiments, we additionally report the \emph{Mean Absolute Percentage Error (MAPE)}, computed over the same masked set $\mathcal{I}$:
\begin{equation}
\mathrm{MAPE}
\;=\;
\frac{100}{|\mathcal{I}|}\sum_{i\in\mathcal{I}}
\frac{|e_i|}{|y_i|+\varepsilon}.
\end{equation}
Here $y_i$ and $\hat y_i$ denote the benchmark and predicted implied volatilities, respectively; the factor $100$ expresses the result in percent.

\noindent For completeness, we also report \emph{Normalized MAD (NMAD)} and \emph{MSE} in the appendix. Define the masked range
\begin{equation}
\mathrm{Range}(y;\mathcal{I})
\;=\;
\max_{i\in\mathcal{I}} y_i
-
\min_{i\in\mathcal{I}} y_i.
\end{equation}
Then
\begin{equation}
\mathrm{NMAD}
\;=\;
\frac{\frac{1}{|\mathcal{I}|}\sum_{i\in\mathcal{I}} |e_i|}{\mathrm{Range}(y;\mathcal{I})+\varepsilon}.
\end{equation}

\begin{equation}
\mathrm{MSE}
\;=\;
\frac{1}{N}\sum_{i=1}^{N} e_i^2.
\end{equation}
where $N$ is the total number of evaluation points.

\newpage
\clearpage
\section{Additional Results}
\subsection{Additional GBM European Call Results}\label{secA2_call}

\begin{table}
\centering
{%
\footnotesize
\begin{tabular}{@{}c c c c c c c c c c@{}}
\hline
\multirow{2}{*}{$\sigma$} & \multirow{2}{*}{TTM $\tau$}
& \multicolumn{4}{c}{Call option price $(C)$}
& \multicolumn{4}{c}{Hedge ratio $(\Delta)$} \\
\cline{3-10}
& & \multicolumn{2}{c}{NMAD} & \multicolumn{2}{c}{MSE}
  & \multicolumn{2}{c}{NMAD} & \multicolumn{2}{c}{MSE} \\
\hline
\multirow{7}{*}{0.125}
& 0.24 & 0.006 & (0.001) & 0.051 & (0.023) & 0.027 & (0.006) & 0.001 & (0.000) \\
& 0.28 & 0.007 & (0.001) & 0.057 & (0.026) & 0.028 & (0.006) & 0.001 & (0.000) \\
& 0.32 & 0.007 & (0.002) & 0.062 & (0.028) & 0.028 & (0.006) & 0.001 & (0.000) \\
& 0.36 & 0.007 & (0.002) & 0.066 & (0.031) & 0.027 & (0.006) & 0.001 & (0.000) \\
& 0.40 & 0.007 & (0.002) & 0.068 & (0.033) & 0.027 & (0.006) & 0.001 & (0.000) \\
& 0.44 & 0.007 & (0.002) & 0.069 & (0.034) & 0.026 & (0.006) & 0.001 & (0.000) \\
& 0.48 & 0.007 & (0.002) & 0.069 & (0.035) & 0.025 & (0.006) & 0.001 & (0.000) \\
\hline
\multirow{7}{*}{0.150}
& 0.24 & 0.008 & (0.002) & 0.085 & (0.037) & 0.031 & (0.007) & 0.001 & (0.000) \\
& 0.28 & 0.008 & (0.002) & 0.095 & (0.042) & 0.031 & (0.007) & 0.001 & (0.000) \\
& 0.32 & 0.009 & (0.002) & 0.102 & (0.046) & 0.030 & (0.007) & 0.001 & (0.000) \\
& 0.36 & 0.009 & (0.002) & 0.107 & (0.050) & 0.030 & (0.007) & 0.001 & (0.000) \\
& 0.40 & 0.009 & (0.002) & 0.110 & (0.052) & 0.029 & (0.007) & 0.001 & (0.000) \\
& 0.44 & 0.009 & (0.002) & 0.110 & (0.054) & 0.028 & (0.006) & 0.001 & (0.000) \\
& 0.48 & 0.009 & (0.002) & 0.109 & (0.055) & 0.027 & (0.006) & 0.001 & (0.000) \\
\hline
\multirow{7}{*}{0.175}
& 0.24 & 0.010 & (0.002) & 0.132 & (0.056) & 0.034 & (0.007) & 0.001 & (0.001) \\
& 0.28 & 0.010 & (0.002) & 0.146 & (0.063) & 0.034 & (0.007) & 0.001 & (0.001) \\
& 0.32 & 0.010 & (0.002) & 0.156 & (0.069) & 0.034 & (0.008) & 0.001 & (0.001) \\
& 0.36 & 0.011 & (0.003) & 0.164 & (0.074) & 0.033 & (0.008) & 0.001 & (0.001) \\
& 0.40 & 0.011 & (0.003) & 0.168 & (0.077) & 0.032 & (0.007) & 0.001 & (0.001) \\
& 0.44 & 0.011 & (0.003) & 0.169 & (0.080) & 0.031 & (0.007) & 0.001 & (0.001) \\
& 0.48 & 0.010 & (0.003) & 0.167 & (0.082) & 0.030 & (0.007) & 0.001 & (0.000) \\
\hline
\end{tabular}
}%
\caption{FINN performance against Black--Scholes (GBM) for European call options. Entries report mean (standard deviation) across seeds. NMAD is normalized mean absolute deviation.}
\label{finn_gbm_nmad}
\end{table}

\clearpage
\subsection{GBM European Put Results}\label{secA2}
Here we present the results and summary statistics of running the FINN on the put options:

\begin{table}
\centering
\label{tab:finn_gbm_put}
{%
\begin{tabular}{ccrrrr}
\toprule
\multicolumn{1}{c}{$\sigma$} & \multicolumn{1}{c}{TTM $\tau$} &
\multicolumn{2}{c}{Put price $(P)$} &
\multicolumn{2}{c}{Delta $(\Delta)$} \\
\cmidrule(lr){3-4}\cmidrule(lr){5-6}
& & RMAD & RMSE & RMAD & RMSE \\
\midrule
\multirow{7}{*}{0.125}
& 0.24 & 0.021 (0.006) & 0.021 (0.006) & 0.040 (0.013) & 0.043 (0.013) \\
& 0.28 & 0.023 (0.007) & 0.022 (0.006) & 0.041 (0.013) & 0.044 (0.013) \\
& 0.32 & 0.025 (0.008) & 0.023 (0.007) & 0.042 (0.013) & 0.044 (0.013) \\
& 0.36 & 0.025 (0.008) & 0.024 (0.007) & 0.042 (0.013) & 0.043 (0.013) \\
& 0.40 & 0.026 (0.009) & 0.024 (0.007) & 0.042 (0.013) & 0.043 (0.012) \\
& 0.44 & 0.027 (0.009) & 0.024 (0.008) & 0.042 (0.013) & 0.042 (0.012) \\
& 0.48 & 0.027 (0.009) & 0.024 (0.008) & 0.041 (0.013) & 0.041 (0.012) \\
\midrule
\multirow{7}{*}{0.150}
& 0.24 & 0.032 (0.009) & 0.029 (0.006) & 0.047 (0.008) & 0.047 (0.007) \\
& 0.28 & 0.034 (0.009) & 0.030 (0.006) & 0.049 (0.008) & 0.048 (0.007) \\
& 0.32 & 0.036 (0.009) & 0.032 (0.006) & 0.050 (0.008) & 0.049 (0.007) \\
& 0.36 & 0.037 (0.009) & 0.032 (0.007) & 0.051 (0.008) & 0.049 (0.007) \\
& 0.40 & 0.038 (0.009) & 0.033 (0.007) & 0.051 (0.009) & 0.049 (0.008) \\
& 0.44 & 0.039 (0.009) & 0.033 (0.007) & 0.051 (0.009) & 0.048 (0.008) \\
& 0.48 & 0.039 (0.010) & 0.033 (0.007) & 0.050 (0.010) & 0.047 (0.008) \\
\midrule
\multirow{7}{*}{0.175}
& 0.24 & 0.039 (0.009) & 0.034 (0.008) & 0.052 (0.012) & 0.050 (0.011) \\
& 0.28 & 0.041 (0.010) & 0.035 (0.008) & 0.054 (0.012) & 0.050 (0.011) \\
& 0.32 & 0.043 (0.010) & 0.036 (0.009) & 0.054 (0.012) & 0.050 (0.011) \\
& 0.36 & 0.044 (0.011) & 0.037 (0.009) & 0.053 (0.013) & 0.049 (0.011) \\
& 0.40 & 0.044 (0.011) & 0.037 (0.009) & 0.053 (0.013) & 0.048 (0.011) \\
& 0.44 & 0.044 (0.012) & 0.037 (0.010) & 0.051 (0.012) & 0.047 (0.011) \\
& 0.48 & 0.044 (0.012) & 0.036 (0.010) & 0.050 (0.012) & 0.046 (0.011) \\
\bottomrule
\end{tabular}
}
\caption{FINN testing results under GBM for European put options across volatility levels and maturities. Results are averaged across strikes $K\in[75,125]$ over 10000 test points and 10 independent runs. Parentheses report standard deviations across runs.}
\end{table}

\begin{table}
\centering
{%
\begin{tabular}{ccrrrr}
\toprule
\multicolumn{1}{c}{$\sigma$} & \multicolumn{1}{c}{TTM $\tau$} &
\multicolumn{2}{c}{Put price $(P)$} &
\multicolumn{2}{c}{Delta $(\Delta)$} \\
\cmidrule(lr){3-4}\cmidrule(lr){5-6}
& & NMAD & MSE & NMAD & MSE \\
\midrule
\multirow{7}{*}{0.125}
& 0.24 & 0.006 (0.002) & 0.059 (0.033) & 0.025 (0.008) & 0.001 (0.001) \\
& 0.28 & 0.006 (0.002) & 0.066 (0.038) & 0.025 (0.008) & 0.001 (0.001) \\
& 0.32 & 0.006 (0.002) & 0.072 (0.041) & 0.025 (0.008) & 0.001 (0.000) \\
& 0.36 & 0.007 (0.002) & 0.076 (0.044) & 0.025 (0.008) & 0.001 (0.000) \\
& 0.40 & 0.007 (0.002) & 0.079 (0.047) & 0.024 (0.008) & 0.001 (0.000) \\
& 0.44 & 0.007 (0.002) & 0.080 (0.049) & 0.024 (0.007) & 0.001 (0.000) \\
& 0.48 & 0.007 (0.002) & 0.081 (0.050) & 0.023 (0.007) & 0.001 (0.000) \\
\midrule
\multirow{7}{*}{0.150}
& 0.24 & 0.008 (0.002) & 0.108 (0.047) & 0.028 (0.005) & 0.001 (0.000) \\
& 0.28 & 0.009 (0.002) & 0.120 (0.050) & 0.028 (0.004) & 0.001 (0.000) \\
& 0.32 & 0.009 (0.002) & 0.131 (0.053) & 0.028 (0.004) & 0.001 (0.000) \\
& 0.36 & 0.009 (0.002) & 0.139 (0.057) & 0.028 (0.005) & 0.001 (0.000) \\
& 0.40 & 0.009 (0.002) & 0.146 (0.060) & 0.028 (0.005) & 0.001 (0.000) \\
& 0.44 & 0.009 (0.002) & 0.150 (0.063) & 0.027 (0.005) & 0.001 (0.000) \\
& 0.48 & 0.009 (0.002) & 0.151 (0.066) & 0.026 (0.005) & 0.001 (0.000) \\
\midrule
\multirow{7}{*}{0.175}
& 0.24 & 0.010 (0.002) & 0.150 (0.068) & 0.029 (0.007) & 0.001 (0.000) \\
& 0.28 & 0.010 (0.002) & 0.166 (0.077) & 0.029 (0.007) & 0.001 (0.000) \\
& 0.32 & 0.010 (0.002) & 0.177 (0.084) & 0.029 (0.007) & 0.001 (0.000) \\
& 0.36 & 0.011 (0.003) & 0.184 (0.089) & 0.028 (0.007) & 0.001 (0.000) \\
& 0.40 & 0.011 (0.003) & 0.188 (0.093) & 0.027 (0.007) & 0.001 (0.000) \\
& 0.44 & 0.011 (0.003) & 0.190 (0.096) & 0.026 (0.006) & 0.001 (0.000) \\
& 0.48 & 0.011 (0.003) & 0.189 (0.099) & 0.026 (0.006) & 0.001 (0.000) \\
\bottomrule
\end{tabular}
}
\caption{FINN testing results under GBM for European put options across volatility levels and maturities. Results are averaged across strikes $K\in[75,125]$ over 10{,}000 test points and 10 independent runs. NMAD normalizes the mean absolute deviation by the empirical range of benchmark values.}
\label{tab:finn_gbm_put_nmad_mse}
\end{table}

\newpage
\clearpage
\subsection{Additional Heston European Call Results}\label{secA3_call}

\begin{table}
\centering
{%
\begin{tabular}{ccrrrr}
\toprule
\multicolumn{1}{c}{$\xi$} & \multicolumn{1}{c}{TTM $\tau$} &
\multicolumn{2}{c}{Call price $(C)$} &
\multicolumn{2}{c}{Delta $(\Delta)$} \\
\cmidrule(lr){3-4}\cmidrule(lr){5-6}
& & NMAD & MSE & NMAD & MSE \\
\midrule
\multirow{7}{*}{0.125}
& 0.24 & 0.008 (0.004) & 0.104 (0.102) & 0.032 (0.016) & 0.001 (0.001) \\
& 0.28 & 0.009 (0.004) & 0.122 (0.115) & 0.033 (0.016) & 0.001 (0.001) \\
& 0.32 & 0.010 (0.004) & 0.139 (0.128) & 0.034 (0.016) & 0.001 (0.001) \\
& 0.36 & 0.010 (0.005) & 0.155 (0.139) & 0.034 (0.016) & 0.002 (0.001) \\
& 0.40 & 0.011 (0.005) & 0.169 (0.148) & 0.035 (0.016) & 0.002 (0.001) \\
& 0.44 & 0.011 (0.005) & 0.183 (0.155) & 0.035 (0.015) & 0.002 (0.001) \\
& 0.48 & 0.011 (0.004) & 0.195 (0.161) & 0.035 (0.015) & 0.002 (0.001) \\
\midrule
\multirow{7}{*}{0.150}
& 0.24 & 0.009 (0.004) & 0.115 (0.109) & 0.034 (0.015) & 0.001 (0.001) \\
& 0.28 & 0.010 (0.004) & 0.136 (0.127) & 0.035 (0.015) & 0.002 (0.001) \\
& 0.32 & 0.010 (0.005) & 0.156 (0.143) & 0.036 (0.015) & 0.002 (0.001) \\
& 0.36 & 0.011 (0.005) & 0.174 (0.157) & 0.037 (0.015) & 0.002 (0.001) \\
& 0.40 & 0.011 (0.005) & 0.191 (0.168) & 0.038 (0.015) & 0.002 (0.001) \\
& 0.44 & 0.012 (0.005) & 0.207 (0.177) & 0.038 (0.015) & 0.002 (0.001) \\
& 0.48 & 0.012 (0.005) & 0.221 (0.184) & 0.038 (0.014) & 0.002 (0.001) \\
\midrule
\multirow{7}{*}{0.175}
& 0.24 & 0.010 (0.004) & 0.138 (0.128) & 0.037 (0.014) & 0.002 (0.001) \\
& 0.28 & 0.011 (0.004) & 0.164 (0.147) & 0.038 (0.014) & 0.002 (0.001) \\
& 0.32 & 0.012 (0.004) & 0.191 (0.164) & 0.040 (0.014) & 0.002 (0.001) \\
& 0.36 & 0.012 (0.004) & 0.217 (0.180) & 0.041 (0.014) & 0.002 (0.001) \\
& 0.40 & 0.013 (0.005) & 0.242 (0.194) & 0.042 (0.014) & 0.002 (0.001) \\
& 0.44 & 0.014 (0.005) & 0.266 (0.205) & 0.043 (0.013) & 0.002 (0.001) \\
& 0.48 & 0.014 (0.005) & 0.288 (0.215) & 0.043 (0.013) & 0.002 (0.001) \\
\bottomrule
\end{tabular}
}
\caption{FINN performance against the Heston stochastic volatility model for European call options. Entries report NMAD and MSE with standard deviations in parentheses (across seeds). NMAD normalizes mean absolute deviation by the empirical range of benchmark values on the test grid.}
\label{table_heston_nmad_mse}
\end{table}

\begin{table}
\centering
{%
\begin{tabular}{ccrrrr}
\toprule
\multicolumn{1}{c}{$\xi$} & \multicolumn{1}{c}{TTM $\tau$} &
\multicolumn{2}{c}{Put price $(P)$} &
\multicolumn{2}{c}{Delta $(\Delta)$} \\
\cmidrule(lr){3-4}\cmidrule(lr){5-6}
& & RMAD & RMSE & RMAD & RMSE \\
\midrule
\multirow{7}{*}{0.125}
& 0.24 & 0.030 (0.018) & 0.027 (0.016) & 0.048 (0.023) & 0.048 (0.021) \\
& 0.28 & 0.033 (0.019) & 0.029 (0.016) & 0.051 (0.023) & 0.050 (0.022) \\
& 0.32 & 0.036 (0.020) & 0.031 (0.017) & 0.054 (0.024) & 0.052 (0.022) \\
& 0.36 & 0.038 (0.021) & 0.033 (0.017) & 0.056 (0.023) & 0.053 (0.021) \\
& 0.40 & 0.039 (0.021) & 0.034 (0.018) & 0.057 (0.023) & 0.054 (0.021) \\
& 0.44 & 0.041 (0.021) & 0.035 (0.018) & 0.058 (0.023) & 0.055 (0.021) \\
& 0.48 & 0.043 (0.021) & 0.037 (0.017) & 0.060 (0.023) & 0.056 (0.020) \\
\midrule
\multirow{7}{*}{0.150}
& 0.24 & 0.030 (0.018) & 0.027 (0.016) & 0.053 (0.022) & 0.052 (0.020) \\
& 0.28 & 0.033 (0.019) & 0.029 (0.017) & 0.056 (0.023) & 0.054 (0.021) \\
& 0.32 & 0.035 (0.020) & 0.031 (0.017) & 0.058 (0.024) & 0.056 (0.021) \\
& 0.36 & 0.037 (0.021) & 0.033 (0.018) & 0.060 (0.024) & 0.057 (0.021) \\
& 0.40 & 0.039 (0.021) & 0.034 (0.018) & 0.062 (0.024) & 0.058 (0.021) \\
& 0.44 & 0.041 (0.022) & 0.036 (0.019) & 0.064 (0.025) & 0.059 (0.021) \\
& 0.48 & 0.043 (0.022) & 0.037 (0.019) & 0.065 (0.025) & 0.060 (0.021) \\
\midrule
\multirow{7}{*}{0.175}
& 0.24 & 0.033 (0.018) & 0.030 (0.016) & 0.058 (0.021) & 0.058 (0.019) \\
& 0.28 & 0.036 (0.019) & 0.032 (0.017) & 0.062 (0.022) & 0.061 (0.020) \\
& 0.32 & 0.039 (0.020) & 0.035 (0.017) & 0.065 (0.023) & 0.063 (0.020) \\
& 0.36 & 0.041 (0.021) & 0.036 (0.018) & 0.068 (0.023) & 0.065 (0.020) \\
& 0.40 & 0.044 (0.021) & 0.038 (0.018) & 0.070 (0.023) & 0.066 (0.020) \\
& 0.44 & 0.046 (0.021) & 0.040 (0.018) & 0.072 (0.023) & 0.068 (0.020) \\
& 0.48 & 0.048 (0.021) & 0.042 (0.018) & 0.074 (0.023) & 0.069 (0.020) \\
\bottomrule
\end{tabular}
}
\caption{FINN testing results under Heston for European put options across vol-of-vol levels and maturities. Results are averaged across strikes $K\in[75,125]$ over 10{,}000 test points and 10 independent runs; parentheses report standard deviations. RMAD and RMSE are computed relative to the Heston benchmark on the test grid.}
\label{tab:heston_put_rmad_rmse}
\end{table}

\begin{table}
\centering
{%
\begin{tabular}{ccrrrr}
\toprule
\multicolumn{1}{c}{$\xi$} & \multicolumn{1}{c}{TTM $\tau$} &
\multicolumn{2}{c}{Put price $(P)$} &
\multicolumn{2}{c}{Delta $(\Delta)$} \\
\cmidrule(lr){3-4}\cmidrule(lr){5-6}
& & NMAD & MSE & NMAD & MSE \\
\midrule
\multirow{7}{*}{0.125}
& 0.24 & 0.007 (0.004) & 0.117 (0.109) & 0.027 (0.013) & 0.001 (0.001) \\
& 0.28 & 0.008 (0.005) & 0.137 (0.127) & 0.028 (0.013) & 0.001 (0.001) \\
& 0.32 & 0.008 (0.005) & 0.155 (0.142) & 0.029 (0.013) & 0.001 (0.001) \\
& 0.36 & 0.009 (0.005) & 0.170 (0.153) & 0.029 (0.012) & 0.001 (0.001) \\
& 0.40 & 0.009 (0.005) & 0.184 (0.163) & 0.030 (0.012) & 0.001 (0.001) \\
& 0.44 & 0.009 (0.005) & 0.197 (0.171) & 0.030 (0.012) & 0.001 (0.001) \\
& 0.48 & 0.010 (0.005) & 0.210 (0.179) & 0.030 (0.012) & 0.001 (0.001) \\
\midrule
\multirow{7}{*}{0.150}
& 0.24 & 0.007 (0.004) & 0.119 (0.130) & 0.030 (0.012) & 0.001 (0.001) \\
& 0.28 & 0.008 (0.005) & 0.139 (0.149) & 0.031 (0.013) & 0.001 (0.001) \\
& 0.32 & 0.008 (0.005) & 0.156 (0.166) & 0.031 (0.013) & 0.001 (0.001) \\
& 0.36 & 0.009 (0.005) & 0.173 (0.181) & 0.032 (0.013) & 0.001 (0.001) \\
& 0.40 & 0.009 (0.005) & 0.189 (0.194) & 0.032 (0.013) & 0.002 (0.001) \\
& 0.44 & 0.009 (0.005) & 0.204 (0.207) & 0.033 (0.013) & 0.002 (0.001) \\
& 0.48 & 0.010 (0.005) & 0.219 (0.217) & 0.033 (0.013) & 0.002 (0.001) \\
\midrule
\multirow{7}{*}{0.175}
& 0.24 & 0.008 (0.004) & 0.139 (0.141) & 0.032 (0.012) & 0.002 (0.001) \\
& 0.28 & 0.009 (0.005) & 0.163 (0.163) & 0.034 (0.012) & 0.002 (0.001) \\
& 0.32 & 0.009 (0.005) & 0.184 (0.181) & 0.035 (0.012) & 0.002 (0.001) \\
& 0.36 & 0.009 (0.005) & 0.204 (0.198) & 0.036 (0.012) & 0.002 (0.001) \\
& 0.40 & 0.010 (0.005) & 0.224 (0.212) & 0.036 (0.012) & 0.002 (0.001) \\
& 0.44 & 0.010 (0.005) & 0.243 (0.225) & 0.037 (0.012) & 0.002 (0.001) \\
& 0.48 & 0.011 (0.005) & 0.263 (0.236) & 0.038 (0.012) & 0.002 (0.001) \\

\bottomrule
\end{tabular}
}
\caption{FINN testing results under Heston for European put options across vol-of-vol levels and maturities. Results are averaged across strikes $K\in[75,125]$ over 10{,}000 test points and 10 independent runs; parentheses report standard deviations. NMAD normalizes mean absolute deviation by the empirical range of benchmark values on the test grid.}
\label{tab:heston_put_nmad_mse}
\end{table}

\newpage
\clearpage
\subsection{Additional American Put Results}

\begin{table}
\centering
{%
\begin{tabular}{@{}c c c c c c@{}}
\hline
\multirow{2}{*}{$\sigma$} & \multirow{2}{*}{TTM $\tau$}
& \multicolumn{2}{c}{Put option price $(P)$}
& \multicolumn{2}{c}{Hedge ratio $(\Delta)$} \\
\cline{3-6}
& & NMAD & MSE & NMAD & MSE \\
\hline
\multirow{7}{*}{0.125}
& 0.240 & 0.013 (0.006) & 0.288 (0.232) & 0.046 (0.016) & 0.004 (0.003) \\
& 0.280 & 0.013 (0.007) & 0.326 (0.269) & 0.046 (0.017) & 0.004 (0.003) \\
& 0.320 & 0.013 (0.007) & 0.353 (0.303) & 0.047 (0.019) & 0.004 (0.004) \\
& 0.360 & 0.014 (0.008) & 0.390 (0.346) & 0.048 (0.020) & 0.004 (0.004) \\
& 0.400 & 0.014 (0.008) & 0.420 (0.384) & 0.049 (0.021) & 0.004 (0.004) \\
& 0.440 & 0.015 (0.008) & 0.450 (0.420) & 0.050 (0.022) & 0.005 (0.005) \\
& 0.480 & 0.015 (0.008) & 0.482 (0.458) & 0.052 (0.023) & 0.005 (0.005) \\
\hline
\multirow{7}{*}{0.150}
& 0.240 & 0.009 (0.003) & 0.124 (0.079) & 0.041 (0.009) & 0.003 (0.002) \\
& 0.280 & 0.009 (0.003) & 0.129 (0.091) & 0.040 (0.009) & 0.003 (0.002) \\
& 0.320 & 0.009 (0.003) & 0.131 (0.100) & 0.040 (0.008) & 0.003 (0.001) \\
& 0.360 & 0.009 (0.004) & 0.137 (0.113) & 0.040 (0.008) & 0.003 (0.001) \\
& 0.400 & 0.009 (0.004) & 0.139 (0.122) & 0.041 (0.008) & 0.003 (0.001) \\
& 0.440 & 0.009 (0.004) & 0.144 (0.131) & 0.042 (0.008) & 0.003 (0.001) \\
& 0.480 & 0.009 (0.004) & 0.151 (0.140) & 0.043 (0.008) & 0.003 (0.001) \\
\hline
\multirow{7}{*}{0.175}
& 0.240 & 0.007 (0.003) & 0.096 (0.075) & 0.035 (0.013) & 0.002 (0.002) \\
& 0.280 & 0.007 (0.003) & 0.094 (0.080) & 0.032 (0.013) & 0.002 (0.002) \\
& 0.320 & 0.007 (0.003) & 0.095 (0.081) & 0.031 (0.013) & 0.002 (0.002) \\
& 0.360 & 0.007 (0.003) & 0.095 (0.086) & 0.030 (0.013) & 0.002 (0.002) \\
& 0.400 & 0.007 (0.002) & 0.104 (0.084) & 0.031 (0.013) & 0.002 (0.002) \\
& 0.440 & 0.010 (0.006) & 0.213 (0.275) & 0.032 (0.012) & 0.002 (0.002) \\
& 0.480 & 0.010 (0.006) & 0.225 (0.257) & 0.033 (0.012) & 0.002 (0.001) \\
\hline
\end{tabular}
}%
\caption{American put option FINN prediction NMAD/MSE results with finite-difference benchmark for different $\sigma$ values. Entries report mean (standard deviation) across seeds.}
\label{tab:finn_results_put_nmad_mse}
\end{table}

\newpage
\subsection{Full Delta-Gamma Hedging Results}
\label{app:full_gamma_results}
The following tables provide the complete, granular results for the delta-gamma hedging experiments across all tested time-to-maturity parameters.

\begin{table}
\centering
{%
\small\setlength{\tabcolsep}{6pt}
\begin{tabular}{cclrr}
\toprule
\multirow{2}{*}{Hedge TTM $\tau_h$} & \multirow{2}{*}{Option TTM $\tau$} & \multirow{2}{*}{Metric} & \multicolumn{2}{c}{Error} \\
\cmidrule(lr){4-5}
& & & RMAD & RMSE \\
\midrule
\multirow{21}{*}{0.12} & \multirow{3}{*}{0.24} & Price $(C)$ & 0.003 (0.001) & 0.003 (0.001) \\
 &  & Delta $(\Delta)$ & 0.008 (0.002) & 0.009 (0.002) \\
 &  & Gamma $(\Gamma)$ & 0.059 (0.017) & 0.061 (0.020) \\
\cmidrule(lr){2-5}
 & \multirow{3}{*}{0.28} & Price $(C)$ & 0.003 (0.001) & 0.003 (0.001) \\
 &  & Delta $(\Delta)$ & 0.008 (0.002) & 0.008 (0.002) \\
 &  & Gamma $(\Gamma)$ & 0.058 (0.017) & 0.060 (0.020) \\
\cmidrule(lr){2-5}
 & \multirow{3}{*}{0.32} & Price $(C)$ & 0.003 (0.001) & 0.003 (0.001) \\
 &  & Delta $(\Delta)$ & 0.008 (0.002) & 0.008 (0.002) \\
 &  & Gamma $(\Gamma)$ & 0.057 (0.017) & 0.060 (0.020) \\
\cmidrule(lr){2-5}
 & \multirow{3}{*}{0.36} & Price $(C)$ & 0.003 (0.001) & 0.003 (0.001) \\
 &  & Delta $(\Delta)$ & 0.008 (0.002) & 0.008 (0.002) \\
 &  & Gamma $(\Gamma)$ & 0.056 (0.018) & 0.059 (0.021) \\
\cmidrule(lr){2-5}
 & \multirow{3}{*}{0.40} & Price $(C)$ & 0.003 (0.001) & 0.003 (0.001) \\
 &  & Delta $(\Delta)$ & 0.007 (0.002) & 0.008 (0.002) \\
 &  & Gamma $(\Gamma)$ & 0.055 (0.018) & 0.059 (0.021) \\
\cmidrule(lr){2-5}
 & \multirow{3}{*}{0.44} & Price $(C)$ & 0.003 (0.001) & 0.003 (0.002) \\
 &  & Delta $(\Delta)$ & 0.007 (0.002) & 0.008 (0.002) \\
 &  & Gamma $(\Gamma)$ & 0.054 (0.018) & 0.058 (0.021) \\
\cmidrule(lr){2-5}
 & \multirow{3}{*}{0.48} & Price $(C)$ & 0.003 (0.001) & 0.003 (0.002) \\
 &  & Delta $(\Delta)$ & 0.007 (0.002) & 0.008 (0.002) \\
 &  & Gamma $(\Gamma)$ & 0.053 (0.017) & 0.058 (0.020) \\
\midrule
\multirow{21}{*}{0.36} & \multirow{3}{*}{0.24} & Price $(C)$ & 0.003 (0.002) & 0.003 (0.002) \\
 &  & Delta $(\Delta)$ & 0.005 (0.002) & 0.006 (0.002) \\
 &  & Gamma $(\Gamma)$ & 0.046 (0.012) & 0.049 (0.014) \\
\cmidrule(lr){2-5}
 & \multirow{3}{*}{0.28} & Price $(C)$ & 0.003 (0.002) & 0.003 (0.002) \\
 &  & Delta $(\Delta)$ & 0.005 (0.002) & 0.006 (0.002) \\
 &  & Gamma $(\Gamma)$ & 0.044 (0.012) & 0.048 (0.014) \\
\cmidrule(lr){2-5}
 & \multirow{3}{*}{0.32} & Price $(C)$ & 0.003 (0.002) & 0.003 (0.002) \\
 &  & Delta $(\Delta)$ & 0.005 (0.002) & 0.006 (0.002) \\
 &  & Gamma $(\Gamma)$ & 0.042 (0.012) & 0.046 (0.014) \\
\cmidrule(lr){2-5}
 & \multirow{3}{*}{0.36} & Price $(C)$ & 0.003 (0.002) & 0.003 (0.002) \\
 &  & Delta $(\Delta)$ & 0.005 (0.002) & 0.006 (0.002) \\
 &  & Gamma $(\Gamma)$ & 0.040 (0.012) & 0.045 (0.014) \\
\cmidrule(lr){2-5}
 & \multirow{3}{*}{0.40} & Price $(C)$ & 0.003 (0.002) & 0.003 (0.002) \\
 &  & Delta $(\Delta)$ & 0.005 (0.002) & 0.006 (0.002) \\
 &  & Gamma $(\Gamma)$ & 0.038 (0.012) & 0.044 (0.014) \\
\cmidrule(lr){2-5}
 & \multirow{3}{*}{0.44} & Price $(C)$ & 0.003 (0.002) & 0.003 (0.002) \\
 &  & Delta $(\Delta)$ & 0.005 (0.002) & 0.006 (0.002) \\
 &  & Gamma $(\Gamma)$ & 0.036 (0.011) & 0.043 (0.014) \\
\cmidrule(lr){2-5}
 & \multirow{3}{*}{0.48} & Price $(C)$ & 0.003 (0.002) & 0.003 (0.002) \\
 &  & Delta $(\Delta)$ & 0.005 (0.002) & 0.006 (0.002) \\
 &  & Gamma $(\Gamma)$ & 0.034 (0.011) & 0.042 (0.014) \\
\bottomrule
\end{tabular}
}
\caption{FINN delta--gamma hedging performance at $\sigma=0.125$ (RMAD/RMSE). Standard deviations are in parentheses.}
\label{tab:gamma_rmad_rmse_vol0125}
\end{table}

\begin{table}
\centering
{%
\small\setlength{\tabcolsep}{6pt}
\begin{tabular}{cclrr}
\toprule
\multirow{2}{*}{Hedge TTM $\tau_h$} & \multirow{2}{*}{Option TTM $\tau$} & \multirow{2}{*}{Metric} & \multicolumn{2}{c}{Error} \\
\cmidrule(lr){4-5}
& & & RMAD & RMSE \\
\midrule
\multirow{21}{*}{0.12} & \multirow{3}{*}{0.24} & Price $(C)$ & 0.003 (0.001) & 0.002 (0.001) \\
 &  & Delta $(\Delta)$ & 0.005 (0.002) & 0.006 (0.002) \\
 &  & Gamma $(\Gamma)$ & 0.039 (0.015) & 0.039 (0.010) \\
\cmidrule(lr){2-5}
 & \multirow{3}{*}{0.28} & Price $(C)$ & 0.003 (0.001) & 0.002 (0.001) \\
 &  & Delta $(\Delta)$ & 0.005 (0.002) & 0.006 (0.002) \\
 &  & Gamma $(\Gamma)$ & 0.038 (0.015) & 0.038 (0.010) \\
\cmidrule(lr){2-5}
 & \multirow{3}{*}{0.32} & Price $(C)$ & 0.003 (0.001) & 0.002 (0.001) \\
 &  & Delta $(\Delta)$ & 0.005 (0.002) & 0.006 (0.002) \\
 &  & Gamma $(\Gamma)$ & 0.037 (0.015) & 0.038 (0.010) \\
\cmidrule(lr){2-5}
 & \multirow{3}{*}{0.36} & Price $(C)$ & 0.003 (0.001) & 0.002 (0.001) \\
 &  & Delta $(\Delta)$ & 0.005 (0.002) & 0.006 (0.002) \\
 &  & Gamma $(\Gamma)$ & 0.036 (0.015) & 0.037 (0.010) \\
\cmidrule(lr){2-5}
 & \multirow{3}{*}{0.40} & Price $(C)$ & 0.003 (0.001) & 0.002 (0.001) \\
 &  & Delta $(\Delta)$ & 0.005 (0.002) & 0.006 (0.002) \\
 &  & Gamma $(\Gamma)$ & 0.035 (0.014) & 0.037 (0.010) \\
\cmidrule(lr){2-5}
 & \multirow{3}{*}{0.44} & Price $(C)$ & 0.003 (0.001) & 0.002 (0.001) \\
 &  & Delta $(\Delta)$ & 0.005 (0.002) & 0.006 (0.002) \\
 &  & Gamma $(\Gamma)$ & 0.033 (0.014) & 0.036 (0.010) \\
\cmidrule(lr){2-5}
 & \multirow{3}{*}{0.48} & Price $(C)$ & 0.003 (0.001) & 0.002 (0.001) \\
 &  & Delta $(\Delta)$ & 0.005 (0.002) & 0.006 (0.002) \\
 &  & Gamma $(\Gamma)$ & 0.032 (0.014) & 0.035 (0.010) \\
\midrule
\multirow{21}{*}{0.36} & \multirow{3}{*}{0.24} & Price $(C)$ & 0.004 (0.002) & 0.004 (0.003) \\
 &  & Delta $(\Delta)$ & 0.005 (0.002) & 0.006 (0.002) \\
 &  & Gamma $(\Gamma)$ & 0.039 (0.012) & 0.043 (0.014) \\
\cmidrule(lr){2-5}
 & \multirow{3}{*}{0.28} & Price $(C)$ & 0.004 (0.002) & 0.004 (0.003) \\
 &  & Delta $(\Delta)$ & 0.005 (0.002) & 0.006 (0.002) \\
 &  & Gamma $(\Gamma)$ & 0.036 (0.011) & 0.041 (0.013) \\
\cmidrule(lr){2-5}
 & \multirow{3}{*}{0.32} & Price $(C)$ & 0.004 (0.002) & 0.004 (0.003) \\
 &  & Delta $(\Delta)$ & 0.005 (0.002) & 0.006 (0.002) \\
 &  & Gamma $(\Gamma)$ & 0.034 (0.011) & 0.040 (0.013) \\
\cmidrule(lr){2-5}
 & \multirow{3}{*}{0.36} & Price $(C)$ & 0.004 (0.002) & 0.004 (0.003) \\
 &  & Delta $(\Delta)$ & 0.005 (0.002) & 0.006 (0.002) \\
 &  & Gamma $(\Gamma)$ & 0.032 (0.010) & 0.038 (0.013) \\
\cmidrule(lr){2-5}
 & \multirow{3}{*}{0.40} & Price $(C)$ & 0.004 (0.002) & 0.004 (0.003) \\
 &  & Delta $(\Delta)$ & 0.005 (0.002) & 0.006 (0.002) \\
 &  & Gamma $(\Gamma)$ & 0.030 (0.010) & 0.037 (0.013) \\
\cmidrule(lr){2-5}
 & \multirow{3}{*}{0.44} & Price $(C)$ & 0.004 (0.002) & 0.004 (0.003) \\
 &  & Delta $(\Delta)$ & 0.005 (0.002) & 0.006 (0.002) \\
 &  & Gamma $(\Gamma)$ & 0.029 (0.010) & 0.036 (0.012) \\
\cmidrule(lr){2-5}
 & \multirow{3}{*}{0.48} & Price $(C)$ & 0.004 (0.002) & 0.004 (0.003) \\
 &  & Delta $(\Delta)$ & 0.005 (0.002) & 0.006 (0.002) \\
 &  & Gamma $(\Gamma)$ & 0.027 (0.010) & 0.035 (0.012) \\
\bottomrule
\end{tabular}
}
\caption{FINN delta--gamma hedging performance at $\sigma=0.150$ (RMAD/RMSE). Standard deviations are in parentheses.}
\label{tab:gamma_rmad_rmse_vol0150}
\end{table}

\begin{table}
\centering
{%
\small\setlength{\tabcolsep}{6pt}
\begin{tabular}{cclrr}
\toprule
\multirow{2}{*}{Hedge TTM $\tau_h$} & \multirow{2}{*}{Option TTM $\tau$} & \multirow{2}{*}{Metric} & \multicolumn{2}{c}{Error} \\
\cmidrule(lr){4-5}
& & & RMAD & RMSE \\
\midrule
\multirow{21}{*}{0.12} & \multirow{3}{*}{0.24} & Price $(C)$ & 0.003 (0.001) & 0.002 (0.001) \\
 &  & Delta $(\Delta)$ & 0.005 (0.002) & 0.005 (0.002) \\
 &  & Gamma $(\Gamma)$ & 0.033 (0.008) & 0.036 (0.010) \\
\cmidrule(lr){2-5}
 & \multirow{3}{*}{0.28} & Price $(C)$ & 0.003 (0.001) & 0.002 (0.001) \\
 &  & Delta $(\Delta)$ & 0.004 (0.002) & 0.005 (0.002) \\
 &  & Gamma $(\Gamma)$ & 0.032 (0.008) & 0.036 (0.010) \\
\cmidrule(lr){2-5}
 & \multirow{3}{*}{0.32} & Price $(C)$ & 0.003 (0.001) & 0.002 (0.001) \\
 &  & Delta $(\Delta)$ & 0.004 (0.002) & 0.005 (0.002) \\
 &  & Gamma $(\Gamma)$ & 0.031 (0.008) & 0.036 (0.010) \\
\cmidrule(lr){2-5}
 & \multirow{3}{*}{0.36} & Price $(C)$ & 0.003 (0.001) & 0.002 (0.001) \\
 &  & Delta $(\Delta)$ & 0.004 (0.002) & 0.005 (0.002) \\
 &  & Gamma $(\Gamma)$ & 0.029 (0.007) & 0.035 (0.010) \\
\cmidrule(lr){2-5}
 & \multirow{3}{*}{0.40} & Price $(C)$ & 0.003 (0.001) & 0.002 (0.001) \\
 &  & Delta $(\Delta)$ & 0.004 (0.002) & 0.005 (0.002) \\
 &  & Gamma $(\Gamma)$ & 0.028 (0.007) & 0.034 (0.010) \\
\cmidrule(lr){2-5}
 & \multirow{3}{*}{0.44} & Price $(C)$ & 0.003 (0.001) & 0.002 (0.001) \\
 &  & Delta $(\Delta)$ & 0.004 (0.002) & 0.005 (0.002) \\
 &  & Gamma $(\Gamma)$ & 0.027 (0.007) & 0.033 (0.010) \\
\cmidrule(lr){2-5}
 & \multirow{3}{*}{0.48} & Price $(C)$ & 0.003 (0.001) & 0.002 (0.001) \\
 &  & Delta $(\Delta)$ & 0.004 (0.002) & 0.005 (0.002) \\
 &  & Gamma $(\Gamma)$ & 0.025 (0.007) & 0.032 (0.009) \\
\midrule
\multirow{21}{*}{0.36} & \multirow{3}{*}{0.24} & Price $(C)$ & 0.002 (0.002) & 0.003 (0.002) \\
 &  & Delta $(\Delta)$ & 0.005 (0.002) & 0.006 (0.002) \\
 &  & Gamma $(\Gamma)$ & 0.031 (0.010) & 0.036 (0.012) \\
\cmidrule(lr){2-5}
 & \multirow{3}{*}{0.28} & Price $(C)$ & 0.002 (0.002) & 0.003 (0.002) \\
 &  & Delta $(\Delta)$ & 0.005 (0.002) & 0.006 (0.002) \\
 &  & Gamma $(\Gamma)$ & 0.030 (0.010) & 0.036 (0.012) \\
\cmidrule(lr){2-5}
 & \multirow{3}{*}{0.32} & Price $(C)$ & 0.002 (0.002) & 0.003 (0.002) \\
 &  & Delta $(\Delta)$ & 0.005 (0.002) & 0.006 (0.002) \\
 &  & Gamma $(\Gamma)$ & 0.029 (0.010) & 0.035 (0.012) \\
\cmidrule(lr){2-5}
 & \multirow{3}{*}{0.36} & Price $(C)$ & 0.002 (0.002) & 0.003 (0.002) \\
 &  & Delta $(\Delta)$ & 0.005 (0.002) & 0.006 (0.002) \\
 &  & Gamma $(\Gamma)$ & 0.028 (0.010) & 0.035 (0.012) \\
\cmidrule(lr){2-5}
 & \multirow{3}{*}{0.40} & Price $(C)$ & 0.002 (0.002) & 0.003 (0.002) \\
 &  & Delta $(\Delta)$ & 0.005 (0.002) & 0.006 (0.002) \\
 &  & Gamma $(\Gamma)$ & 0.028 (0.010) & 0.035 (0.012) \\
\cmidrule(lr){2-5}
 & \multirow{3}{*}{0.44} & Price $(C)$ & 0.002 (0.002) & 0.003 (0.002) \\
 &  & Delta $(\Delta)$ & 0.005 (0.002) & 0.006 (0.002) \\
 &  & Gamma $(\Gamma)$ & 0.027 (0.009) & 0.034 (0.012) \\
\cmidrule(lr){2-5}
 & \multirow{3}{*}{0.48} & Price $(C)$ & 0.002 (0.002) & 0.003 (0.002) \\
 &  & Delta $(\Delta)$ & 0.005 (0.002) & 0.006 (0.002) \\
 &  & Gamma $(\Gamma)$ & 0.026 (0.009) & 0.034 (0.012) \\
\bottomrule
\end{tabular}
}
\caption{FINN delta--gamma hedging performance at $\sigma=0.175$ (RMAD/RMSE). Standard deviations are in parentheses.}
\label{tab:gamma_rmad_rmse_vol0175}
\end{table}

\begin{table}
\centering
{%
\small\setlength{\tabcolsep}{6pt}
\begin{tabular}{cclrr}
\toprule
\multirow{2}{*}{Hedge TTM $\tau_h$} & \multirow{2}{*}{Option TTM $\tau$} & \multirow{2}{*}{Metric} & \multicolumn{2}{c}{Error} \\
\cmidrule(lr){4-5}
& & & NMAD & MSE \\
\midrule
\multirow{21}{*}{0.12} & \multirow{3}{*}{0.24} & Price $(C)$ & 0.003 (0.001) & 0.001 (0.001) \\
 &  & Delta $(\Delta)$ & 0.007 (0.002) & 0.000 (0.000) \\
 &  & Gamma $(\Gamma)$ & 0.051 (0.013) & 0.000 (0.000) \\
\cmidrule(lr){2-5}
 & \multirow{3}{*}{0.28} & Price $(C)$ & 0.003 (0.001) & 0.001 (0.001) \\
 &  & Delta $(\Delta)$ & 0.007 (0.002) & 0.000 (0.000) \\
 &  & Gamma $(\Gamma)$ & 0.050 (0.013) & 0.000 (0.000) \\
\cmidrule(lr){2-5}
 & \multirow{3}{*}{0.32} & Price $(C)$ & 0.003 (0.001) & 0.001 (0.001) \\
 &  & Delta $(\Delta)$ & 0.007 (0.002) & 0.000 (0.000) \\
 &  & Gamma $(\Gamma)$ & 0.049 (0.013) & 0.000 (0.000) \\
\cmidrule(lr){2-5}
 & \multirow{3}{*}{0.36} & Price $(C)$ & 0.003 (0.001) & 0.001 (0.001) \\
 &  & Delta $(\Delta)$ & 0.007 (0.002) & 0.000 (0.000) \\
 &  & Gamma $(\Gamma)$ & 0.048 (0.013) & 0.000 (0.000) \\
\cmidrule(lr){2-5}
 & \multirow{3}{*}{0.40} & Price $(C)$ & 0.003 (0.001) & 0.001 (0.001) \\
 &  & Delta $(\Delta)$ & 0.007 (0.002) & 0.000 (0.000) \\
 &  & Gamma $(\Gamma)$ & 0.047 (0.013) & 0.000 (0.000) \\
\cmidrule(lr){2-5}
 & \multirow{3}{*}{0.44} & Price $(C)$ & 0.003 (0.001) & 0.001 (0.001) \\
 &  & Delta $(\Delta)$ & 0.007 (0.002) & 0.000 (0.000) \\
 &  & Gamma $(\Gamma)$ & 0.046 (0.013) & 0.000 (0.000) \\
\cmidrule(lr){2-5}
 & \multirow{3}{*}{0.48} & Price $(C)$ & 0.003 (0.001) & 0.001 (0.001) \\
 &  & Delta $(\Delta)$ & 0.007 (0.002) & 0.000 (0.000) \\
 &  & Gamma $(\Gamma)$ & 0.045 (0.013) & 0.000 (0.000) \\
\midrule
\multirow{21}{*}{0.36} & \multirow{3}{*}{0.24} & Price $(C)$ & 0.004 (0.002) & 0.001 (0.001) \\
 &  & Delta $(\Delta)$ & 0.005 (0.002) & 0.000 (0.000) \\
 &  & Gamma $(\Gamma)$ & 0.042 (0.013) & 0.000 (0.000) \\
\cmidrule(lr){2-5}
 & \multirow{3}{*}{0.28} & Price $(C)$ & 0.004 (0.002) & 0.001 (0.001) \\
 &  & Delta $(\Delta)$ & 0.005 (0.002) & 0.000 (0.000) \\
 &  & Gamma $(\Gamma)$ & 0.040 (0.013) & 0.000 (0.000) \\
\cmidrule(lr){2-5}
 & \multirow{3}{*}{0.32} & Price $(C)$ & 0.004 (0.002) & 0.001 (0.001) \\
 &  & Delta $(\Delta)$ & 0.005 (0.002) & 0.000 (0.000) \\
 &  & Gamma $(\Gamma)$ & 0.038 (0.012) & 0.000 (0.000) \\
\cmidrule(lr){2-5}
 & \multirow{3}{*}{0.36} & Price $(C)$ & 0.004 (0.002) & 0.001 (0.001) \\
 &  & Delta $(\Delta)$ & 0.005 (0.002) & 0.000 (0.000) \\
 &  & Gamma $(\Gamma)$ & 0.036 (0.012) & 0.000 (0.000) \\
\cmidrule(lr){2-5}
 & \multirow{3}{*}{0.40} & Price $(C)$ & 0.004 (0.002) & 0.001 (0.001) \\
 &  & Delta $(\Delta)$ & 0.005 (0.002) & 0.000 (0.000) \\
 &  & Gamma $(\Gamma)$ & 0.035 (0.012) & 0.000 (0.000) \\
\cmidrule(lr){2-5}
 & \multirow{3}{*}{0.44} & Price $(C)$ & 0.004 (0.002) & 0.001 (0.001) \\
 &  & Delta $(\Delta)$ & 0.005 (0.002) & 0.000 (0.000) \\
 &  & Gamma $(\Gamma)$ & 0.033 (0.011) & 0.000 (0.000) \\
\cmidrule(lr){2-5}
 & \multirow{3}{*}{0.48} & Price $(C)$ & 0.004 (0.002) & 0.001 (0.001) \\
 &  & Delta $(\Delta)$ & 0.005 (0.002) & 0.000 (0.000) \\
 &  & Gamma $(\Gamma)$ & 0.031 (0.011) & 0.000 (0.000) \\
\bottomrule
\end{tabular}
}
\caption{FINN delta--gamma hedging performance at $\sigma=0.125$ (NMAD/MSE). Standard deviations are in parentheses.}
\label{tab:gamma_nmad_mse_vol0125}
\end{table}

\begin{table}
\centering
{%
\small\setlength{\tabcolsep}{6pt}
\begin{tabular}{cclrr}
\toprule
\multirow{2}{*}{Hedge TTM $\tau_h$} & \multirow{2}{*}{Option TTM $\tau$} & \multirow{2}{*}{Metric} & \multicolumn{2}{c}{Error} \\
\cmidrule(lr){4-5}
& & & NMAD & MSE \\
\midrule
\multirow{21}{*}{0.12} & \multirow{3}{*}{0.24} & Price $(C)$ & 0.003 (0.001) & 0.001 (0.001) \\
 &  & Delta $(\Delta)$ & 0.005 (0.002) & 0.000 (0.000) \\
 &  & Gamma $(\Gamma)$ & 0.033 (0.013) & 0.000 (0.000) \\
\cmidrule(lr){2-5}
 & \multirow{3}{*}{0.28} & Price $(C)$ & 0.003 (0.001) & 0.001 (0.001) \\
 &  & Delta $(\Delta)$ & 0.005 (0.002) & 0.000 (0.000) \\
 &  & Gamma $(\Gamma)$ & 0.032 (0.013) & 0.000 (0.000) \\
\cmidrule(lr){2-5}
 & \multirow{3}{*}{0.32} & Price $(C)$ & 0.003 (0.001) & 0.001 (0.001) \\
 &  & Delta $(\Delta)$ & 0.005 (0.002) & 0.000 (0.000) \\
 &  & Gamma $(\Gamma)$ & 0.031 (0.013) & 0.000 (0.000) \\
\cmidrule(lr){2-5}
 & \multirow{3}{*}{0.36} & Price $(C)$ & 0.003 (0.001) & 0.001 (0.001) \\
 &  & Delta $(\Delta)$ & 0.005 (0.002) & 0.000 (0.000) \\
 &  & Gamma $(\Gamma)$ & 0.030 (0.012) & 0.000 (0.000) \\
\cmidrule(lr){2-5}
 & \multirow{3}{*}{0.40} & Price $(C)$ & 0.003 (0.001) & 0.001 (0.001) \\
 &  & Delta $(\Delta)$ & 0.005 (0.002) & 0.000 (0.000) \\
 &  & Gamma $(\Gamma)$ & 0.029 (0.012) & 0.000 (0.000) \\
\cmidrule(lr){2-5}
 & \multirow{3}{*}{0.44} & Price $(C)$ & 0.003 (0.001) & 0.001 (0.001) \\
 &  & Delta $(\Delta)$ & 0.005 (0.002) & 0.000 (0.000) \\
 &  & Gamma $(\Gamma)$ & 0.028 (0.012) & 0.000 (0.000) \\
\cmidrule(lr){2-5}
 & \multirow{3}{*}{0.48} & Price $(C)$ & 0.003 (0.001) & 0.001 (0.001) \\
 &  & Delta $(\Delta)$ & 0.005 (0.002) & 0.000 (0.000) \\
 &  & Gamma $(\Gamma)$ & 0.027 (0.012) & 0.000 (0.000) \\
\midrule
\multirow{21}{*}{0.36} & \multirow{3}{*}{0.24} & Price $(C)$ & 0.004 (0.002) & 0.001 (0.001) \\
 &  & Delta $(\Delta)$ & 0.005 (0.002) & 0.000 (0.000) \\
 &  & Gamma $(\Gamma)$ & 0.035 (0.014) & 0.000 (0.000) \\
\cmidrule(lr){2-5}
 & \multirow{3}{*}{0.28} & Price $(C)$ & 0.004 (0.002) & 0.001 (0.001) \\
 &  & Delta $(\Delta)$ & 0.005 (0.002) & 0.000 (0.000) \\
 &  & Gamma $(\Gamma)$ & 0.033 (0.013) & 0.000 (0.000) \\
\cmidrule(lr){2-5}
 & \multirow{3}{*}{0.32} & Price $(C)$ & 0.004 (0.002) & 0.001 (0.001) \\
 &  & Delta $(\Delta)$ & 0.005 (0.002) & 0.000 (0.000) \\
 &  & Gamma $(\Gamma)$ & 0.032 (0.013) & 0.000 (0.000) \\
\cmidrule(lr){2-5}
 & \multirow{3}{*}{0.36} & Price $(C)$ & 0.004 (0.002) & 0.001 (0.001) \\
 &  & Delta $(\Delta)$ & 0.005 (0.002) & 0.000 (0.000) \\
 &  & Gamma $(\Gamma)$ & 0.030 (0.013) & 0.000 (0.000) \\
\cmidrule(lr){2-5}
 & \multirow{3}{*}{0.40} & Price $(C)$ & 0.004 (0.002) & 0.001 (0.001) \\
 &  & Delta $(\Delta)$ & 0.005 (0.002) & 0.000 (0.000) \\
 &  & Gamma $(\Gamma)$ & 0.029 (0.013) & 0.000 (0.000) \\
\cmidrule(lr){2-5}
 & \multirow{3}{*}{0.44} & Price $(C)$ & 0.004 (0.002) & 0.001 (0.001) \\
 &  & Delta $(\Delta)$ & 0.005 (0.002) & 0.000 (0.000) \\
 &  & Gamma $(\Gamma)$ & 0.028 (0.012) & 0.000 (0.000) \\
\cmidrule(lr){2-5}
 & \multirow{3}{*}{0.48} & Price $(C)$ & 0.004 (0.002) & 0.001 (0.001) \\
 &  & Delta $(\Delta)$ & 0.005 (0.002) & 0.000 (0.000) \\
 &  & Gamma $(\Gamma)$ & 0.026 (0.012) & 0.000 (0.000) \\
\bottomrule
\end{tabular}
}
\caption{FINN delta--gamma hedging performance at $\sigma=0.150$ (NMAD/MSE). Standard deviations are in parentheses.}
\label{tab:gamma_nmad_mse_vol0150}
\end{table}

\begin{table}
\centering
{%
\small\setlength{\tabcolsep}{6pt}
\begin{tabular}{cclrr}
\toprule
\multirow{2}{*}{Hedge TTM $\tau_h$} & \multirow{2}{*}{Option TTM $\tau$} & \multirow{2}{*}{Metric} & \multicolumn{2}{c}{Error} \\
\cmidrule(lr){4-5}
& & & NMAD & MSE \\
\midrule
\multirow{21}{*}{0.12} & \multirow{3}{*}{0.24} & Price $(C)$ & 0.003 (0.001) & 0.001 (0.001) \\
 &  & Delta $(\Delta)$ & 0.005 (0.002) & 0.000 (0.000) \\
 &  & Gamma $(\Gamma)$ & 0.028 (0.010) & 0.000 (0.000) \\
\cmidrule(lr){2-5}
 & \multirow{3}{*}{0.28} & Price $(C)$ & 0.003 (0.001) & 0.001 (0.001) \\
 &  & Delta $(\Delta)$ & 0.005 (0.002) & 0.000 (0.000) \\
 &  & Gamma $(\Gamma)$ & 0.027 (0.010) & 0.000 (0.000) \\
\cmidrule(lr){2-5}
 & \multirow{3}{*}{0.32} & Price $(C)$ & 0.003 (0.001) & 0.001 (0.001) \\
 &  & Delta $(\Delta)$ & 0.005 (0.002) & 0.000 (0.000) \\
 &  & Gamma $(\Gamma)$ & 0.026 (0.010) & 0.000 (0.000) \\
\cmidrule(lr){2-5}
 & \multirow{3}{*}{0.36} & Price $(C)$ & 0.003 (0.001) & 0.001 (0.001) \\
 &  & Delta $(\Delta)$ & 0.005 (0.002) & 0.000 (0.000) \\
 &  & Gamma $(\Gamma)$ & 0.025 (0.010) & 0.000 (0.000) \\
\cmidrule(lr){2-5}
 & \multirow{3}{*}{0.40} & Price $(C)$ & 0.003 (0.001) & 0.001 (0.001) \\
 &  & Delta $(\Delta)$ & 0.005 (0.002) & 0.000 (0.000) \\
 &  & Gamma $(\Gamma)$ & 0.024 (0.009) & 0.000 (0.000) \\
\cmidrule(lr){2-5}
 & \multirow{3}{*}{0.44} & Price $(C)$ & 0.003 (0.001) & 0.001 (0.001) \\
 &  & Delta $(\Delta)$ & 0.005 (0.002) & 0.000 (0.000) \\
 &  & Gamma $(\Gamma)$ & 0.023 (0.009) & 0.000 (0.000) \\
\cmidrule(lr){2-5}
 & \multirow{3}{*}{0.48} & Price $(C)$ & 0.003 (0.001) & 0.001 (0.001) \\
 &  & Delta $(\Delta)$ & 0.005 (0.002) & 0.000 (0.000) \\
 &  & Gamma $(\Gamma)$ & 0.022 (0.009) & 0.000 (0.000) \\
\midrule
\multirow{21}{*}{0.36} & \multirow{3}{*}{0.24} & Price $(C)$ & 0.003 (0.002) & 0.001 (0.001) \\
 &  & Delta $(\Delta)$ & 0.005 (0.002) & 0.000 (0.000) \\
 &  & Gamma $(\Gamma)$ & 0.028 (0.012) & 0.000 (0.000) \\
\cmidrule(lr){2-5}
 & \multirow{3}{*}{0.28} & Price $(C)$ & 0.003 (0.002) & 0.001 (0.001) \\
 &  & Delta $(\Delta)$ & 0.005 (0.002) & 0.000 (0.000) \\
 &  & Gamma $(\Gamma)$ & 0.027 (0.012) & 0.000 (0.000) \\
\cmidrule(lr){2-5}
 & \multirow{3}{*}{0.32} & Price $(C)$ & 0.003 (0.002) & 0.001 (0.001) \\
 &  & Delta $(\Delta)$ & 0.005 (0.002) & 0.000 (0.000) \\
 &  & Gamma $(\Gamma)$ & 0.026 (0.011) & 0.000 (0.000) \\
\cmidrule(lr){2-5}
 & \multirow{3}{*}{0.36} & Price $(C)$ & 0.003 (0.002) & 0.001 (0.001) \\
 &  & Delta $(\Delta)$ & 0.005 (0.002) & 0.000 (0.000) \\
 &  & Gamma $(\Gamma)$ & 0.025 (0.011) & 0.000 (0.000) \\
\cmidrule(lr){2-5}
 & \multirow{3}{*}{0.40} & Price $(C)$ & 0.003 (0.002) & 0.001 (0.001) \\
 &  & Delta $(\Delta)$ & 0.005 (0.002) & 0.000 (0.000) \\
 &  & Gamma $(\Gamma)$ & 0.025 (0.011) & 0.000 (0.000) \\
\cmidrule(lr){2-5}
 & \multirow{3}{*}{0.44} & Price $(C)$ & 0.003 (0.002) & 0.001 (0.001) \\
 &  & Delta $(\Delta)$ & 0.005 (0.002) & 0.000 (0.000) \\
 &  & Gamma $(\Gamma)$ & 0.024 (0.011) & 0.000 (0.000) \\
\cmidrule(lr){2-5}
 & \multirow{3}{*}{0.48} & Price $(C)$ & 0.003 (0.002) & 0.001 (0.001) \\
 &  & Delta $(\Delta)$ & 0.005 (0.002) & 0.000 (0.000) \\
 &  & Gamma $(\Gamma)$ & 0.023 (0.010) & 0.000 (0.000) \\
\bottomrule
\end{tabular}
}
\caption{FINN delta--gamma hedging performance at $\sigma=0.175$ (NMAD/MSE). Standard deviations are in parentheses.}
\label{tab:gamma_nmad_mse_vol0175}
\end{table}

\newpage
\clearpage
\subsection{Additional Results for Section Learning Prices Where Option Markets Do Not Exist}\label{app:ipo}

\begin{figure}[H]
    \centering
    \begin{subfigure}[t]{0.32\textwidth}
        \centering
        \includegraphics[width=\linewidth]{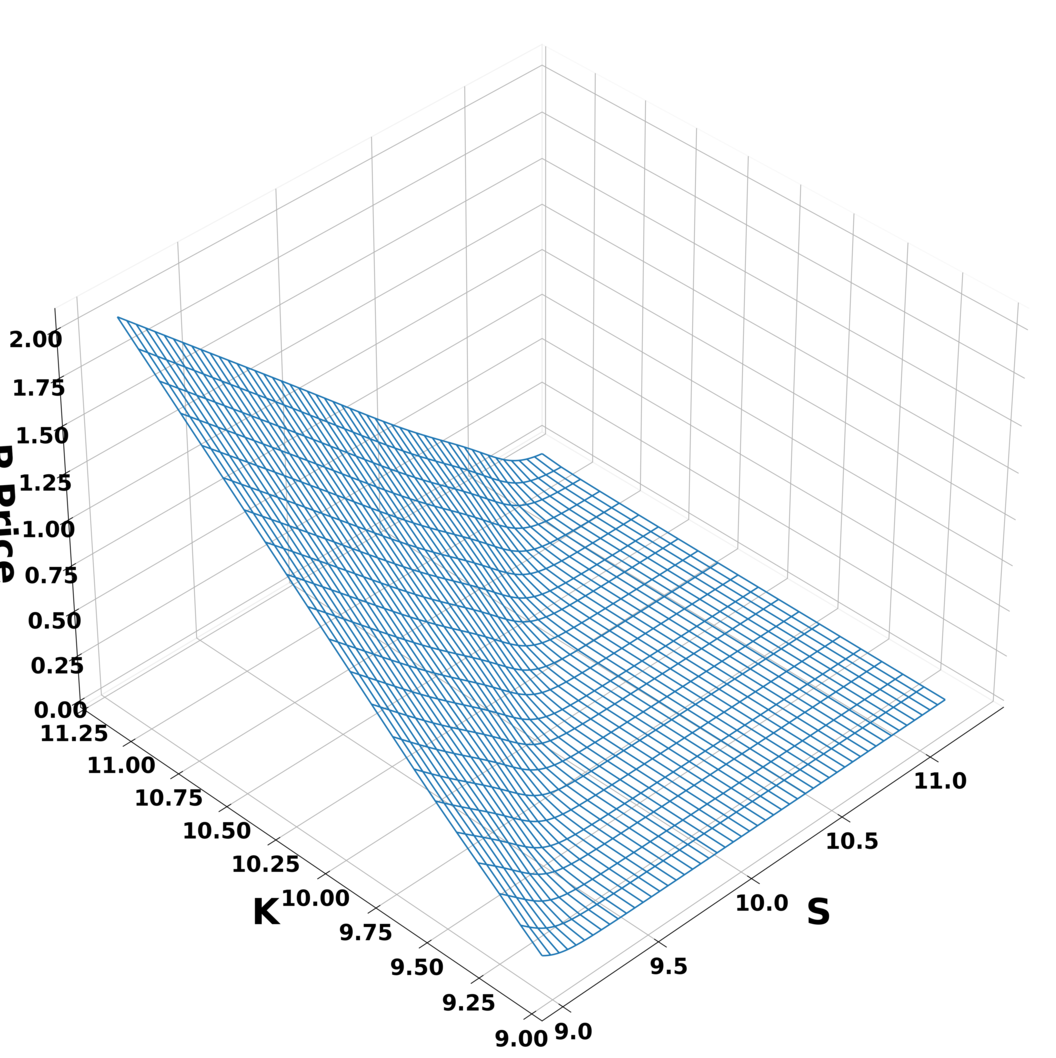}
        \caption{\texttt{COPL} Price}
        \label{fig:copl_put_price_surface}
    \end{subfigure}\hfill
    \begin{subfigure}[t]{0.32\textwidth}
        \centering
        \includegraphics[width=\linewidth]{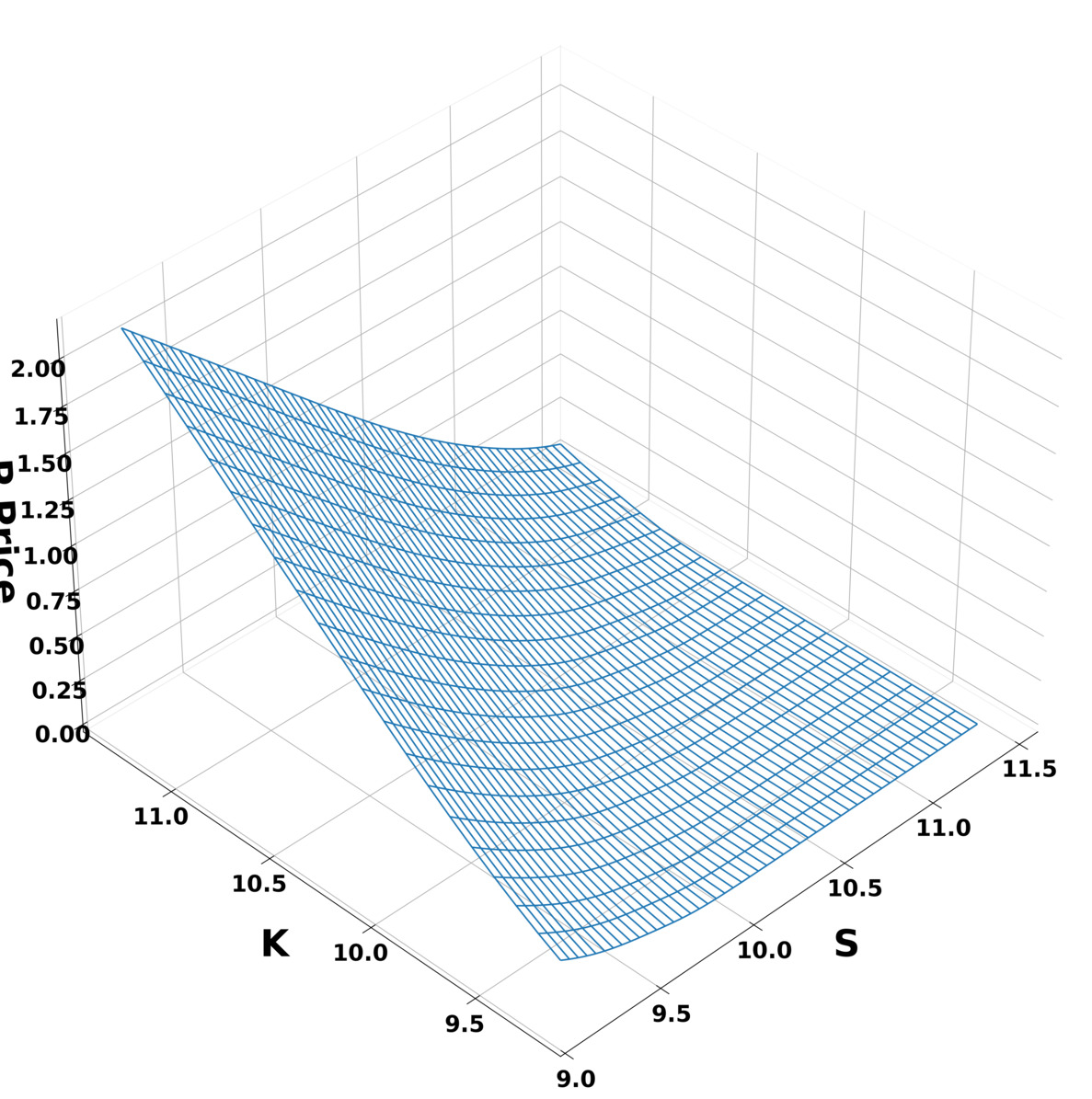}
        \caption{\texttt{GTEN} Price}
        \label{fig:gten_put_price_surface}
    \end{subfigure}\hfill
    \begin{subfigure}[t]{0.32\textwidth}
        \centering
        \includegraphics[width=0.8\linewidth]{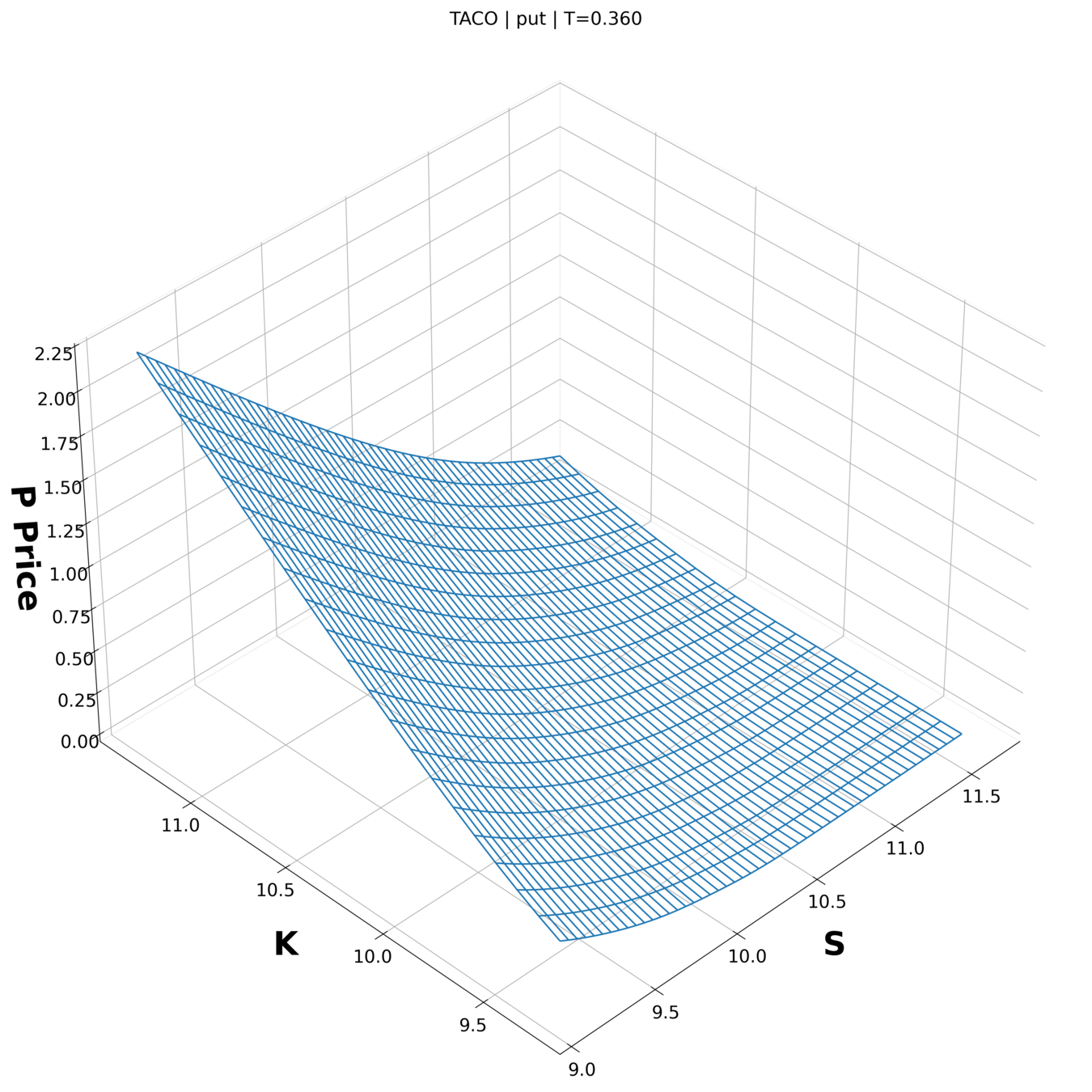}
        \caption{\texttt{TACO} Price}
        \label{fig:taco_put_price_surface}
    \end{subfigure}

    \vspace{0.6em}

    \begin{subfigure}[t]{0.32\textwidth}
        \centering
        \includegraphics[width=\linewidth]{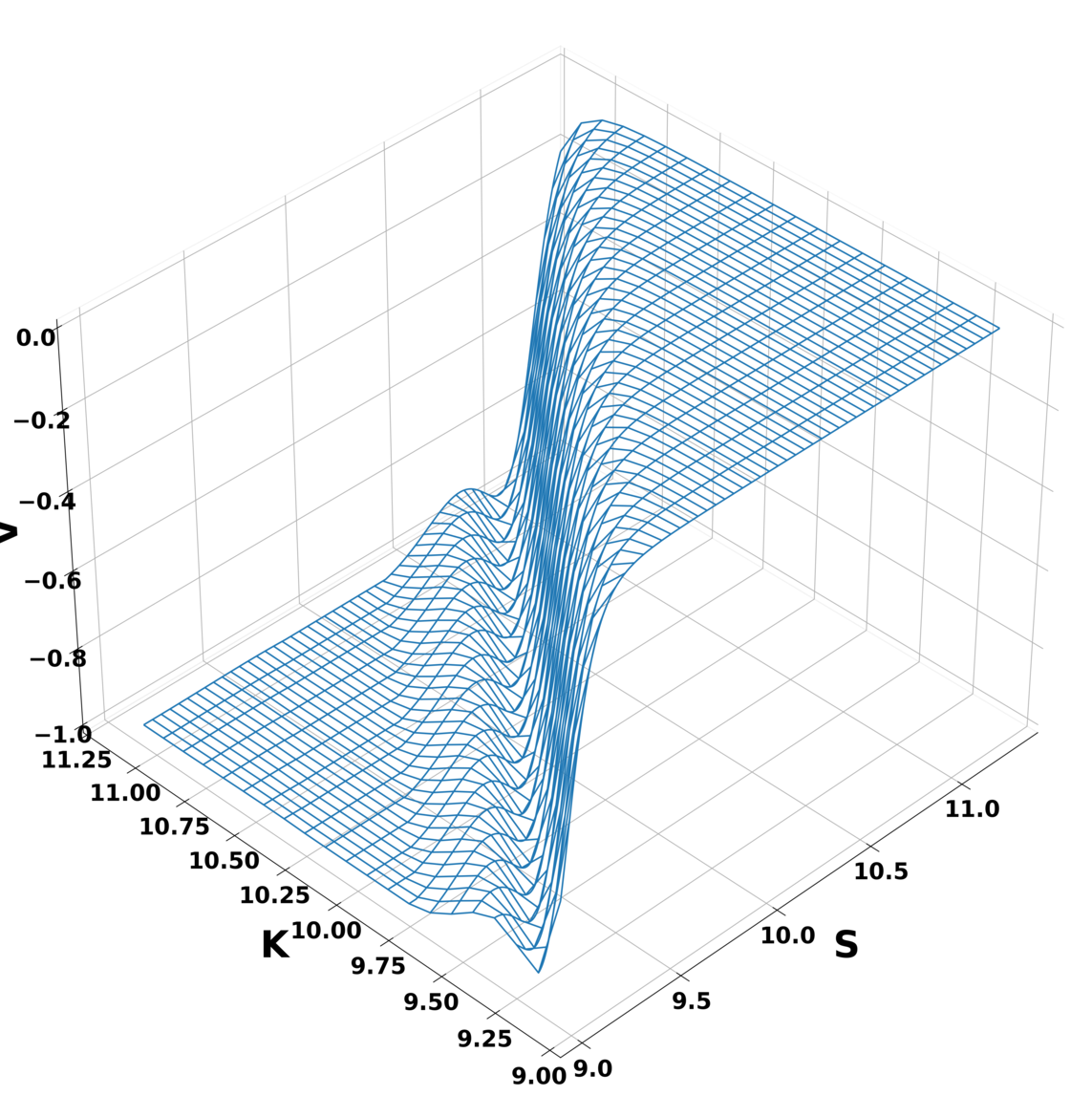}
        \caption{\texttt{COPL} $\Delta_{put}$}
        \label{fig:copl_put_delta_surface}
    \end{subfigure}\hfill
    \begin{subfigure}[t]{0.32\textwidth}
        \centering
        \includegraphics[width=\linewidth]{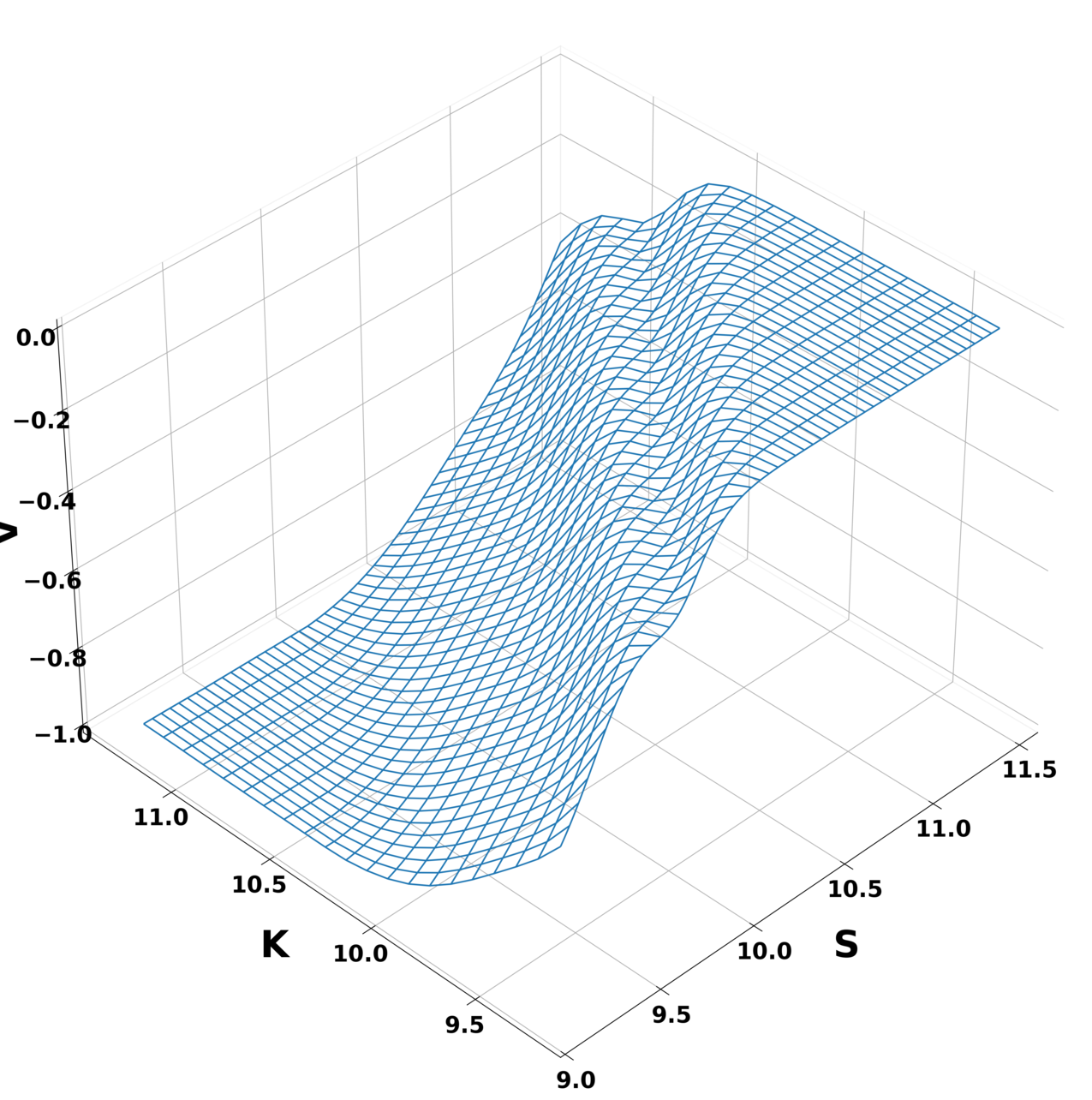}
        \caption{\texttt{GTEN} $\Delta_{put}$}
        \label{fig:gten_put_delta_surface}
    \end{subfigure}\hfill
    \begin{subfigure}[t]{0.32\textwidth}
        \centering
        \includegraphics[width=\linewidth]{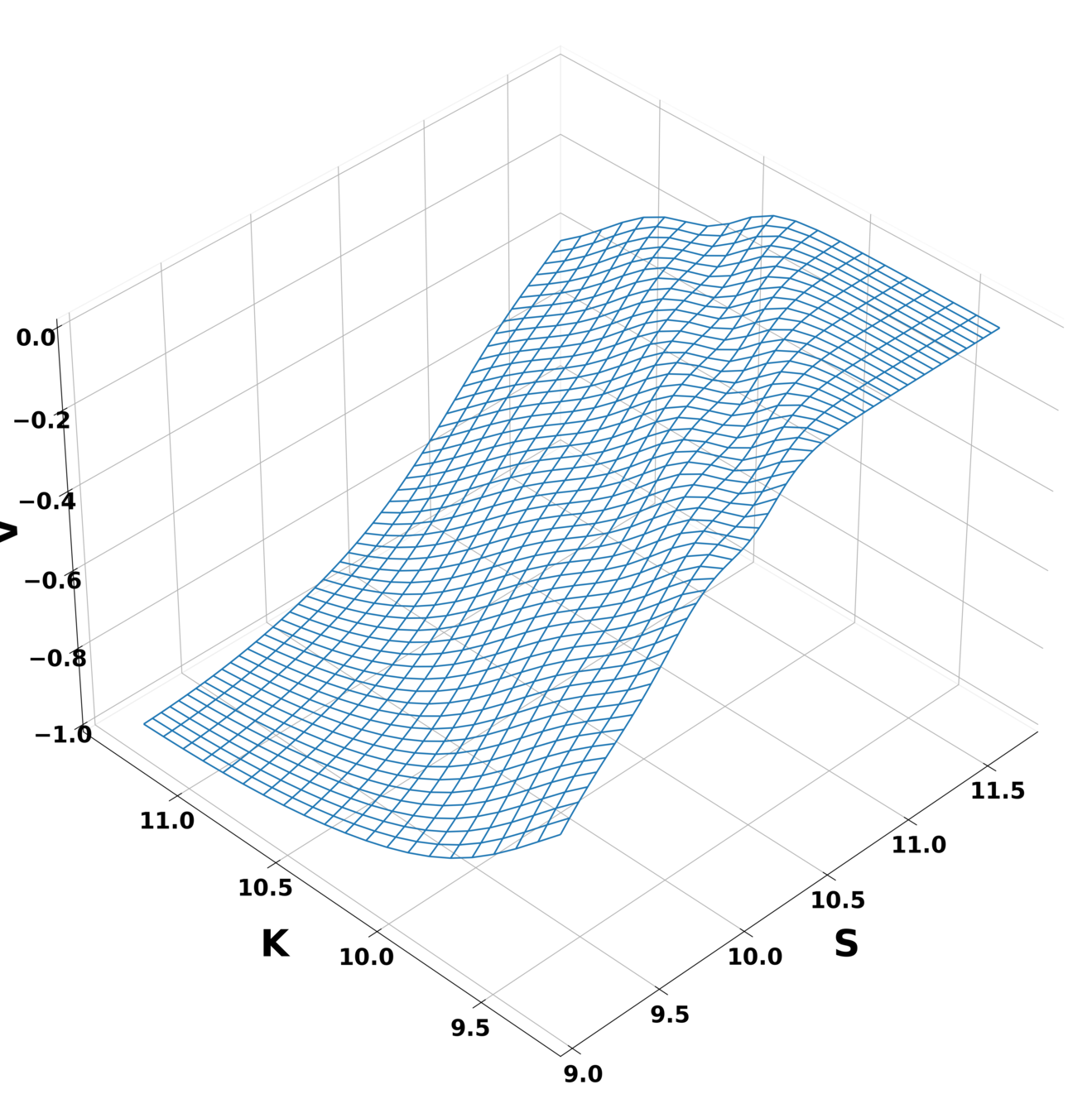}
        \caption{\texttt{TACO} $\Delta_{put}$}
        \label{fig:taco_put_delta_surface}
    \end{subfigure}

    \caption{FINN-implied European put option surfaces for instruments without listed options (TTM $=0.36$). Top row: option prices. Bottom row: corresponding hedge ratio delta.}
    \label{fig:put_price_delta_surfaces_all}
\end{figure}

\newpage
\clearpage
\bibliographystyle{unsrtnat}
\bibliography{references}  






\end{document}